\pdfoutput=1

\documentclass[11pt]{article}

\usepackage[final]{acl}

\usepackage{times}
\usepackage{latexsym}

\usepackage[T1]{fontenc}

\usepackage[utf8]{inputenc}

\usepackage{microtype}

\usepackage{inconsolata}

\usepackage{graphicx}
\usepackage{subfig}
\usepackage{amsmath}
\usepackage{bbding}
\usepackage{amssymb}
\usepackage{booktabs}
\usepackage{multirow}
\usepackage{arydshln}
\usepackage{colortbl}
\usepackage{dashrule}
\usepackage{fontawesome5}  
\usepackage{float}
\usepackage{overpic}
\usepackage{subfig}
\usepackage{xcolor}
\usepackage{tcolorbox}
\usepackage[colorinlistoftodos]{todonotes}

\title{LinguaLIFT: An Effective Two-stage Instruction Tuning Framework for Low-Resource Language Reasoning}

\author{Hongbin Zhang\textsuperscript{\textdagger\textdaggerdbl},Kehai Chen\textsuperscript{\textdagger}\thanks{Corresponding Author},Xuefeng Bai\textsuperscript{\textdagger},Yang Xiang\textsuperscript{\textdagger},Min Zhang\textsuperscript{\textdagger} \\
\textsuperscript{\textdagger}Institute of Computing and Intelligence, Harbin Institute of Technology, Shenzhen, China \\
\textsuperscript{\textdaggerdbl}Peng Cheng Laboratory, Shenzhen, China \\
\texttt{azure.starzhang@gmail.com,\{chenkehai,baixuefeng,zhangmin2021\}@hit.edu.cn,} \\
\texttt{xiangy@pcl.ac.cn} \\}

\begin{document}
\maketitle
\begin{abstract}
Large language models (LLMs) have exhibited impressive multilingual reasoning capabilities, driven by extensive multilingual pre-training corpora and instruction fine-tuning data. 
However, a performance gap exists between high- and low-resource language reasoning tasks due to the language imbalance in the pre-training corpus, which is exacerbated by evaluation bias in existing reasoning benchmarks lacking low-resource language coverage.
To alleviate this issue, we propose LinguaLIFT, a two-stage instruction tuning framework for advancing low-resource language reasoning. 
LinguaLIFT employs a language alignment layer to capture multilingual alignment in a code-switched tuning way without requiring multilingual instruction or parallel data, thereby transferring the cross-lingual reasoning capabilities to low-resource languages through English-only instruction tuning data.
To comprehensively evaluate the multilingual reasoning capabilities, we introduce the Multilingual Math World Problem (MMWP) benchmark, which spans 21 low-resource, 17 medium-resource, and 10 high-resource languages. 
Experimental results show that LinguaLIFT outperforms several competitive baselines across MMWP and four widely used benchmarks.
\footnote{Our code and data will be released once accepted.} 
\end{abstract}

\section{Introduction}
Large Language Models (LLMs) have demonstrated impressive reasoning capabilities, opening up a new paradigm for artificial intelligence \citep{NEURIPS2022_9d560961,zhao2023survey}.
These capabilities arise from self-supervised pre-training on large-scale multilingual data, followed by supervised fine-tuning on multilingual instruction data~\citep{NEURIPS2022_b1efde53,muennighoff-etal-2023-crosslingual,JMLR:v25:23-0870}. 
As a result, LLMs have achieved remarkable performance across various natural language processing tasks~\citep{chen2024llm,chen-etal-2024-dual,zhang-etal-2024-paying}, especially high-resource language mathematical reasoning tasks~\citep{shi2023language,she-etal-2024-mapo}.

\begin{figure}[!t]
    \centering
    \includegraphics[width=1\linewidth]{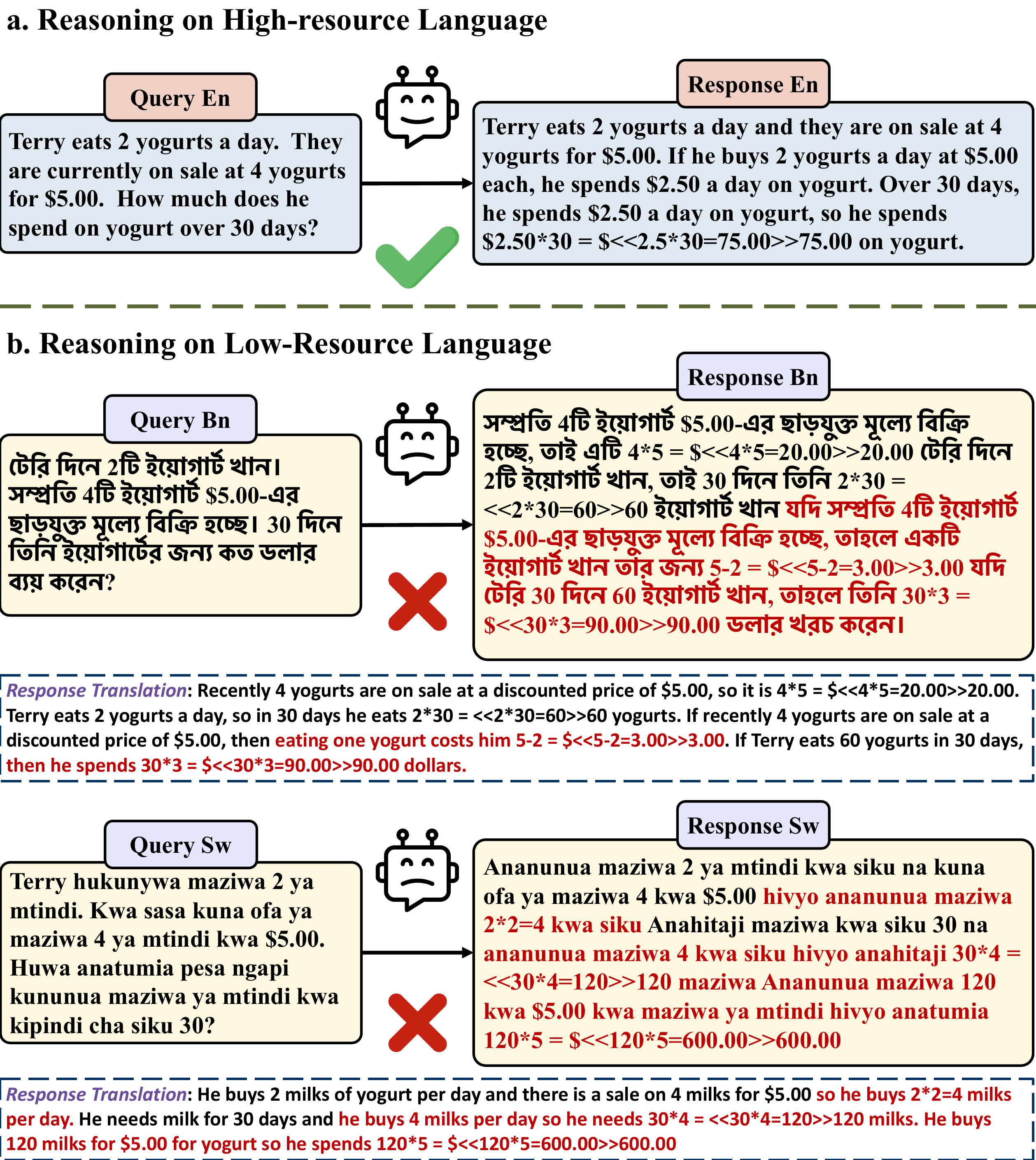}
    \caption{Examples from the MGSM~\citep{shi2023language} dataset, where the mathematical problems share the same meaning across languages, but LLMs generate different answers. The red text marks the erroneous reasoning in the responses. Translations of the responses are provided in dashed boxes.}
    \label{fig:intro}
\end{figure}

However, a substantial performance gap still exists between high- and low-resource language reasoning~\citep{shi2023language,zhu-etal-2024-question}, primarily due to the language imbalance within the large-scale multilingual corpora during pre-training, where high-resource languages dominate~\citep{touvron2023llama,lu-etal-2024-llamax}. 
This gap is further exacerbated by evaluation bias inherent in existing multilingual reasoning benchmarks, which exhibit limited coverage of low-resource languages~\cite{shi2023language,chen2023breaking}. 
As illustrated in Figure ~\ref{fig:intro}, while LLMs can answer correctly to the query in English, they struggle with the same query in low-resource languages (e.g., Bengali (Bn), Swahili (Sw)). 
These discrepancies emphasize LLMs' fundamental limitation in low-resource language reasoning.

To alleviate this issue, this study proposes a two-stage instruction tuning framework (LinguaLIFT). 
The first stage introduces a language alignment layer to adapt a pre-trained multilingual encoder, utilizing code-switched translation data to enhance multilingual alignment. 
Building on this strengthened alignment, the second stage fine-tunes the LLM on English instruction data while keeping the alignment layer frozen, thereby transferring reasoning capabilities from English to low-resource languages.
This study builds a new benchmark named Multilingual Math World Problem (MMWP), spanning 21 low-resource, 17 medium-resource, and 10 high-resource languages. 
Experiments demonstrate that LinguaLIFT significantly outperforms several strong competitive methods on the MMWP and four widely used benchmarks, such as MGSM~\citep{shi2023language}, MSVAMP~\citep{chen2023breaking}, XNLI~\citep{conneau-etal-2018-xnli} and X-CSQA~\citep{lin-etal-2021-common}.

To summarize, our primary contributions are:
\begin{itemize}
    \item The proposed LinguaLIFT significantly enhances low-resource language reasoning without requiring multilingual or parallel data.
    \item This study introduces MMWP, a new mathematical reasoning benchmark covering languages at varying resource levels, to evaluate multilingual reasoning tasks comprehensively.
    \item Experiments across mathematical and commonsense reasoning demonstrate the effectiveness and generalizability of LinguaLIFT in advancing low-resource language reasoning.
\end{itemize}

\section{Related Work}
\paragraph{Leveraging cross-lingual transfer within multilingual models.}
While recent research highlights multilingual models' robust cross-lingual transfer capabilities~\citep{conneau-etal-2020-unsupervised,fitzgerald-etal-2023-massive,shaham-etal-2024-multilingual}—achieved through alignment improvements via multilingual instruction or parallel data~\citep{pfeiffer-etal-2020-mad,pan-etal-2021-multilingual,feng-etal-2022-language,li2023bactrian,chen-etal-2024-monolingual}—such approaches remain impractical in low-resource scenarios due to data scarcity.
To address this limitation, LinguaLIFT advances code-switched tuning, demonstrating success in cross-lingual transfer~\citep{yang-etal-2020-csp,li-etal-2024-prealign}, to enhance multilingual alignment without requiring multilingual instructions or parallel corpora.

\paragraph{Improving multilingual mathematical reasoning tasks.}
Recent efforts to improve multilingual mathematical reasoning for LLMs can be categorized into three ways: 1) \textbf{Prompting close-source LLMs:} \citet{qin-etal-2023-cross}, \citet{shi2023language}, and \citet{huang-etal-2023-languages} designed prompts for closed-source LLMs like ChatGPT, translating non-English contexts into English for reasoning. This approach is limited by translation quality and does not improve multilingual understanding or works well for open-source LLMs \citep{zhu-etal-2024-question}. 2) \textbf{Instruction-tuning open-source LLMs:} \textit{MathOctopus}~\citep{chen2023breaking}, \textit{xCoT}~\citep{chai2024xcot}, and \textit{mCoT}~\citep{lai-nissim-2024-mcot} adopted a translate-training method, translating English reasoning datasets into non-English and instruction-tuning LLMs. \textit{QAlign}~\citep{zhu-etal-2024-question} and \textit{MAPO}~\citep{she-etal-2024-mapo} proposed approaches to transfer mathematical reasoning capabilities from English to non-English. While these methods improve multilingual reasoning, they incur high translation costs and errors, making them impractical for low-resource languages. 3) \textbf{Bridging existing skilled LLMs to multilingualism:} \textit{Langbridge}~\citep{yoon-etal-2024-langbridge} combines pre-trained multilingual models with skilled reasoning LLMs, but a performance gap persists between low-resource and high-resource languages. While \textit{MindMerger}~\citep{huang2024mindmerger} shows significant improvement, it relies heavily on parallel and multilingual reasoning data.

\begin{figure*}[!htbp]
    \centering
    \includegraphics[width=0.95\linewidth]{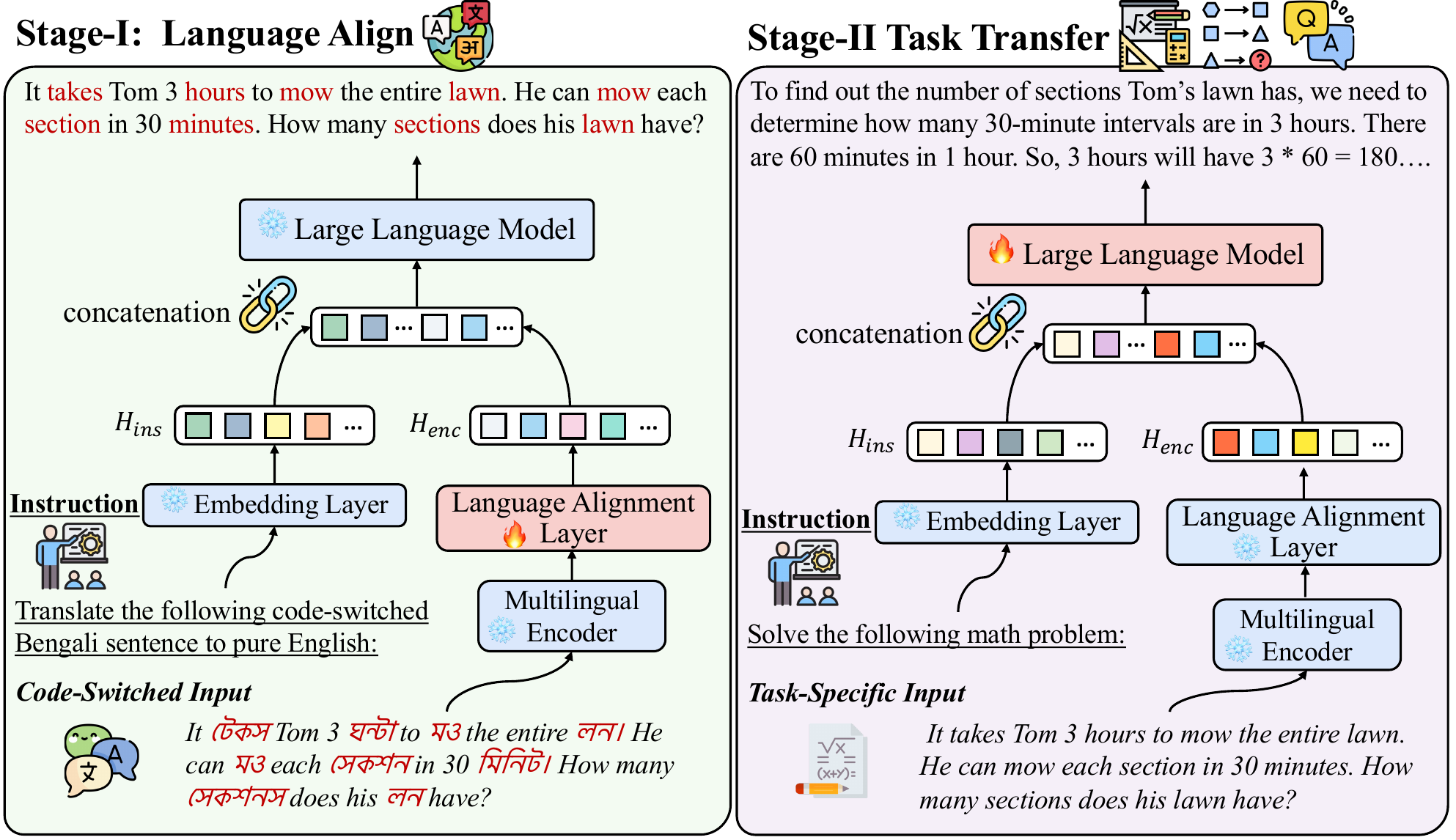}
    \caption{Overview of the proposed LinguaLIFT approach. \textbf{Stage-I (Language Align)}: A language alignment layer is introduced into the LLM to adapt the pre-trained multilingual encoder, thereby enhancing multilingual alignment through code-switched tuning. \textbf{Stage-II (Task Transfer)}: The LLM is fine-tuned on high-quality, English-only instruction data with the language alignment layer frozen, allowing the LLM to transfer reasoning capabilities learned from English to low-resource languages.}
    \label{methodology}
\end{figure*}

Unlike prior approaches that depend on costly translated data or impractical parallel corpora in low-resource settings, LinguaLIFT advances low-resource language reasoning without requiring translated multilingual or parallel datasets.

\section{Methodology}
\label{sec:method}
To harness cross-lingual transfer for low-resource languages reasoning, this study proposes leveraging the language alignment established by code-switched tuning (Section \S\ref{method.language_align}) to transfer reasoning capabilities from English to low-resource languages (Section \S\ref{method.task_transfer}), as illustrated in Figure~\ref{methodology}.
Formally, given an input sequence $x$ with $l_x$ tokens, a multilingual pre-trained encoder generates a language-agnostic representation $X\in \mathbb{R}^{l_x \times d_1}$, which is subsequently aligned to the input embedding space of the LLM via a multi-layer perceptron alignment layer, yielding $\hat{X}\in \mathbb{R}^{l_x \times d_2}$.
To leverage the built-in capabilities of the LLM for instruction following, the instruction context $q$ (with $l_q$ tokens) is mapped into $Q\in \mathbb{R}^{l_q \times d_2}$ via the LLM's embedding layer. The combined input sequence is constructed by concatenating the start token <bos>, $Q$, alignment boundary tokens <enc\_start> and <enc\_end>, and $\hat{X}$, forming:
\begin{equation}
    (Q, \hat{X}) = [\text{<bos>}; Q; \text{<enc\_start>}; \hat{X}; \text{<enc\_end>}].
\label{input_embedding}
\end{equation}
This structured input is then processed by the LLM to generate the final response.

\subsection{Language Align Stage}
\label{method.language_align}
Due to the scarcity of parallel corpora in low-resource languages, this study proposes enhancing the model's multilingual alignment through code-switched tuning. 
This involves two key parts: (i) multilingual alignment lexicon construction and (ii) code-switched translation instruction tuning.

\paragraph{Multilingual alignment lexicon construction.}  
An unsupervised word translation method MUSE~\citep{lample2018word} is adopted to construct the multilingual alignment lexicons without parallel data. 
Specifically, a set of unique words $\mathcal{W}=\{w\}_i^{N}$ is extracted from English monolingual corpus $\mathcal{D}$, where $N$ is the number of unique words, excluding named entities and stop words. Word translations are obtained by retrieving the top nearest neighbor via projection of word embeddings into the target space.
Further implementation details and validation steps in Appendix~\ref{apdx:align-vocab} ensure the robustness and reliability of our constructed lexicons.

\paragraph{Code-switched translation instruction tuning.}
The constructed multilingual alignment lexicons are used to generate code-switched inputs for instruction tuning.
Formally, our training objective is to predict the English translation response tokens $y$ given the instruction context $q$ and code-switched input tokens $x_{l}$ in language $l$.
The input embeddings for this task, represented as $(Q, \hat{X_l})$ from Equation (\ref{input_embedding}), serve as the LLMs' input. 
The optimization objective is defined as follows:
\begin{equation}
    -\underset{\theta}{\arg \min } \sum_{l \in L} \sum_{i=1}^{T} \log p_{\theta,\psi ,\phi} (y_{i} \mid (Q,\hat{X}_{l}), y_{<i}),
\end{equation}
where $T$ is the target sequence length, $y_{<i}$ represents the preceding tokens, $y_i$ denotes the $i$-th token, and $L$ is the set of target languages. Trainable parameters include $\theta$ (language alignment layer and boundary tokens), while $\psi $ (multilingual encoder) and $\phi$ (LLM) remain frozen.

\subsection{Task Transfer Stage}
\label{method.task_transfer}
Following the language alignment established in the \textit{Language Align Stage}, the LLM is fine-tuned using high-quality English reasoning instruction data while keeping the language alignment layer frozen. This strategy preserves the acquired multilingual alignment and cross-lingual transfer capabilities, unlocking the potential of LLM to transfer strong reasoning capabilities learned from English to low-resource languages. 
The training minimizes the negative log-likelihood of the response $y$, given the combined instruction context and query input $(Q, \hat{X}_{en})$:
\begin{equation}
    -\underset{\phi}{\arg \min }  \sum_{i=1}^{T} \log p_{\tilde{\theta},\psi ,\phi}\left(y_{i} \mid (Q,\hat{X}_{en}), y_{<i}\right),
\end{equation}
where $\phi$ denotes the trainable LLM parameter, and $\tilde{\theta}$ is initialized from the checkpoint of $\theta$ trained at the first stage. The multilingual encoder parameters $\psi$ and $\tilde{\theta}$ are frozen throughout the training.

\section{The MMWP Benchmark}
Existing multilingual mathematical reasoning benchmarks~\citep{shi2023language,chen2023breaking} predominantly focus on 7 high-resource languages and 3 low-resource languages, leaving a significant gap in coverage for low-resource languages. This imbalance introduces evaluation bias, as models are optimized for high-resource languages while their performance on low-resource languages remains underexplored, hindering the comprehensive development of multilingual models. 
To fill this gap, we build a new Multilingual Math World Problem (MMWP) benchmark and describe its collection process in this section. 

\paragraph{Source data.} We use AsDiV~\citep{miao-etal-2020-diverse} and MAWPS~\citep{koncel-kedziorski-etal-2016-mawps} as base datasets. From the AsDiv official test set, we randomly select 500 examples and another 500 from MAWPS, where all the problems require multiple steps to solve, as described by \citet{miao-etal-2020-diverse}. We filter out duplicates and problems with non-numeric answers, resulting in 811 examples.

\paragraph{Target language selection.} Our study employs a typologically diverse sample of 48 languages representing 12 language families and 12 writing systems, encompassing low-, medium-, and high-resource levels. The selection comprises 21 low-resource, 17 medium-resource, and 10 high-resource languages. Language categorization list and selection criteria are detailed in Appendices~\ref{apdx:lang_list} and \ref{apdx:resource-level-categorization}, respectively.

\paragraph{Translation and post-calibration.} 
English questions are translated into target languages using Google Translate. Translation quality is verified through combined automated and human evaluations. Five annotators then refine inaccuracies via GPT-4-guided post-calibration, followed by reviews and adjustments from two language experts prominent in minor languages. Detailed accuracy assurance procedures are provided in Appendix~\ref{apdx:quality-estimation}.

\section{Experiments}
\label{sec:exp}
\subsection{Datasets}

\paragraph{Evaluation dataset.} We use the MMWP and the latest multilingual benchmarks, \textbf{MGSM}~\citep{shi2023language} and \textbf{MSVAMP}~\citep{chen2023breaking} to evaluate the performance of LLMs in multilingual mathematical reasoning through zero-shot chain-of-thought reasoning~\citep{NEURIPS2022_9d560961} setting.
To further assess the task generalization of LinguaLIFT, we incorporate several challenging multilingual datasets, including \textbf{X-CSQA}~\citep{lin-etal-2021-common} for commonsense reasoning and \textbf{XNLI}~\citep{conneau-etal-2018-xnli} for natural language inference. 

\paragraph{Training dataset.} We utilize English-only instruction data following prior work~\citep{lu-etal-2024-llamax,huang2024mindmerger}, which include \textbf{MetaMathQA}~\citep{yu2024metamath} for mathematical reasoning, \textbf{MultiNLI}~\citep{williams-etal-2018-broad} for natural language inference, and a set of unified commonsense reasoning tasks, comprising the \textbf{X-CSQA}, \textbf{OpenBookQA}~\citep{mihaylov-etal-2018-suit}, \textbf{ARC}~\citep{clark2018think}, and \textbf{QASC}~\citep{Khot_Clark_Guerquin_Jansen_Sabharwal_2020} datasets. Additionally, to further explore the potential of the proposed method, we incorporate the recent advanced mathematical reasoning instruction dataset, \textbf{OpenMathInstruct-2}~\citep{toshniwal2024openmathinstruct-2}. 
Statistics of the datasets involved are presented in Appendix~\ref{apdx:dataset}, and the prompts for each task are given in Appendix~\ref{apdx:prompts}.

\subsection{Baselines}
Three categories of baselines are considered:
(1) \textbf{Mono-SFT}: \textit{MAmmoTH}~\citep{yue2024mammoth}, \textit{WizardMath}~\citep{luo2023wizardmath}, \textit{MetaMath}~\citep{yu2024metamath}, and \textit{OpenMath2}~\citep{toshniwal2024openmathinstruct-2} fine-tune on respective English instruction dataset.
(2) \textbf{Multi-SFT}: \textit{MathOctopus}~\citep{chen2023breaking}, \textit{QAlign}~\citep{zhu-etal-2024-question}, \textit{MAPO}~\citep{she-etal-2024-mapo} fine-tune on translated multilingual instruction datasets (e.g., MGSM8K-Instruct~\citep{chen2023breaking}) or parallel corpus.
(3) \textbf{Leveraging External Tools or Models}: \textit{Translate-En}~\citep{shi2023language} use SOTA translation models~\citep{nllbteam2022languageleftbehindscaling} to translate the query into English; \textit{LangBridge}~\citep{yoon-etal-2024-langbridge} and \textit{MindMerger}~\citep{huang2024mindmerger} replace the input of LLMs with the hidden states output by mT5~\citep{xue-etal-2021-mt5}. To ensure a fair comparison with MindMerger (the previous SOTA), we further consider continuing training (\textit{Add Low-Resource}) or retraining (\textit{Low-Resource Retrain}) with open-source low-resource parallel data.
More training details are presented in Appendix~\ref{apdx:trainDetails}.

\subsection{Experimental Results}

\begin{table}[!t]
\centering
\resizebox{\linewidth}{!}{
\begin{tabular}{lccccc}
\hline
\multirow{2}{*}{\textbf{LLaMA-2-7B as base model}} &
  \multicolumn{1}{c}{\textbf{Use}} &
  \multirow{2}{*}{\textbf{LR.}} &
  \multirow{2}{*}{\textbf{MR.}} &
  \multirow{2}{*}{\textbf{HR.}} &
  \multirow{2}{*}{\textbf{Avg.}} \\
 & \multicolumn{1}{c}{\textbf{Multi.}} &  &  &  &  \\ \hline
\rowcolor[HTML]{F1F1F1} 
\multicolumn{6}{c}{\cellcolor[HTML]{F1F1F1}\textsc{baseline}}                                    \\
\rowcolor[HTML]{F5EFF8} 
\multicolumn{6}{c}{\cellcolor[HTML]{F5EFF8}\textit{\textbf{Mono-SFT}}}                           \\
\rowcolor[HTML]{F5EFF8} 
MAmmoTH\textsuperscript{\textdagger}                        & \XSolidBrush & 6.36 & 20.4 & 26.1          & 15.5 \\
\rowcolor[HTML]{F5EFF8} 
WizardMath\textsuperscript{\textdagger}                     & \XSolidBrush & 10.1 & 28.0 & 32.2          & 21.0 \\
\rowcolor[HTML]{F5EFF8} 
MetaMath\textsuperscript{\textdagger}                       & \XSolidBrush & 13.6 & 37.2 & 41.8          & 27.8 \\
\rowcolor[HTML]{F5EFF8} 
OpenMath2\textsuperscript{\textdaggerdbl}                   & \XSolidBrush & 19.3 & 55.5 & \textbf{62.7} & 41.1 \\
\rowcolor[HTML]{E6EEF7} 
\multicolumn{6}{c}{\cellcolor[HTML]{E6EEF7}\textit{\textbf{Multi-SFT}}}                          \\
\rowcolor[HTML]{E6EEF7} 
MathOctopus-Parallel\textsuperscript{\textdagger}           & \Checkmark & 11.0 & 20.8 & 25.2          & 17.4 \\
\rowcolor[HTML]{E6EEF7} 
MathOctopus-MAPO-DPO\textsuperscript{\textdagger}           & \Checkmark & 18.6 & 29.8 & 33.1          & 25.6 \\
\rowcolor[HTML]{E6EEF7} 
MetaMathOctopus-MAPO-DPO\textsuperscript{\textdagger}       & \Checkmark & 16.4 & 36.7 & 43.9          & 29.3 \\
\rowcolor[HTML]{E6EEF7} 
QAlign-MetaMathQA\textsuperscript{\textdagger}              & \Checkmark & 17.1 & 39.0 & 44.5          & 30.6 \\
\rowcolor[HTML]{E1EFE1} 
\multicolumn{6}{c}{\cellcolor[HTML]{E1EFE1}\textit{\textbf{Leveraging External Tools or Models}}}                                                                 \\
\rowcolor[HTML]{E1EFE1} 
Translate-En\textsuperscript{\textdaggerdbl}~(MetaMath)     & \XSolidBrush & 27.6 & 36.4 & 40.6          & 33.4 \\
\rowcolor[HTML]{E1EFE1} 
LangBridge\textsuperscript{\textdagger}~(MetaMath)          & \XSolidBrush & 33.4 & 36.8 & 39.4          & 35.9 \\
\rowcolor[HTML]{E1EFE1} 
MindMerger-Soft\textsuperscript{\textdagger}~(MetaMath)     & \Checkmark & 36.6 & 40.7 & 43.2          & 39.4 \\
\rowcolor[HTML]{E1EFE1} 
\multicolumn{1}{r}{\cellcolor[HTML]{E1EFE1}(Add Low-Resource)\textsuperscript{\textdaggerdbl}}     & \Checkmark & 36.7          & 39.8          & 41.5         & 38.8          \\
\rowcolor[HTML]{E1EFE1} 
\multicolumn{1}{r}{\cellcolor[HTML]{E1EFE1}(Low-Resource Retrain)\textsuperscript{\textdaggerdbl}} & \Checkmark & 33.4          & 36.8          & 39.4         & 35.9          \\
\rowcolor[HTML]{E1EFE1} 
Translate-En\textsuperscript{\textdaggerdbl}~(OpenMath2)    & \XSolidBrush & 43.1 & 54.7 & 61.5          & 51.0 \\
\rowcolor[HTML]{E1EFE1} 
LangBridge\textsuperscript{\textdaggerdbl}~(OpenMath2)      & \XSolidBrush & 47.9 & 53.8 & 59.2          & 52.3 \\
\rowcolor[HTML]{E1EFE1} 
MindMerger-Soft\textsuperscript{\textdaggerdbl}~(OpenMath2) & \Checkmark & 49.5 & 54.8 & 61.2          & 53.8 \\ \hline
\rowcolor[HTML]{FBE5D6} 
\multicolumn{6}{c}{\cellcolor[HTML]{FBE5D6}\textsc{our methods}}                                 \\
\rowcolor[HTML]{FBE5D6} 
\multicolumn{1}{l}{\cellcolor[HTML]{FBE5D6}\textbf{LinguaLIFT~(MetaMathQA)}}                         & \XSolidBrush & 41.2          & 44.3          & 45.5         & 43.2          \\
\rowcolor[HTML]{FBE5D6} 
\multicolumn{1}{l}{\cellcolor[HTML]{FBE5D6}\textbf{LinguaLIFT\textsuperscript{$\Diamond$}~(MetaMathQA)}}                         & \XSolidBrush & 38.6          & 42.7          & 44.1         & 41.2          \\
\rowcolor[HTML]{FBE5D6} 
\multicolumn{1}{l}{\cellcolor[HTML]{FBE5D6}\textbf{LinguaLIFT~(OpenMathInstruct-2)}}                        & \XSolidBrush & \textbf{55.4} & \textbf{61.2} & 62.5         & \textbf{58.9} \\ 
\rowcolor[HTML]{FBE5D6} 
\multicolumn{1}{l}{\cellcolor[HTML]{FBE5D6}\textbf{LinguaLIFT\textsuperscript{$\Diamond$}~(OpenMathInstruct-2)}}                         & \XSolidBrush & 53.5          & 59.0          & 60.8         & 57.0          \\ \hline
\end{tabular}
}
\caption{Results on the MMWP benchmark, where ``LR.'', ``MR.'', and ``HR.'' denote mean accuracy for low-, medium-, and high-resource languages, respectively. ``Use Multi.'' denotes whether multilingual or parallel corpus is utilized. Bold values indicate the highest score across systems. $\dagger$ marks results obtained using officially released models, $\Diamond$ denotes the LoRA~\citep{hu2022lora} implementation, and $\ddagger$ signifies results from our local implementation. The LLM and extra data used in the third baseline category are specified in brackets.}
\label{main-result-MMWP-48}
\end{table}

\paragraph{LinguaLIFT demonstrates significant improvements across low-resource languages.}
Table~\ref{main-result-MMWP-48} presents experimental results on the \textsc{MMWP} test set, grouped by language resource levels. We highlight seven key findings: (1) Mono-SFT models exhibit substantial performance drops in low-resource languages. (2) Multi-SFT models outperform monolingual counterparts in low-resource settings but retain a notable performance gap between low- and high-resource languages. (3) Models using external translation systems or pre-trained multilingual models generalize poorly to unseen languages and underperform in low-resource contexts. 
(4) Even with supplementary low-resource data, MindMerger fails to improve low-resource language reasoning performance. (5) LinguaLIFT substantially improves low-resource reasoning accuracy, surpassing all baselines without requiring multilingual or parallel data. (6) Integrating advanced English reasoning datasets (e.g., OpenMathInstruct-2) further enhances reasoning performance, underscoring the need to adapt to evolving high-quality English reasoning data. 

\paragraph{Robust generalization and task versatility.}
\begin{table}[!htbp]
\centering
\resizebox{\linewidth}{!}{
\begin{tabular}{ccccccc}
\hline
\multicolumn{1}{c|}{} &
  \multicolumn{3}{c|}{\textbf{MGSM}} &
  \multicolumn{3}{c}{\textbf{MSVAMP}} \\
\multicolumn{1}{c|}{\multirow{-2}{*}{\textbf{LLaMA-2-7B as base model}}} &
  \textbf{LR.} &
  \textbf{HR.} &
  \multicolumn{1}{c|}{\textbf{Avg.}} &
  \textbf{LR.} &
  \textbf{HR.} &
  \textbf{Avg.} \\ \hline
\rowcolor[HTML]{F1F1F1} 
\multicolumn{7}{c}{\cellcolor[HTML]{F1F1F1}\textsc{baseline}} \\
\rowcolor[HTML]{F5EFF8} 
\multicolumn{7}{l}{\cellcolor[HTML]{F5EFF8}\textit{\textbf{Mono-SFT}}} \\
\rowcolor[HTML]{F5EFF8} 
\multicolumn{1}{c|}{\cellcolor[HTML]{F5EFF8}MAmmoTH\textsuperscript{\textdagger}} &
  3.40 &
  32.4 &
  \multicolumn{1}{c|}{\cellcolor[HTML]{F5EFF8}21.9} &
  6.57 &
  40.1 &
  30.1 \\
\rowcolor[HTML]{F5EFF8} 
\multicolumn{1}{c|}{\cellcolor[HTML]{F5EFF8}WizardMath\textsuperscript{\textdagger}} &
  4.00 &
  37.7 &
  \multicolumn{1}{c|}{\cellcolor[HTML]{F5EFF8}25.5} &
  15.7 &
  48.5 &
  38.7 \\
\rowcolor[HTML]{F5EFF8} 
\multicolumn{1}{c|}{\cellcolor[HTML]{F5EFF8}MetaMath\textsuperscript{\textdagger}} &
  4.60 &
  51.7 &
  \multicolumn{1}{c|}{\cellcolor[HTML]{F5EFF8}34.6} &
  15.2 &
  61.2 &
  47.4 \\
\rowcolor[HTML]{F5EFF8} 
\multicolumn{1}{c|}{\cellcolor[HTML]{F5EFF8}OpenMath2\textsuperscript{\textdaggerdbl}} &
  5.60 &
  60.2 &
  \multicolumn{1}{c|}{\cellcolor[HTML]{F5EFF8}40.4} &
  18.8 &
  70.5 &
  55.0 \\
\rowcolor[HTML]{E6EEF7} 
\multicolumn{7}{l}{\cellcolor[HTML]{E6EEF7}\textit{\textbf{Multi-SFT}}} \\
\rowcolor[HTML]{E6EEF7} 
\multicolumn{1}{c|}{\cellcolor[HTML]{E6EEF7}MathOctopus-Parallel\textsuperscript{\textdagger}} &
  28.0 &
  42.4 &
  \multicolumn{1}{c|}{\cellcolor[HTML]{E6EEF7}37.2} &
  33.6 &
  43.8 &
  40.8 \\
\rowcolor[HTML]{E6EEF7} 
\multicolumn{1}{c|}{\cellcolor[HTML]{E6EEF7}MathOctopus-MAPO-DPO\textsuperscript{\textdagger}} &
  30.6 &
  43.4 &
  \multicolumn{1}{c|}{\cellcolor[HTML]{E6EEF7}38.8} &
  52.5 &
  58.8 &
  56.9 \\
\rowcolor[HTML]{E6EEF7} 
\multicolumn{1}{c|}{\cellcolor[HTML]{E6EEF7}MetaMathOctopus-MAPO-DPO\textsuperscript{\textdagger}} &
  31.0 &
  55.6 &
  \multicolumn{1}{c|}{\cellcolor[HTML]{E6EEF7}46.7} &
  57.8 &
  67.4 &
  64.5 \\
\rowcolor[HTML]{E6EEF7} 
\multicolumn{1}{c|}{\cellcolor[HTML]{E6EEF7}QAlign-MetaMathQA\textsuperscript{\textdagger}} &
  26.3 &
  55.9 &
  \multicolumn{1}{c|}{\cellcolor[HTML]{E6EEF7}45.2} &
  48.4 &
  61.5 &
  57.6 \\
\rowcolor[HTML]{E1EFE1} 
\multicolumn{7}{l}{\cellcolor[HTML]{E1EFE1}\textit{\textbf{Leveraging External Tools or Models}}} \\
\rowcolor[HTML]{E1EFE1} 
\multicolumn{1}{c|}{\cellcolor[HTML]{E1EFE1}Translate-En\textsuperscript{\textdaggerdbl}~(MetaMath)} &
  39.6 &
  55.1 &
  \multicolumn{1}{c|}{\cellcolor[HTML]{E1EFE1}49.4} &
  47.4 &
  52.3 &
  50.9 \\
\rowcolor[HTML]{E1EFE1} 
\multicolumn{1}{c|}{\cellcolor[HTML]{E1EFE1}LangBridge\textsuperscript{\textdagger}~(MetaMath)} &
  38.4 &
  51.9 &
  \multicolumn{1}{c|}{\cellcolor[HTML]{E1EFE1}47.0} &
  43.9 &
  54.5 &
  51.4 \\
\rowcolor[HTML]{E1EFE1} 
\multicolumn{1}{c|}{\cellcolor[HTML]{E1EFE1}MindMerger-Soft\textsuperscript{\textdagger}~(MetaMath)} &
  53.1 &
  57.9 &
  \multicolumn{1}{c|}{\cellcolor[HTML]{E1EFE1}56.2} &
  52.7 &
  60.6 &
  58.2 \\
\rowcolor[HTML]{E1EFE1} 
\multicolumn{1}{c|}{\cellcolor[HTML]{E1EFE1}Translate-En\textsuperscript{\textdaggerdbl}~(OpenMath2)} &
  41.3 &
  63.8 &
  \multicolumn{1}{c|}{\cellcolor[HTML]{E1EFE1}55.6} &
  51.9 &
  62.2 &
  59.1 \\
\rowcolor[HTML]{E1EFE1} 
\multicolumn{1}{c|}{\cellcolor[HTML]{E1EFE1}LangBridge\textsuperscript{\textdaggerdbl}~(OpenMath2)} &
  42.6 &
  60.8 &
  \multicolumn{1}{c|}{\cellcolor[HTML]{E1EFE1}54.2} &
  47.9 &
  64.4 &
  59.5 \\
\rowcolor[HTML]{E1EFE1} 
\multicolumn{1}{c|}{\cellcolor[HTML]{E1EFE1}MindMerger-Soft\textsuperscript{\textdaggerdbl}~(OpenMath2)} &
  60.5 &
  \textbf{67.5} &
  \multicolumn{1}{c|}{\cellcolor[HTML]{E1EFE1}65.0} &
  63.4 &
  74.2 &
  70.9 \\ \hline
\rowcolor[HTML]{FBE5D6} 
\multicolumn{7}{c}{\cellcolor[HTML]{FBE5D6}\textsc{our methods}} \\
\rowcolor[HTML]{FBE5D6} 
\multicolumn{1}{c|}{\cellcolor[HTML]{FBE5D6}\textbf{LinguaLIFT (MetaMathQA)}} &
  55.4 &
  58.8 &
  \multicolumn{1}{c|}{\cellcolor[HTML]{FBE5D6}57.6} &
  56.1 &
  60.6 &
  59.3 \\
\rowcolor[HTML]{FBE5D6} 
\multicolumn{1}{c|}{\cellcolor[HTML]{FBE5D6}\textbf{LinguaLIFT\textsuperscript{$\Diamond$} (MetaMathQA)}} &
  51.2 &
  54.7 &
  \multicolumn{1}{c|}{\cellcolor[HTML]{FBE5D6}53.5} &
  51.5 &
  56.1 &
  54.7 \\
\rowcolor[HTML]{FBE5D6} 
\multicolumn{1}{c|}{\cellcolor[HTML]{FBE5D6}\textbf{LinguaLIFT (OpenMathInstruct-2)}} &
  \textbf{63.8} &
  66.5 &
  \multicolumn{1}{c|}{\cellcolor[HTML]{FBE5D6}\textbf{65.5}} &
  \textbf{67.2} &
  \textbf{74.3} &
  \textbf{72.2} \\ 
\rowcolor[HTML]{FBE5D6} 
\multicolumn{1}{c|}{\cellcolor[HTML]{FBE5D6}\textbf{LinguaLIFT\textsuperscript{$\Diamond$} (OpenMathInstruct-2)}} &
  59.9 &
  62.9 &
  \multicolumn{1}{c|}{\cellcolor[HTML]{FBE5D6}61.8} &
  63.5 &
  70.5 &
  68.4 \\ \hline
\end{tabular}
}
\caption{Results on MGSM and MSVAMP datasets. The usage of abbreviation, symbol and boldface are the same in Table~\ref{main-result-MMWP-48}.}
\label{tab:abbr-mgsm-msvamp}
\end{table}

Experimental results across four multilingual reasoning benchmarks (Tables~\ref{tab:abbr-mgsm-msvamp} and~\ref{tab:abbr-xnl-xcsqa}) demonstrate LinguaLIFT's robust generalization capabilities in out-of-domain scenarios. The framework achieves significant improvements over existing baselines in mathematical reasoning tasks (MGSM, MSVAMP) for low-resource languages while exhibiting strong adaptability to cross-domain commonsense reasoning challenges (XNLI, X-CSQA). By excelling in numerical computation and understanding semantics, LinguaLIFT demonstrates its task-agnostic generalization ability across various tasks, effectively transferring learned patterns between different reasoning paradigms and thus enhancing performance in low-resource language reasoning. Furthermore, the model enhances reasoning accuracy even in high-resource languages, underscoring its dual capacity to generalize and adapt across tasks and resource levels.

\begin{table}[!htbp]
\centering
\resizebox{\linewidth}{!}{
\begin{tabular}{ccccccc}
\hline
\multicolumn{1}{c|}{} &
  \multicolumn{3}{c|}{\textbf{XNLI}} &
  \multicolumn{3}{c}{\textbf{X-CSQA}} \\
\multicolumn{1}{c|}{\multirow{-2}{*}{\textbf{LLaMA-2-7B as base model}}} &
  \textbf{LR.} &
  \textbf{HR.} &
  \multicolumn{1}{c|}{\textbf{Avg.}} &
  \textbf{LR.} &
  \textbf{HR.} &
  \textbf{Avg.} \\ \hline
\rowcolor[HTML]{F1F1F1} 
\multicolumn{7}{c}{\cellcolor[HTML]{F1F1F1}\textsc{baselines}} \\
\rowcolor[HTML]{F5EFF8} 
\multicolumn{7}{l}{\cellcolor[HTML]{F5EFF8}\textit{\textbf{Mono-SFT(English-only Task Data)}}} \\
\rowcolor[HTML]{F5EFF8} 
\multicolumn{1}{c|}{\cellcolor[HTML]{F5EFF8}Mono-SFT\textsuperscript{\textasteriskcentered}} &
  58.7 &
  80.1 &
  \multicolumn{1}{c|}{\cellcolor[HTML]{F5EFF8}68.7} &
  28.6 &
  58.6 &
  51.3 \\
\rowcolor[HTML]{E6EEF7} 
\multicolumn{7}{l}{\cellcolor[HTML]{E6EEF7}\textit{\textbf{Multi-SFT(Multilingual Task Data with Query Translation)}}} \\
\rowcolor[HTML]{E6EEF7} 
\multicolumn{1}{c|}{\cellcolor[HTML]{E6EEF7}Multi-SFT\textsuperscript{\textasteriskcentered}} &
  63.6 &
  81.5 &
  \multicolumn{1}{c|}{\cellcolor[HTML]{E6EEF7}71.9} &
  29.4 &
  48.6 &
  43.8 \\
\rowcolor[HTML]{E6EEF7} 
\multicolumn{1}{c|}{\cellcolor[HTML]{E6EEF7}QAlign\textsuperscript{\textasteriskcentered}} &
  67.1 &
  80.9 &
  \multicolumn{1}{c|}{\cellcolor[HTML]{E6EEF7}73.5} &
  35.5 &
  57.9 &
  52.3 \\
\rowcolor[HTML]{E1EFE1} 
\multicolumn{7}{l}{\cellcolor[HTML]{E1EFE1}\textit{\textbf{Freezing LLM with External Tools or Models}}} \\
\rowcolor[HTML]{E1EFE1} 
\multicolumn{1}{c|}{\cellcolor[HTML]{E1EFE1}LangBridge\textsuperscript{\textasteriskcentered}} &
  73.9 &
  79.4 &
  \multicolumn{1}{c|}{\cellcolor[HTML]{E1EFE1}76.5} &
  30.9 &
  37.8 &
  36.1 \\
\rowcolor[HTML]{E1EFE1} 
\multicolumn{1}{c|}{\cellcolor[HTML]{E1EFE1}Translate-En\textsuperscript{\textasteriskcentered}} &
  71.1 &
  79.6 &
  \multicolumn{1}{c|}{\cellcolor[HTML]{E1EFE1}75.1} &
  42.7 &
  55.5 &
  52.3 \\
\rowcolor[HTML]{E1EFE1} 
\multicolumn{1}{c|}{\cellcolor[HTML]{E1EFE1}MindMerger-Soft\textsuperscript{\textasteriskcentered}} &
  74.4 &
  83.1 &
  \multicolumn{1}{c|}{\cellcolor[HTML]{E1EFE1}78.4} &
  47.9 &
  65.4 &
  61.0 \\ \hline
\rowcolor[HTML]{FBE5D6} 
\multicolumn{7}{c}{\cellcolor[HTML]{FBE5D6}\textsc{our methods}} \\
\rowcolor[HTML]{FBE5D6} 
\multicolumn{1}{c|}{\cellcolor[HTML]{FBE5D6}\textbf{LinguaLIFT}} &
  \textbf{77.6} &
  \textbf{83.3} &
  \multicolumn{1}{c|}{\cellcolor[HTML]{FBE5D6}\textbf{80.3}} &
  \textbf{49.6} &
  \textbf{65.5} &
  \textbf{61.5} \\ \hline
\end{tabular}
}
\caption{Results on XNLI and X-CSQA datasets. The usage of abbreviation is the same in Table~\ref{tab:abbr-mgsm-msvamp}. The asterisk symbol (\textasteriskcentered) indicates results are taken from the published results of \citet{huang2024mindmerger}.}
\label{tab:abbr-xnl-xcsqa}
\end{table}

Overall, LinguaLIFT established itself as a universally applicable and strong potential method for LLMs to advance low-resource language performance across various reasoning tasks. More experimental results are provided in Appendix \ref{apdx:detailsResults}.

\section{Analysis}

\subsection{Two-stage Training Ablation}
\begin{figure}[!htbp]
    \vspace{-5pt}
	\centering
	\includegraphics[width=.85\linewidth]{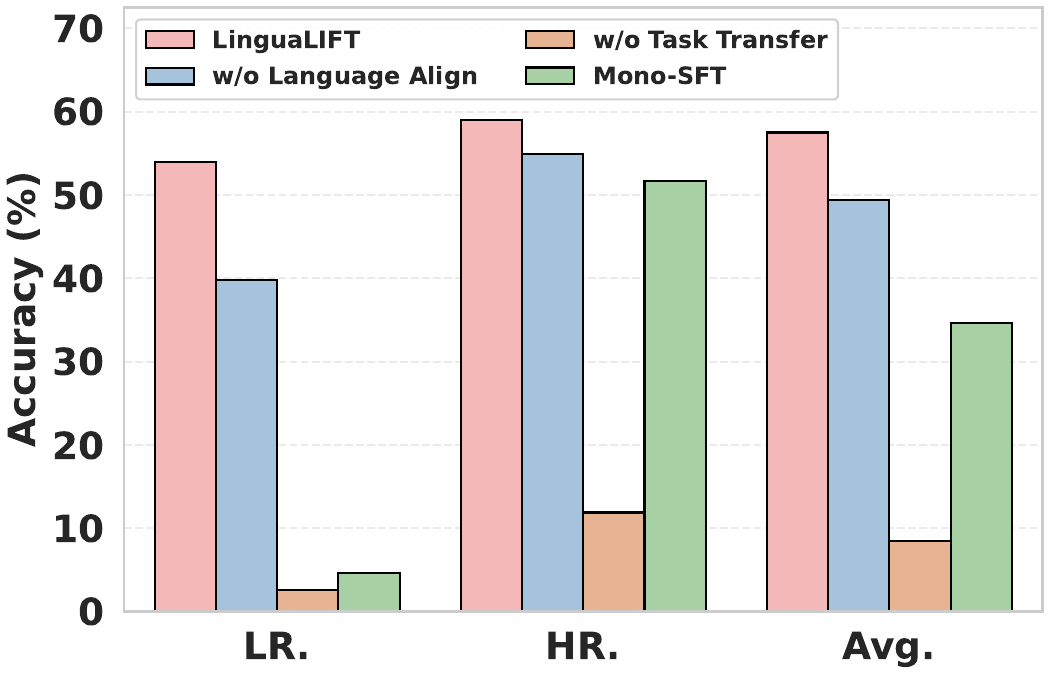}
	\caption{Ablation study of two-stage training on MGSM, showing average accuracy for low-resource (LR.), high-resource (HR.), and all languages (Avg.).}
	\label{stage_ablation}
\end{figure}

As shown in Figure~\ref{stage_ablation}, ablating the \textit{Language Align Stage} results in a significant performance decline for low-resource languages, empirically validating its role in cross-lingual knowledge transfer. Similarly, removing the \textit{Task Transfer Stage} degrades performance across all languages. These results underscore both stages' mutually dependent contributions to enhancing reasoning tasks.
Further details are provided in Appendix \ref{apdx:two-stage-ablation}.

\subsection{Trainable Modules Ablation}
\begin{figure}[!htbp]
    \vspace{-5pt}
	\centering
	\includegraphics[width=.85\linewidth]{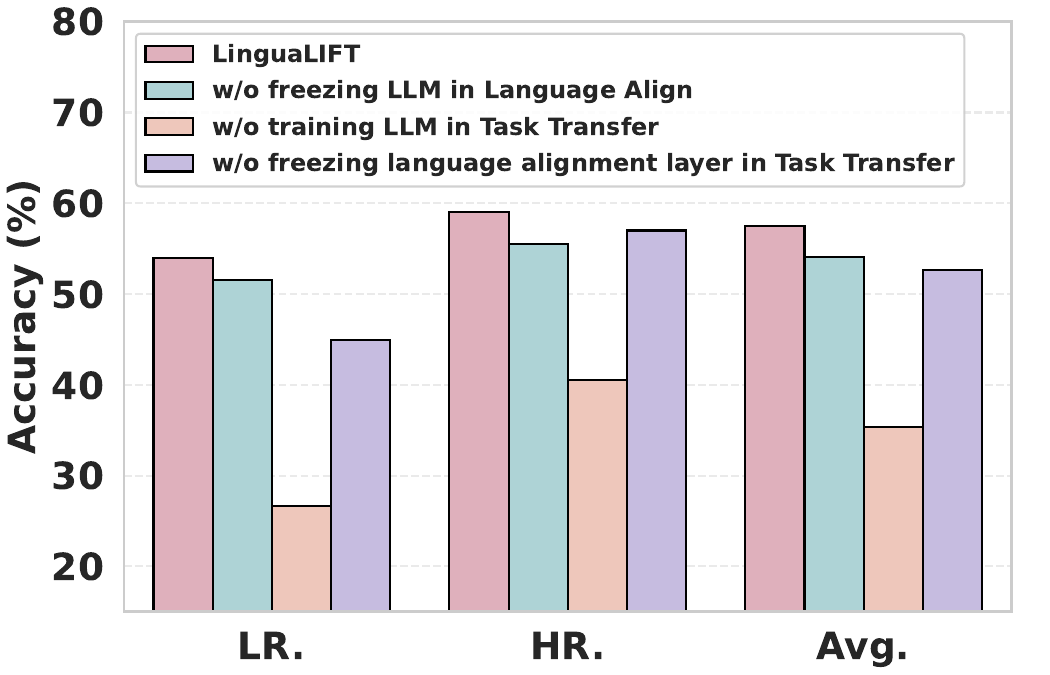}
	\caption{Trainable modules ablation on MGSM. The usage of abbreviation is the same in Figure~\ref{stage_ablation}.}
	\label{trainable_ablation}
\end{figure}

As shown in Figure~\ref{trainable_ablation}, training the LLM during the Language Align stage impairs high-resource language reasoning, likely due to catastrophic forgetting of acquired knowledge. Freezing the LLM during the Task Transfer stage similarly degrades performance across all languages. Notably, training the language alignment layer in this stage undermines low-resource reasoning, as it disrupts the pre-established multilingual alignment. These findings necessitate sequential training: first optimizing the alignment layer, followed by LLM fine-tuning.
Further details are available in Appendix \ref{apdx:trainable-ablation}.

\subsection{Multilingual Encoder Ablation}
\begin{figure}[!htbp]
    \vspace{-5pt}
    \centering
    \includegraphics[width=.85\linewidth]{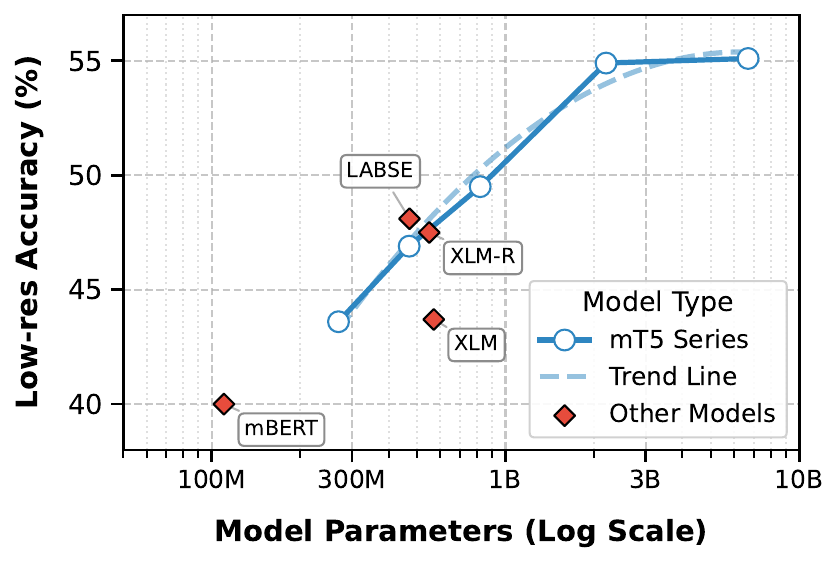}
    \caption{Accuracy (\%) of LinguaLIFT across encoders' scales and types on MGSM's low-resource reasoning.}
    \label{fig:mt5}
\end{figure}
Figure~\ref{fig:mt5} evaluates low-resource language reasoning on the MGSM test set using the mT5 series~\citep{xue-etal-2021-mt5} across various scales and four pre-trained multilingual encoders: mBERT~\citep{devlin-etal-2019-bert}, XLM~\citep{lample2019cross}, XLM-R~\citep{conneau-etal-2020-unsupervised}, and LaBSE~\citep{feng-etal-2022-language}. 
Performance improved significantly as mT5 scaled from 270M to 2.2B parameters but stagnated beyond this threshold, indicating a plateau at larger scales. Notably, LaBSE, leveraging parallel corpora for robust cross-lingual alignment, outperformed other encoders.
More detailed information is provided in Appendix~\ref{apdx:encoder}.

\subsection{Generalizes Well to Various LLMs}
\begin{table}[!htbp]
\centering
\resizebox{\columnwidth}{!}{%
\begin{tabular}{ccccccc}
\hline
\multicolumn{1}{l|}{\textbf{}}                        & \multicolumn{3}{c|}{\textbf{Mistral-7B}} & \multicolumn{3}{c}{\textbf{Llama-2-13B}} \\
\multicolumn{1}{c|}{\textbf{Methods}} &
  \textbf{LR.} &
  \textbf{HR.} &
  \multicolumn{1}{c|}{\textbf{Avg.}} &
  \textbf{LR.} &
  \textbf{HR.} &
  \textbf{Avg.} \\ \hline
\multicolumn{7}{c}{\textsc{baseline}}                                                                                                       \\
\rowcolor[HTML]{E6F9E6} 
\multicolumn{7}{l}{\cellcolor[HTML]{E6F9E6}\textit{\textbf{Mono-SFT(English-only Task Data)}}}                                              \\
\rowcolor[HTML]{E6F9E6} 
\multicolumn{1}{c|}{\cellcolor[HTML]{E6F9E6}MetaMath\textsuperscript{\textasteriskcentered}} &
  \cellcolor[HTML]{E6F9E6}30.0 &
  \cellcolor[HTML]{E6F9E6}66.5 &
  \multicolumn{1}{c|}{\cellcolor[HTML]{E6F9E6}55.6} &
  8.53 &
  59.7 &
  44.4 \\
\rowcolor[HTML]{FFFCC1} 
\multicolumn{7}{l}{\cellcolor[HTML]{FFFCC1}\textit{\textbf{Multi-SFT(Multilingual Task Data with Query Translation)}}}                      \\
\rowcolor[HTML]{FFFCC1} 
\multicolumn{1}{c|}{\cellcolor[HTML]{FFFCC1}MathOctopus-Parallel\textsuperscript{\textasteriskcentered}} &
  \cellcolor[HTML]{FFFCC1}48.1 &
  \cellcolor[HTML]{FFFCC1}51.1 &
  \multicolumn{1}{c|}{\cellcolor[HTML]{FFFCC1}50.2} &
  41.6 &
  47.7 &
  45.8 \\
\rowcolor[HTML]{FFFCC1} 
\multicolumn{1}{c|}{\cellcolor[HTML]{FFFCC1}QAlign-MetaMathQA\textsuperscript{\textasteriskcentered}} &
  \cellcolor[HTML]{FFFCC1}50.7 &
  \cellcolor[HTML]{FFFCC1}59.3 &
  \multicolumn{1}{c|}{\cellcolor[HTML]{FFFCC1}56.7} &
  44.7 &
  62.4 &
  57.1 \\
\rowcolor[HTML]{DCF2FC} 
\multicolumn{7}{l}{\cellcolor[HTML]{DCF2FC}\textit{\textbf{Freezing LLM with External Tools or Models}}}                                    \\
\rowcolor[HTML]{DCF2FC} 
\multicolumn{1}{c|}{\cellcolor[HTML]{DCF2FC}Translate-En\textsuperscript{\textasteriskcentered}} &
  \cellcolor[HTML]{DCF2FC}53.7 &
  \cellcolor[HTML]{DCF2FC}60.1 &
  \multicolumn{1}{c|}{\cellcolor[HTML]{DCF2FC}58.2} &
  44.4 &
  55.5 &
  52.2 \\
\rowcolor[HTML]{DCF2FC} 
\multicolumn{1}{c|}{\cellcolor[HTML]{DCF2FC}LangBridge\textsuperscript{\textasteriskcentered}} &
  \cellcolor[HTML]{DCF2FC}52.4 &
  \cellcolor[HTML]{DCF2FC}65.8 &
  \multicolumn{1}{c|}{\cellcolor[HTML]{DCF2FC}61.8} &
  41.3 &
  52.9 &
  49.5 \\
\rowcolor[HTML]{DCF2FC} 
\multicolumn{1}{c|}{\cellcolor[HTML]{DCF2FC}MindMerger-Soft\textsuperscript{\textasteriskcentered}} &
  \cellcolor[HTML]{DCF2FC}56.8 &
  \cellcolor[HTML]{DCF2FC}69.1 &
  \multicolumn{1}{c|}{\cellcolor[HTML]{DCF2FC}65.4} &
  57.1 &
  65.1 &
  62.7 \\ \hline
\multicolumn{7}{c}{\textsc{our methods}}                                                                                                    \\
\multicolumn{1}{c|}{\textbf{LinguaLIFT (MetaMathQA)}} & 57.7  & 68.8 & \multicolumn{1}{c|}{65.5} & 58.7         & 64.7        & 62.9        \\ \hline
\end{tabular}%
}
\caption{Tesults on MGSM across various LLMs. Abbreviation and symbol are the same in Table~\ref{main-result-MMWP-48}.}
\label{tab:abbr-various-llms}
\end{table}
We extend the evaluations in Section~\ref{sec:exp} to diverse LLMs and scales, including Llama-2-13B~\cite{touvron2023llama} and Mistral-7B~\citep{jiang2023mistral}. Table~\ref{tab:abbr-various-llms} demonstrates LinguaLIFT's broader applicability across LLM scales and architectures, with full results in Appendix~\ref{apdx:differentLLMs}.

\subsection{Analysis of Language Transferability}
\begin{figure}[!htbp]
\vspace{-5pt}
	\centering
    \includegraphics[width=.85\linewidth]{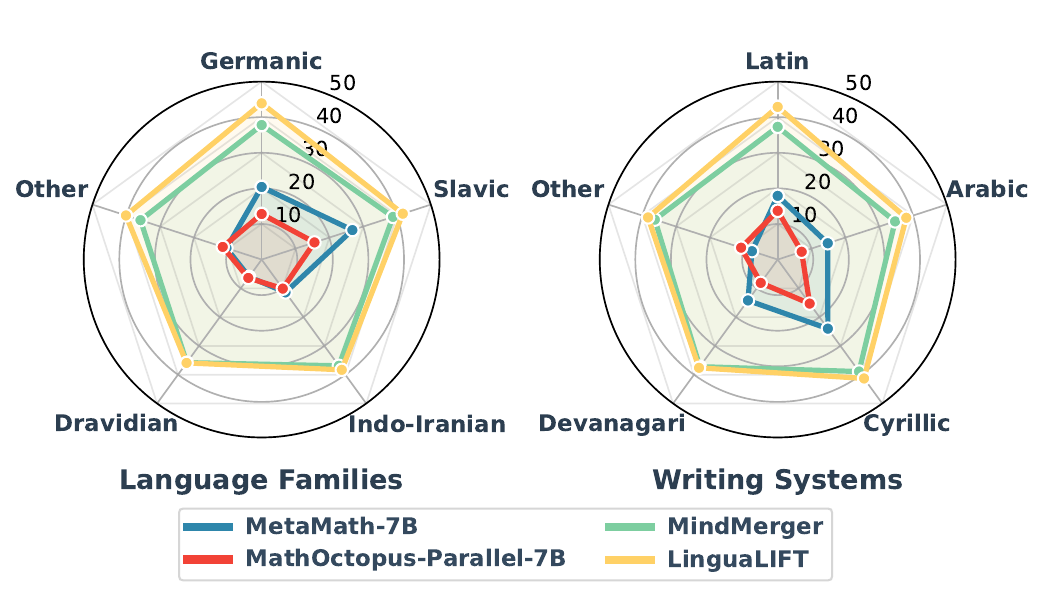}
    \caption{Reasoning accuracy on MMWP group by language families and writing systems.}
    \label{fig:language-family-writing-systems}
\end{figure}
As illustrated in Figure~\ref{fig:language-family-writing-systems}, LinguaLIFT outperformed all competitive baselines, achieving state-of-the-art transferability across language families and writing systems, particularly in languages sharing English's syntax or script (e.g., Germanic, Latin). 
These findings highlight LinguaLIFT as a robust cross-lingual transfer framework, leveraging shared linguistic features to enhance low-resource language reasoning tasks.
More details are provided in Appendix~\ref{apdx:language-family}.

\subsection{Analysis of Code-Switched Ratio}
\begin{figure}[!htbp]
    \centering
    \includegraphics[width=.85\linewidth]{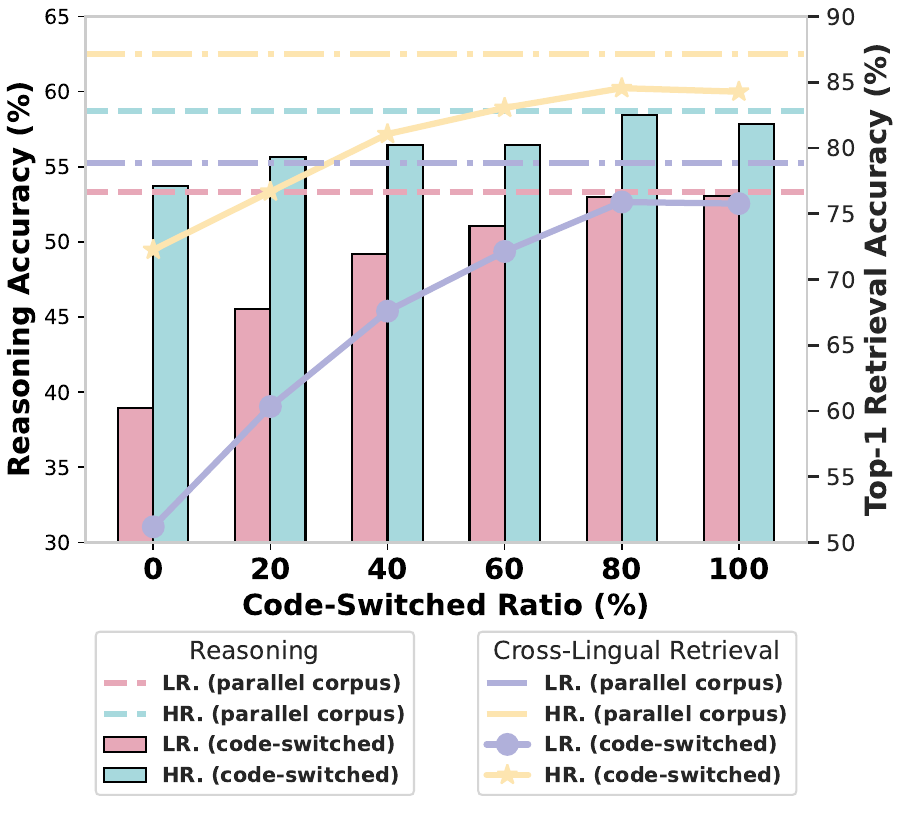}
    \caption{Ablation experiments on the impact of varying code-switch ratios on multilingual alignment and reasoning performance.}
    \label{fig:ablation_ratio}
\end{figure}

We assess language alignment and mathematical reasoning via top-1 retrieval accuracy on Tatoeba~\citep{artetxe-schwenk-2019-massively} and MGSM accuracy, respectively (Figure~\ref{fig:ablation_ratio}). Higher code-switching ratios enhance multilingual alignment, especially for low-resource languages, with proportional gains in reasoning accuracy.
Notably, at an 80\% ratio, the model matches the performance of parallel corpus-trained systems, indicating that word-level code-switching suffices for robust alignment. This demonstrates that explicit parallel corpora are unnecessary, as code-switching inherently promotes multilingual understanding and reasoning in low-resource scenarios.

\subsection{Analysis of Code-Switched Part-of-Speech}
\begin{figure}[!htbp]
    \centering
    \includegraphics[width=1\linewidth]{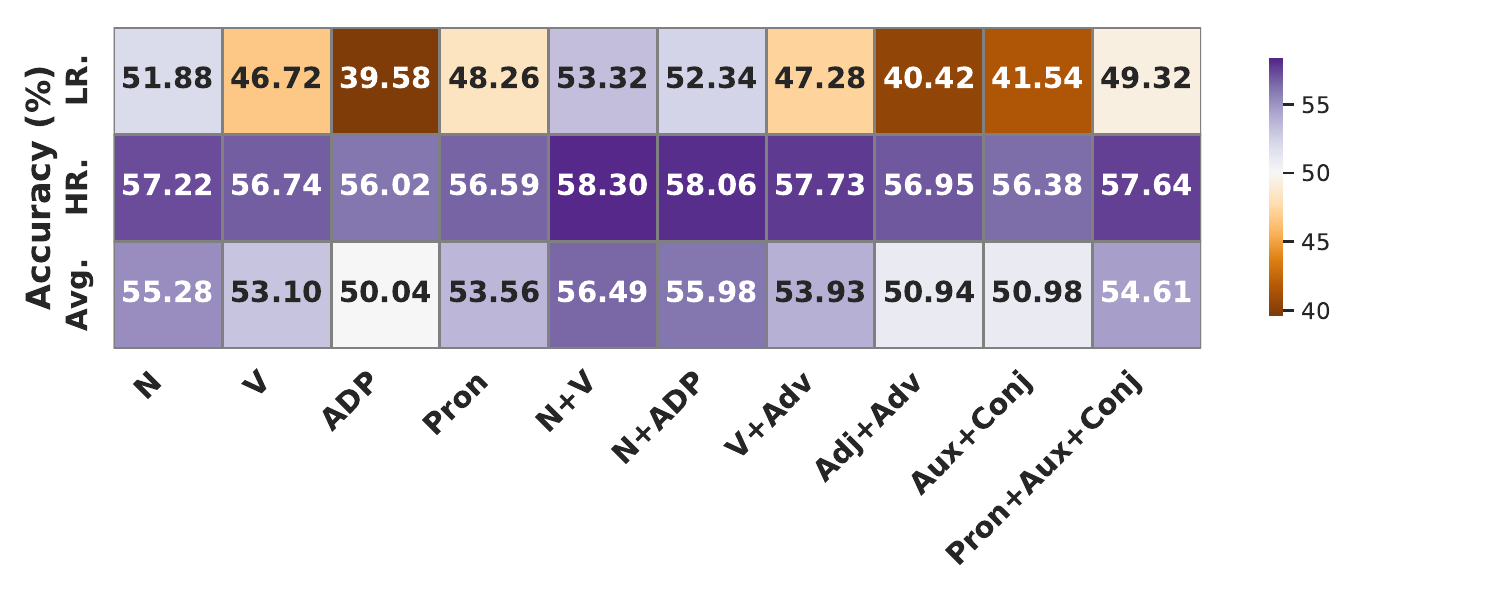}
    \caption{The reasoning performance on MGSM group by the combinations of POS: \textbf{(1) individual POS categories:}Verb (V), Adposition (ADP), Pronoun (Pron); \textbf{(2) syntactic function combinations:}Verb+Adverb (V+Adv), Adjective+Adverb (Adj+Adv), Pronoun+Auxiliary+Conjunction (Pron + Aux + Conj); and \textbf{(3) key syntactic structures:}Subject-Verb (N+V) and Prepositional Phrases (N+ADP).}
    \label{fig:ablation_pos}
\end{figure}

\begin{figure*}[!t]
    \centering
    \includegraphics[width=.85\linewidth]{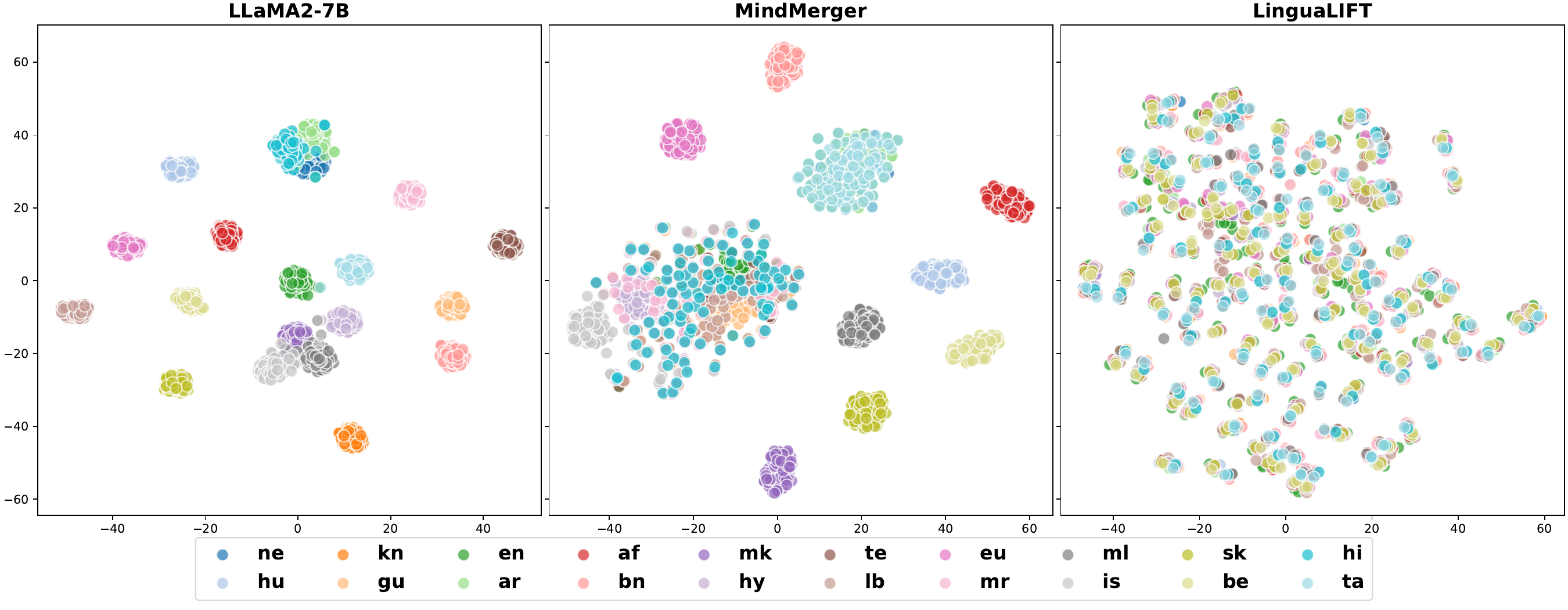}
    \caption{T-SNE visualization in the spaces of the Mono-SFT input embeddings, MindMerger mapping layer outputs, and LinguaLIFT language alignment layer outputs.}
    \label{fig:t-sne}
\end{figure*}
As illustrated in Figure~\ref{fig:ablation_pos}, noun substitutions significantly impair the model's reasoning capabilities among part-of-speech combinations. Syntactic structures involving subject-verb pairs and prepositional phrases demonstrate superior performance compared to other combinations, underscoring the importance of core arguments (subjects, verbs) and their syntactic relations (prepositions) in capturing essential relational information for cross-linguistic reasoning generalization. Conversely, adjective-adverb and auxiliary-conjunction combinations exhibit the lowest performance, indicating that modifiers play a less critical role in reasoning. Further details are available in Appendix~\ref{apdx:code-switch}.

\subsection{Incorporating Multilingual Data}
\begin{figure}[!htbp]
\vspace{-5pt}
    \centering
    \includegraphics[width=.9\linewidth]{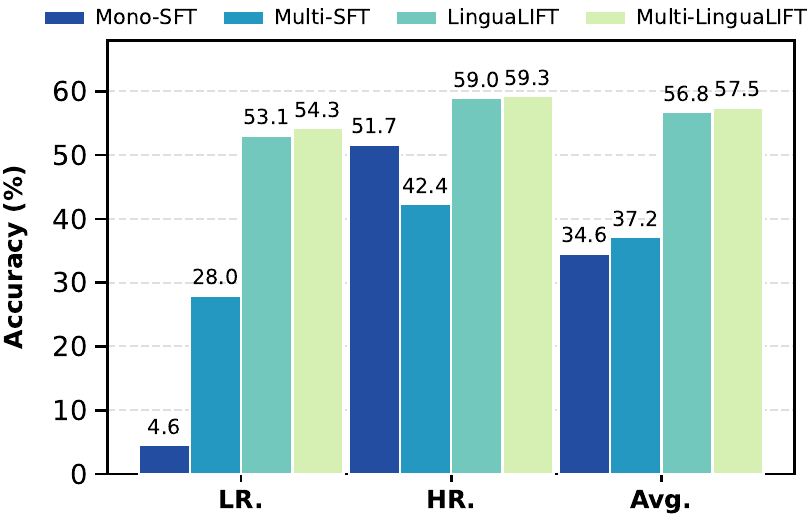}
    \caption{Accuracy (\%) on MGSM when LinguaLIFT leverages multilingual instruction and parallel data.}
    \label{fig:multi-lingualift}
\end{figure}
Beyond using code-switched translation data, LinguaLIFT can adapt to the training setting of previous works by integrating real multilingual data, further advancing reasoning performance. As shown in Figure~\ref{fig:multi-lingualift}, Multi-LinguaLIFT achieves a higher ceiling for low-resource language reasoning, with full experimental details in Appendix~\ref{apdx:incorporate-multilingual}.

\subsection{Alignment Representation Visualization}
For each low-resource language in the MMWP benchmark, we extracted 100 semantically equivalent texts from the Flores-101 dataset~\citep{goyal-etal-2022-flores}. We derived mean pooling representations using multiple methods and visualized them via T-SNE~\citep{JMLR:v9:vandermaaten08a}. 
As shown in Appendix~\ref{apdx:correlation}, cross-lingual alignment positively correlates with reasoning transfer: representations closer to English exhibit more substantial reasoning capability generalization.
Figure~\ref{fig:t-sne} reveals that the \textit{LLaMA2-7B}’s low-resource language representations are scattered and divergent from English, indicating the challenges in knowledge transfer. \textit{MindMerger} achieves partial alignment, with some low-resource language representations overlapping English representations; however, others remain isolated, reflecting its instability in diverse low-resource scenarios. In contrast, \textit{LinguaLIFT} tightly aligns low-resource languages with English, enabling effective reasoning transfer from English instruction data and yielding superior performance.

\subsection{Supplementary Experiments}
More supplementary experiments comprising: (1) quantitative correlations between multilingual alignment and reasoning performance (Appendix \ref{apdx:correlation}), (2) language alignment layer selection (Appendix \ref{apdx:la-layer}), and (3) zero-shot CoT examples in mathematical reasoning (Appendix \ref{apdx:case_study}). 

\section{Conclusion}
This paper proposes LinguaLIFT, a novel two-stage instruction tuning framework that enhances low-resource language reasoning without relying on parallel corpora or multilingual instruction data. 
Additionally, we introduce MMWP, a multilingual benchmark spanning 21 low-resource, 17 medium-resource, and 10 high-resource languages, to comprehensively evaluate multilingual mathematical reasoning tasks.
Experiments on the MMWP and other widely used benchmarks demonstrate its effectiveness in advancing low-resource language reasoning and further alleviating the performance gap between high-resource and low-resource language reasoning in LLMs.

\section*{Limitations}
While our experimental results demonstrate that leveraging code-switched tuning and cross-lingual transfer can help improve low-resource language reasoning, its implementation entails computational considerations requiring a moderately-size multilingual encoder for cross-lingual alignment, which introduces notable resource demands.  Although we employ LoRA for parameter efficiency, this adaptation may underperform full-parameter tuning in specific scenarios. Future work should prioritize developing lightweight adaptation strategies that preserve cross-lingual transfer capabilities while optimizing computational efficiency.

\bibliography{acl_latex}

\clearpage

\appendix
\section{Implementation Details}
\label{apdx:implementation-details}
\subsection{Collection of the Alignment Lexicons}
\label{apdx:align-vocab}
To better construct multilingual alignment lexicons for low-resource language reasoning, we leveraged the Unsupervised Bilingual Lexicon Induction (UBLI)~\citep{zhang-etal-2017-adversarial,lample2018word,dou-etal-2018-unsupervised,artetxe-etal-2019-bilingual,li-etal-2023-bilingual}, which has been proven effective in inducing word translation pairs by aligning independently trained word embeddings in two languages.

Initially, we tokenized the text using the Spacy library's tokenization function\footnote{\href{https://spacy.io/}{https://spacy.io/}}. The resultant word set comprised all words, barring specific named entities, numbers, and date tokens. This step is crucial in ensuring our focus on frequent and pertinent terms that are likely to hold significance in reasoning domains.

Following prior outstanding work  MUSE\footnote{\href{https://github.com/facebookresearch/MUSE}{https://github.com/facebookresearch/MUSE}}~\citep{lample2018word}, we construct multilingual lexicons using adversarial training to establish a linear mapping between source and target spaces without relying on cross-lingual supervision. This process involves training the model to align the word embeddings of the source and target languages in a shared semantic space. For each word in the source language, we identify the most relevant translations in the target language by projecting the source word embeddings into the target space and retrieving the top nearest neighbor words based on cosine similarity. These translated words form the bilingual lexicons, which are essential for enhancing multilingual understanding. Table \ref{tab:lexicon_demo} provides examples of word translations derived from the multilingual alignment lexicons we constructed, illustrating the effectiveness of this approach.

\begin{table}[!htbp]
    \centering
    \includegraphics[width=1\linewidth]{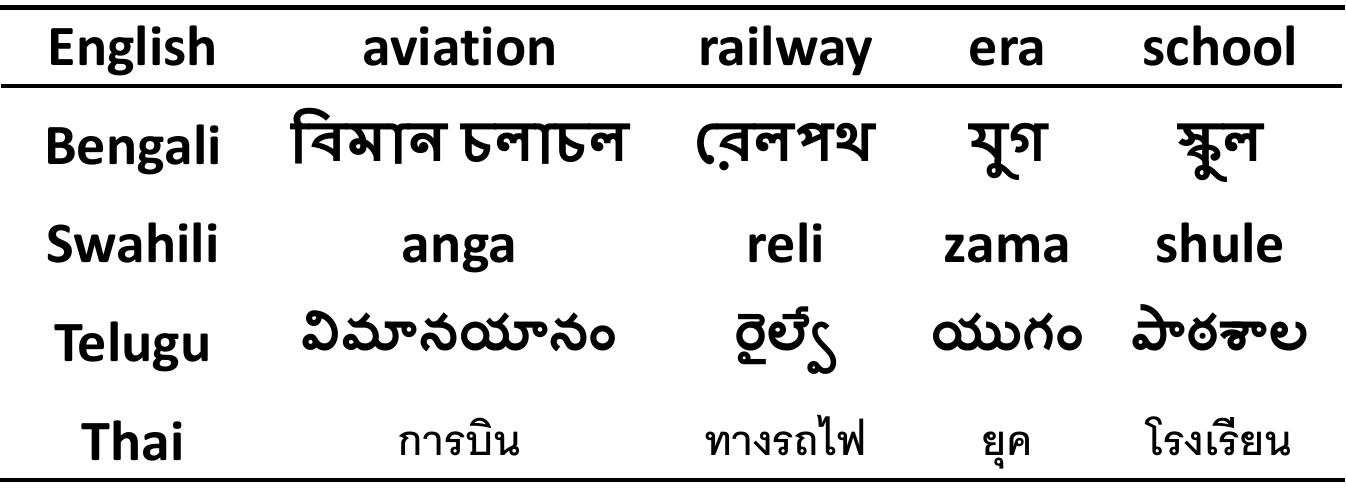}
    \caption{Word translations examples for English to several low-resource languages.}
    \label{tab:lexicon_demo}
\end{table}

The accuracy of the word mappings in our code-switch approach is a critical consideration, especially in the absence of parallel data. To demonstrate the feasibility and effectiveness of our unsupervised method, we address this important question from multiple perspectives:

$\bullet$ \textit{Methodology Validation}: The unsupervised word translation method we adopted is based on MUSE~\citep{lample2018word}, which has been thoroughly validated in previous research. As demonstrated in MUSE (Section 4.2), this approach achieves remarkable performance without any parallel data - up to 80\% accuracy for high-resource languages and around 50\% for low-resource languages when translating to English, which is competitive with supervised methods. Constructing a lexicon in such settings is inherently challenging, yet our method achieves remarkable performance, demonstrating the novelty and effectiveness of our approach.

$\bullet$ \textit{Further evaluation using GPT-4o}: To evaluate the accuracy of the lexicons constructed in Stage 1, we employed GPT-4o to evaluate the quality of the generated word mappings. The evaluation, using P@5 accuracy, yielded impressive results: (i) High-resource languages: 87.4\% accuracy (P@5) and (ii) Low-resource languages: 61.9\% accuracy (P@5).

These results further validate the quality of the unsupervised word mappings produced by our method, underscoring the strength of our approach even in low-resource settings.

This unsupervised approach allowed us to obtain diverse translations while preserving the general meaning of the words, which is crucial for capturing cross-lingual alignment. Moreover, by avoiding context-specific translations, we ensured that the generated translations had broad applicability across multiple tasks and languages. This strategy reduced the reliance on large-scale parallel corpora or pre-existing bilingual dictionaries and provided a robust and scalable solution for building multilingual lexicons in an unsupervised manner.

\subsection{Dataset Statistics}
\label{apdx:dataset}

In this study, we utilize a variety of datasets for training and evaluation purposes, as detailed in Table \ref{dataset-stats}. The datasets are divided into two main categories: \textbf{Training Datasets} and \textbf{Evaluation Datasets}.

The \textbf{Training Datasets} include data from different sources and cover a range of languages and sample sizes. These datasets are primarily used for training the models:

\begin{itemize}
    \item \textbf{MetaMathQA}~\citep{yu2024metamath}: A dataset with 395,000 samples in a single language, primarily used for mathematical question-answering tasks.
    \item \textbf{OpenMathInstruct-2}~\citep{toshniwal2024openmathinstruct-2}: This dataset contains 1,000,000 samples and is also in a single language, focusing on instructional tasks related to mathematics.
    \item \textbf{MGSM8KInstruct}~\citep{chen2023breaking}: A multilingual dataset involving 10 languages and 73,599 samples, designed for general-purpose instruction-based tasks.
    \item \textbf{Multi-NLI}~\citep{williams-etal-2018-broad}: With 392,702 samples, this dataset is in a single language and focuses on natural language inference tasks.
    \item \textbf{QASC}~\citep{Khot_Clark_Guerquin_Jansen_Sabharwal_2020}: A dataset with 8,134 samples used for question-answering tasks based on commonsense reasoning in a single language.
    \item \textbf{ARC}~\citep{clark2018think}: The ARC dataset contains 3,370 samples for answering elementary science questions in a single language.
    \item \textbf{OpenBookQA}~\citep{mihaylov-etal-2018-suit}: This dataset consists of 4,957 samples specifically designed for open-book question answering in a single language.
    \item \textbf{X-CSQA(En)}~\citep{lin-etal-2021-common}: A dataset of 8,888 samples for cross-lingual commonsense question answering in English.
\end{itemize}

\subsection*{Evaluation Datasets}
The \textbf{Evaluation Datasets} are used to assess the performance of the trained models across different tasks and languages:

\begin{itemize}
    \item \textbf{MMWP}: A multilingual dataset with 38,400 samples covering 48 languages. It is used for evaluating mathematical word problems.
    \item \textbf{MGSM}~\citep{shi2023language}: With 11 languages and 2,750 samples, this dataset is used for evaluating general language model performance across multiple languages.
    \item \textbf{MSVAMP}~\citep{chen2023breaking}: A dataset for evaluating model performance on visual question-answering tasks, containing 10,000 samples across 10 languages.
    \item \textbf{XNLI}~\citep{conneau-etal-2018-xnli}: This evaluation dataset covers 15 languages and contains 75,150 samples used for cross-lingual natural language inference tasks.
    \item \textbf{X-CSQA}~\citep{lin-etal-2021-common}: A dataset with 16,000 samples, spanning 16 languages, designed for cross-lingual commonsense question answering.
\end{itemize}

The \textbf{number of languages} (\#Lang) and \textbf{number of samples} (\#Samples) provided in each dataset enable a comprehensive evaluation of models across various linguistic and task-specific domains.

\begin{table}[!htbp]
\centering
\begin{tabular}{@{}ccc@{}}
\toprule
\textbf{Dataset}   & \textbf{\# Lang} & \textbf{\# Samples} \\ \midrule
\multicolumn{3}{c}{\textbf{Traing Datasets}}                \\
MetaMathQA         & 1                & 395,000             \\
OpenMathInstruct-2 & 1                & 1,000,000           \\
MGSM8KInstruct     & 10               & 73,599              \\
MetaMathQA         & 1                & 395,000             \\
Multi-NLI          & 1                & 392,702             \\
QASC               & 1                & 8,134               \\
ARC                & 1                & 3,370               \\
OpenBookQA         & 1                & 4,957               \\
X-CSQA(En)         & 1                & 8,888               \\ \midrule
\multicolumn{3}{c}{\textbf{Evaluation Datasets}}            \\
MMWP               & 48               & 38,928              \\
MGSM               & 11               & 2,750               \\
MSVAMP             & 10               & 10,000              \\
XNLI               & 15               & 75,150              \\
X-CSQA             & 16               & 16,000              \\ \bottomrule
\end{tabular}
\caption{Statistics of involved datasets. "\#Langs" denotes the number of languages covered by the dataset, and "\#Sample" refers to the total number of samples it contains. "Scenario" denotes the usage scenario of the specific dataset discussed in the methodology section.}
\label{dataset-stats}
\end{table}

\subsection{Instruction Tuning Prompts}
\label{apdx:prompts}

The prompt for code-switched tuning is adapted from \citet{zhang-etal-2023-multilingual}, where the source language, source sentence, and target language are replaced with the relevant translations.\\
The prompts for mathematical reasoning tasks, natural language inference, and commonsense question answering are modified from \citet{toshniwal2024openmathinstruct-2} and \citet{lu-etal-2024-llamax}, with the general instruction being replaced by the specific problems from the training data.

\begin{tcolorbox}[
    colframe=blue!50!black,    
    colback=blue!5!white,     
    boxsep=3pt,left=3pt,right=3pt,top=2pt,bottom=2pt,              
    title=\textbf{Prompt for Code-Switched Tuning}
]

Translate the following code-switched sentence from \texttt{\{source\_lang\}} to pure \texttt{\{target\_lang\}}:

\vspace{5pt}
\noindent
\texttt{\{source\_lang\}}: \texttt{\{source\_sentence\}}\\
\texttt{\{target\_lang\}}:
\end{tcolorbox}

\begin{tcolorbox}[
    colframe=blue!50!black,    
    colback=blue!5!white,      
    boxsep=3pt,left=3pt,right=3pt,top=2pt,bottom=2pt,              
    title=\textbf{Prompt for Mathematical Reasoning}
]

Solve the following math problem. Make sure to put the answer (and only the answer) inside \symbol{92}\symbol{92}boxed\{\}.

\vspace{5pt}
\noindent
\texttt{\{instruction\}}
\end{tcolorbox}

\begin{tcolorbox}[
    colframe=blue!50!black,    
    colback=blue!5!white,      
    boxsep=3pt,left=3pt,right=3pt,top=2pt,bottom=2pt,             
    title=\textbf{Prompt for XNLI}
]

I will give you a premise and a hypothesis. Choose the most appropriate relationship from the following options: Entailment, Neutral, Contradiction.

\vspace{5pt}
\noindent
\#\#\# Premise:

\vspace{5pt}
\noindent
\texttt{\{premise\}}

\vspace{5pt}
\noindent
\#\#\# Hypothesis:

\vspace{5pt}
\noindent
\texttt{\{hypo\}}

\vspace{5pt}
\noindent
\#\#\# Answer:

\end{tcolorbox}

\begin{tcolorbox}[
    colframe=blue!50!black,    
    colback=blue!5!white,      
    boxsep=3pt,left=3pt,right=3pt,top=2pt,bottom=2pt,               
    title=\textbf{Prompt for X-CSQA}
]

Question: 

\texttt{\{question\}}\\

Choices:
    
A. \texttt{\{choice\_A\}}

B. \texttt{\{choice\_B\}}
    
C. \texttt{\{choice\_C\}}

D. \texttt{\{choice\_D\}}

E. \texttt{\{choice\_E\}}\\

Answer:

\end{tcolorbox}

\subsection{Training Details}
\label{apdx:trainDetails}
We use mT5~\citep{xue-etal-2021-mt5} as the pre-trained multilingual encoder illustrated in Section \S \ref{sec:method}. We extend LlamaFactory\footnote{\href{https://github.com/hiyouga/LLaMA-Factory}{https://github.com/hiyouga/LLaMA-Factory}} \citep{zheng2024llamafactory} as the training codebase for our experiments. In the first stage, we train only the additional alignment modules (e.g., the language alignment layer and the boundary tokens) for 3 epochs, with a constant learning rate of 6e-4 and a batch size of 256. In the second stage, we fine-tune all parameters of the LLM for 3 epochs. The learning rate is set to 2e-5, with a warm-up ratio of 0.05 and a cosine learning rate scheduler. We also apply a weight decay of 1e-2 and use a batch size of 128. All experiments are conducted on eight NVIDIA A100 GPUs for a day.

\section{Details of Constructing the MMWP Benchmark}
\label{appendix:MMAWPS-Benchmark}

\subsection{Language Categorization List.}
\label{apdx:lang_list}
\begin{itemize}
    \item \textbf{Low-resource languages}: Afrikaans (\textsc{af}), Arabic (\textsc{ar}), Belarusian (\textsc{be}), Bengali (\textsc{bn}), Basque (\textsc{eu}), Gujarati (\textsc{gu}), Hausa (\textsc{ha}), Hindi (\textsc{hi}), Armenian (\textsc{hy}), Icelandic (\textsc{is}), Kannada (\textsc{kn}), Luxembourgish (\textsc{lb}), Macedonian (\textsc{mk}), Malayalam (\textsc{ml}), Marathi (\textsc{mr}), Nepali (\textsc{ne}), Slovak (\textsc{sk}), Swahili (\textsc{sw}), Tamil (\textsc{ta}), Telugu (\textsc{te}), Thai (\textsc{th}).

    \item \textbf{Medium-resource languages}: Bulgarian (\textsc{bg}), Catalan (\textsc{ca}), Czech (\textsc{cs}), Danish (\textsc{da}), Finnish (\textsc{fi}), Croatian (\textsc{hr}), Hungarian (\textsc{hu}), Indonesian (\textsc{id}), Korean (\textsc{ko}), Norwegian Bokmål (\textsc{nb}), Polish (\textsc{pl}), Portuguese (\textsc{pt}), Romanian (\textsc{ro}), Slovenian (\textsc{sl}), Serbian (\textsc{sr}), Ukrainian (\textsc{uk}), Vietnamese (\textsc{vi}).

    \item \textbf{High-resource languages}: German (\textsc{de}), English (\textsc{en}), Spanish (\textsc{es}), French (\textsc{fr}), Italian (\textsc{it}), Japanese (\textsc{ja}), Dutch (\textsc{nl}), Russian (\textsc{ru}), Swedish (\textsc{sv}), Chinese (\textsc{zh}).
\end{itemize}

\subsection{Selection of Target Languages and Categorization of Resource-level.}
\label{apdx:resource-level-categorization}
In our study, we employ a unique categorization method for selecting target languages, which deviates somewhat from the conventional definitions based on the abundance or scarcity of linguistic resources. This deviation is primarily due to our focus on LLMs. Rather than adhering to traditional classifications, we opt to categorize languages based on their language distribution in the pre-training corpus used for the LLMs. 

Guided by the LLaMA2 technical report~\citep{touvron2023llama}, we define low-resource languages as those that constitute less than 0.005\% of the available multilingual datasets. On the other hand, we categorize languages with a representation percentage between 0.005\% and 0.1\% as medium-resource languages. This categorization method allows us to incorporate a diverse set of languages into our study, thereby enabling a more effective and wide-ranging assessment of the multilingual reasoning capabilities of our model.

Furthermore, we believe that the comprehensive evaluation of multilingual tasks necessitates the inclusion of languages from various families and scripts. This diversity is crucial in understanding the robustness and versatility of our model, as it allows us to evaluate its performance and generalization capabilities across different linguistic contexts and resource levels. By incorporating languages from various families, scripts, and resource categories, we can ensure more comprehensive coverage in our multilingual settings evaluation. 

This unique approach to language selection and categorization provides a more nuanced understanding of language resource levels in LLMs while also ensuring a broader and more diverse evaluation of multilingual tasks for LLMs. It also underscores the importance of considering language families and scripts for the assessment of LLMs, thereby contributing to a more comprehensive and inclusive approach to language model development and evaluation.

\subsection{Evaluation and post-calibration process proves the reliability of the MMWP benchmark.} 
\label{apdx:quality-estimation}
To ensure the quality of the proposed MMWP benchmark, we conducted both human and automatic quality estimation evaluations and further post-calibration processes. 

For human evaluation, we employed a total of five annotators for all languages, each responsible for assessing the quality of the translated dataset and performing post-editing and calibration on the translations to ensure consistency and quality across the dataset. These annotators, native Chinese speakers with professional English proficiency and extensive translation and linguistic research experience, reviewed translations for grammatical correctness, fluency, and semantic accuracy. 
To ensure the reliability of the annotations, we hired two experts who are from Lan-bridge, a qualified institution for translation services\footnote{Requirements for translation services: \href{https://www.iso.org/standard/59149.html}{https://www.iso.org/standard/59149.html}.}, recognized by the ISO\footnote{International Organization for Standardization: \href{https://www.iso.org/home.html}{https://www.iso.org/home.html}.}, to serve as instructors to assess the reliability of the annotation. They evaluated the randomly selected translations and reannotated those with significant deviations. The evaluation process also included back-translation using GPT-4, which enabled annotators to compare the translated versions with the original texts. Any substantial deviations were addressed through post-editing to align the translations with the original intent and maintain native-like fluency. 
We carefully measured inter-annotator agreement (IAA) to ensure consistency and reliability. We found substantial agreement among annotators, with a Cohen Kappa score of 0.67 on randomly selected 50 instances, indicating substantial consistency in evaluating the translations. In cases where errors were identified, the annotators worked together to correct them, with an improved IAA of 0.80. The two experts’ determinations are consistent, with an IAA of 0.78.

Complementing the human evaluation, we conducted an automatic evaluation to assess the MMWP dataset's translation quality quantitatively. In this process, we adopted a back-translation strategy, translating the MMWP problems into the target languages and then back into English. Subsequently, we used widely-accepted automatic evaluation metrics such as BLEU \citep{papineni-etal-2002-bleu}, chrF \citep{popovic-2015-chrf}, and TER \citep{ter} to compare the back-translated problems with the original MMWP dataset. These metrics provided a quantitative measure of the overlap between the original and back-translated versions, thereby offering a quantifiable estimation of the translation fidelity. The results of this quality estimation, presented in Table \ref{tab:qe_benchmark}, attest to the reliable quality of the translations.

\begin{table}[!t]
\centering
\begin{tabular}{cccc}
\hline
        & BLEU$\uparrow$ & chrF$\uparrow$ & TER$\downarrow$ \\ \hline
Average & 78.71        & 86.08        & 1.09          \\ \hline
\end{tabular}
\caption{Automatic quality estimation of MMWP using back-translations, showing average results across all languages.}
\label{tab:qe_benchmark}
\end{table}

The results from both the human and automatic evaluations indicate that the dataset is properly constructed and adequately reflects the multilingual nature of the tasks. The human post-editing process ensured that the translated problems maintained their semantic integrity, while the automatic evaluation confirmed that the translations preserved the necessary linguistic structure for multilingual reasoning. Overall, the quality assurance measures we implemented to guarantee that the MMWP benchmark is reliable and effective in evaluating multilingual mathematical reasoning across diverse languages and resource levels.

The prompts for assisting the post-editing process and back-translating the low-resource language texts are presented below.

\begin{tcolorbox}[
    colframe=blue!50!black,    
    colback=blue!5!white,      
    boxsep=3pt,left=3pt,right=3pt,top=2pt,bottom=2pt,               
    title=\textbf{GPT-4 Prompt for Assisting Annotators Post-Editing}
]

You are provided with the following:

\begin{enumerate}
    \item A low-resource source language text.
    \item The back-translated English text (BT).
    \item The reference English translation (REF).
\end{enumerate}

Your task is to evaluate the quality of the back-translated English text (BT) based on three criteria: \textbf{grammatical correctness}, \textbf{fluency}, and \textbf{semantic accuracy}. Then, propose \textbf{three alternative revisions} to improve the BT. For each revision, explain why it was made and how it improves the translation.

\begin{enumerate}
    \item \textbf{Grammatical Correctness}: \\
    Does the back-translated text adhere to standard English grammar rules (e.g., subject-verb agreement, punctuation, tense consistency)?
    
    \item \textbf{Fluency}: \\
    Is the back-translated text natural and smooth? Does it sound like it was written by a native speaker?
    
    \item \textbf{Semantic Accuracy}: \\
    Does the back-translated text accurately reflect the meaning of the source language text? Are there any discrepancies in the interpretation of the source?
\end{enumerate}

\textbf{Provide three revision suggestions for improving the BT:}
\begin{enumerate}
    \item Each revision should aim to enhance either grammatical correctness, fluency, or semantic accuracy.
    \item For each suggestion, explain:
    \begin{itemize}
        \item Why the revision is necessary.
        \item Which aspect of translation quality (grammar, fluency, or accuracy) it improves.
    \end{itemize}
\end{enumerate}

\end{tcolorbox}

\begin{tcolorbox}[
    colframe=blue!50!black,    
    colback=blue!5!white,      
    boxsep=3pt,left=3pt,right=3pt,top=2pt,bottom=2pt,               
    title=\textbf{GPT-4 Prompt for Back-Translating}
]

You are a translation assistant. Directly translate the mathematical problems from \texttt{\{source\_lang\}} to English without additional explanations.

\texttt{\{source\_sentence\}}

\end{tcolorbox}

\section{Complete Experimental Results}
\label{apdx:detailsResults}

\subsection{Evaluation Results on MMWP}
The complete experimental results on MMWP are shown in Table~\ref{tab:MMWP-details-monosft}, \ref{tab:MMWP-details-multisft} and \ref{tab:MMWP-details-external}. These tables present Mono-SFT, Multi-SFT, and Leveraging External Tools or Models category comparison baselines.

\begin{table*}[!htbp]
\centering
\resizebox{.925\linewidth}{!}{
\begin{tabular}{@{}cccccc@{}}
\toprule
\textbf{Resource Level} &
  \textbf{Language} &
  \textbf{MAmmoTH-7B} &
  \textbf{WizardMath-7B} &
  \textbf{MetaMath-7B} &
  \multicolumn{1}{c}{\textbf{OpenMath2}} \\ \midrule
\multirow{21}{*}{\textbf{Low-Resource}}  & \multicolumn{1}{c|}{\textbf{af}} & 20.0 & 25.3 & 32.4 & 48.1 \\
                                         & \multicolumn{1}{c|}{\textbf{ar}} & 14.8 & 18.6 & 23.6 & 33.1 \\
                                         & \multicolumn{1}{c|}{\textbf{be}} & 7.6  & 12.5 & 17.9 & 30.3 \\
                                         & \multicolumn{1}{c|}{\textbf{bn}} & 4.8  & 7.9  & 9.9  & 15.7 \\
                                         & \multicolumn{1}{c|}{\textbf{eu}} & 2.2  & 4.2  & 6.3  & 6.5  \\
                                         & \multicolumn{1}{c|}{\textbf{gu}} & 2.5  & 3.6  & 4.8  & 4.0  \\
                                         & \multicolumn{1}{c|}{\textbf{ha}} & 2.8  & 4.4  & 5.9  & 6.4  \\
                                         & \multicolumn{1}{c|}{\textbf{hi}} & 7.6  & 17.0 & 24.5 & 34.4 \\
                                         & \multicolumn{1}{c|}{\textbf{hy}} & 3.2  & 3.7  & 5.8  & 6.2  \\
                                         & \multicolumn{1}{c|}{\textbf{is}} & 6.5  & 10.1 & 14.3 & 24.3 \\
                                         & \multicolumn{1}{c|}{\textbf{kn}} & 2.1  & 3.6  & 6.4  & 4.8  \\
                                         & \multicolumn{1}{c|}{\textbf{lb}} & 6.8  & 10.9 & 14.6 & 25.7 \\
                                         & \multicolumn{1}{c|}{\textbf{mk}} & 15.9 & 20.1 & 30.1 & 45.6 \\
                                         & \multicolumn{1}{c|}{\textbf{ml}} & 2.5  & 4.2  & 5.2  & 5.7  \\
                                         & \multicolumn{1}{c|}{\textbf{mr}} & 3.0  & 8.4  & 9.6  & 14.9 \\
                                         & \multicolumn{1}{c|}{\textbf{ne}} & 2.0  & 5.8  & 8.4  & 12.3 \\
                                         & \multicolumn{1}{c|}{\textbf{sk}} & 15.0 & 23.7 & 32.3 & 47.5 \\
                                         & \multicolumn{1}{c|}{\textbf{sw}} & 3.1  & 6.5  & 7.5  & 8.4  \\
                                         & \multicolumn{1}{c|}{\textbf{ta}} & 2.6  & 6.0  & 6.3  & 5.4  \\
                                         & \multicolumn{1}{c|}{\textbf{te}} & 2.5  & 4.9  & 6.2  & 4.4  \\
                                         & \multicolumn{1}{c|}{\textbf{th}} & 6.0  & 11.6 & 13.4 & 21.5 \\ \midrule
\multicolumn{2}{c|}{\textbf{Average}}                                       & 6.4  & 10.1 & 13.6 & 19.3 \\ \midrule
\multirow{17}{*}{\textbf{Medium-Resource}} &
  \multicolumn{1}{c|}{\textbf{bg}} &
  23.2 &
  29.1 &
  38.7 &
  59.7 \\
                                         & \multicolumn{1}{c|}{\textbf{ca}} & 23.9 & 31.4 & 38.8 & 59.1 \\
                                         & \multicolumn{1}{c|}{\textbf{cs}} & 19.7 & 28.7 & 37.5 & 56.7 \\
                                         & \multicolumn{1}{c|}{\textbf{da}} & 23.4 & 30.3 & 40.3 & 55.7 \\
                                         & \multicolumn{1}{c|}{\textbf{fi}} & 17.3 & 22.9 & 32.6 & 51.1 \\
                                         & \multicolumn{1}{c|}{\textbf{hr}} & 19.9 & 24.8 & 35.1 & 54.5 \\
                                         & \multicolumn{1}{c|}{\textbf{hu}} & 13.4 & 24.3 & 31.0 & 47.2 \\
                                         & \multicolumn{1}{c|}{\textbf{id}} & 22.2 & 30.0 & 38.4 & 57.6 \\
                                         & \multicolumn{1}{c|}{\textbf{ko}} & 14.4 & 25.9 & 35.0 & 47.4 \\
                                         & \multicolumn{1}{c|}{\textbf{nb}} & 23.2 & 29.4 & 37.9 & 56.5 \\
                                         & \multicolumn{1}{c|}{\textbf{pl}} & 21.2 & 28.0 & 37.9 & 57.3 \\
                                         & \multicolumn{1}{c|}{\textbf{pt}} & 25.5 & 36.3 & 42.9 & 65.0 \\
                                         & \multicolumn{1}{c|}{\textbf{ro}} & 20.4 & 27.1 & 39.5 & 56.1 \\
                                         & \multicolumn{1}{c|}{\textbf{sl}} & 18.3 & 24.5 & 33.1 & 51.7 \\
                                         & \multicolumn{1}{c|}{\textbf{sr}} & 18.4 & 26.8 & 36.1 & 54.0 \\
                                         & \multicolumn{1}{c|}{\textbf{uk}} & 20.7 & 27.6 & 40.3 & 58.8 \\
                                         & \multicolumn{1}{c|}{\textbf{vi}} & 21.8 & 28.4 & 37.0 & 54.8 \\ \midrule
\multicolumn{2}{c|}{\textbf{Average}}                                       & 20.4 & 28.0 & 37.2 & 55.5 \\ \midrule
\multirow{10}{*}{\textbf{High-resource}} & \multicolumn{1}{c|}{\textbf{de}} & 25.2 & 32.3 & 41.9 & 62.8 \\
                                         & \multicolumn{1}{c|}{\textbf{en}} & 36.6 & 37.6 & 42.7 & 75.0 \\
                                         & \multicolumn{1}{c|}{\textbf{es}} & 30.2 & 33.8 & 43.4 & 66.7 \\
                                         & \multicolumn{1}{c|}{\textbf{fr}} & 28.7 & 30.8 & 43.5 & 64.4 \\
                                         & \multicolumn{1}{c|}{\textbf{it}} & 28.1 & 34.7 & 42.9 & 63.1 \\
                                         & \multicolumn{1}{c|}{\textbf{ja}} & 19.7 & 28.6 & 40.8 & 56.4 \\
                                         & \multicolumn{1}{c|}{\textbf{nl}} & 23.9 & 31.2 & 41.2 & 58.9 \\
                                         & \multicolumn{1}{c|}{\textbf{ru}} & 23.9 & 30.3 & 41.6 & 60.8 \\
                                         & \multicolumn{1}{c|}{\textbf{sv}} & 24.2 & 30.3 & 42.5 & 59.7 \\
                                         & \multicolumn{1}{c|}{\textbf{zh}} & 20.7 & 32.1 & 37.9 & 59.2 \\ \midrule
\multicolumn{2}{c|}{\textbf{Average}}                                       & 26.1 & 32.2 & 41.8 & 62.7 \\ \bottomrule
\end{tabular}
}
\caption{Detailed results(Accuracy) of Mono-SFT baselines on the MMWP benchmark across 48 languages.
\label{tab:MMWP-details-monosft}
}
\end{table*}

\begin{table*}[!htbp]
\centering
\resizebox{\linewidth}{!}{
\begin{tabular}{@{}cccccc@{}}
\toprule
\textbf{Resource Level} &
  \textbf{Language} &
  \textbf{MathOctopus-Parallel-7B} &
  \textbf{MathOctopus-MAPO-DPO-7B} &
  \textbf{MetaMathOctopus-MAPO-DPO-7B} &
  \textbf{QAlign-MetaMathQA-7B} \\ \midrule
\multirow{21}{*}{\textbf{Low-Resource}}    & \multicolumn{1}{c|}{\textbf{af}} & 19.1 & 27.5 & 32.2 & 37.4 \\
                                           & \multicolumn{1}{c|}{\textbf{ar}} & 8.6  & 17.1 & 20.4 & 22.4 \\
                                           & \multicolumn{1}{c|}{\textbf{be}} & 14.6 & 23.2 & 21.8 & 18.1 \\
                                           & \multicolumn{1}{c|}{\textbf{bn}} & 20.1 & 28.1 & 28.4 & 26.6 \\
                                           & \multicolumn{1}{c|}{\textbf{eu}} & 5.6  & 11.2 & 9.4  & 7.5  \\
                                           & \multicolumn{1}{c|}{\textbf{gu}} & 6.2  & 10.9 & 5.3  & 6.3  \\
                                           & \multicolumn{1}{c|}{\textbf{ha}} & 5.6  & 10.6 & 8.8  & 6.3  \\
                                           & \multicolumn{1}{c|}{\textbf{hi}} & 11.1 & 24.3 & 14.7 & 22.2 \\
                                           & \multicolumn{1}{c|}{\textbf{hy}} & 6.0  & 8.4  & 6.0  & 6.5  \\
                                           & \multicolumn{1}{c|}{\textbf{is}} & 10.5 & 18.9 & 15.7 & 17.4 \\
                                           & \multicolumn{1}{c|}{\textbf{kn}} & 5.4  & 13.9 & 5.2  & 5.2  \\
                                           & \multicolumn{1}{c|}{\textbf{lb}} & 8.8  & 15.5 & 18.0 & 14.7 \\
                                           & \multicolumn{1}{c|}{\textbf{mk}} & 15.9 & 26.5 & 32.2 & 31.3 \\
                                           & \multicolumn{1}{c|}{\textbf{ml}} & 5.9  & 9.9  & 7.3  & 6.7  \\
                                           & \multicolumn{1}{c|}{\textbf{mr}} & 6.8  & 15.2 & 5.4  & 11.0 \\
                                           & \multicolumn{1}{c|}{\textbf{ne}} & 6.5  & 16.3 & 9.1  & 10.2 \\
                                           & \multicolumn{1}{c|}{\textbf{sk}} & 16.3 & 25.8 & 33.2 & 31.6 \\
                                           & \multicolumn{1}{c|}{\textbf{sw}} & 21.7 & 26.9 & 30.7 & 30.2 \\
                                           & \multicolumn{1}{c|}{\textbf{ta}} & 4.0  & 11.0 & 4.1  & 6.8  \\
                                           & \multicolumn{1}{c|}{\textbf{te}} & 10.2 & 17.9 & 4.0  & 5.1  \\
                                           & \multicolumn{1}{c|}{\textbf{th}} & 21.3 & 31.1 & 33.8 & 35.4 \\ \midrule
\multicolumn{2}{c|}{\textbf{Average}}                                         & 11.0 & 18.6 & 16.4 & 17.1 \\ \midrule
\multirow{17}{*}{\textbf{Medium-Resource}} & \multicolumn{1}{c|}{\textbf{bg}} & 20.6 & 30.1 & 36.4 & 39.2 \\
                                           & \multicolumn{1}{c|}{\textbf{ca}} & 22.8 & 32.9 & 41.8 & 41.2 \\
                                           & \multicolumn{1}{c|}{\textbf{cs}} & 18.9 & 28.0 & 36.3 & 39.1 \\
                                           & \multicolumn{1}{c|}{\textbf{da}} & 22.1 & 28.7 & 39.5 & 40.7 \\
                                           & \multicolumn{1}{c|}{\textbf{fi}} & 19.2 & 27.1 & 35.5 & 36.7 \\
                                           & \multicolumn{1}{c|}{\textbf{hr}} & 21.5 & 28.7 & 33.1 & 36.4 \\
                                           & \multicolumn{1}{c|}{\textbf{hu}} & 19.1 & 30.0 & 31.7 & 35.4 \\
                                           & \multicolumn{1}{c|}{\textbf{id}} & 20.8 & 29.1 & 36.5 & 39.5 \\
                                           & \multicolumn{1}{c|}{\textbf{ko}} & 18.0 & 29.4 & 33.4 & 38.5 \\
                                           & \multicolumn{1}{c|}{\textbf{nb}} & 22.6 & 28.5 & 40.3 & 40.3 \\
                                           & \multicolumn{1}{c|}{\textbf{pl}} & 21.0 & 31.3 & 38.1 & 39.2 \\
                                           & \multicolumn{1}{c|}{\textbf{pt}} & 24.3 & 32.4 & 44.1 & 43.8 \\
                                           & \multicolumn{1}{c|}{\textbf{ro}} & 19.6 & 30.7 & 36.4 & 38.8 \\
                                           & \multicolumn{1}{c|}{\textbf{sl}} & 19.2 & 30.2 & 32.6 & 36.3 \\
                                           & \multicolumn{1}{c|}{\textbf{sr}} & 18.5 & 29.4 & 33.3 & 37.2 \\
                                           & \multicolumn{1}{c|}{\textbf{uk}} & 24.8 & 31.2 & 37.7 & 41.1 \\
                                           & \multicolumn{1}{c|}{\textbf{vi}} & 20.4 & 28.9 & 36.6 & 39.3 \\ \midrule
\multicolumn{2}{c|}{\textbf{Average}}                                         & 20.8 & 29.8 & 36.7 & 39.0 \\ \midrule
\multirow{10}{*}{\textbf{High-resource}}   & \multicolumn{1}{c|}{\textbf{de}} & 27.7 & 33.5 & 41.3 & 45.1 \\
                                           & \multicolumn{1}{c|}{\textbf{en}} & 29.6 & 36.0 & 53.5 & 52.9 \\
                                           & \multicolumn{1}{c|}{\textbf{es}} & 25.8 & 33.7 & 46.4 & 46.0 \\
                                           & \multicolumn{1}{c|}{\textbf{fr}} & 24.8 & 32.4 & 44.1 & 43.4 \\
                                           & \multicolumn{1}{c|}{\textbf{it}} & 24.0 & 31.6 & 44.8 & 43.4 \\
                                           & \multicolumn{1}{c|}{\textbf{ja}} & 25.4 & 33.5 & 40.6 & 43.9 \\
                                           & \multicolumn{1}{c|}{\textbf{nl}} & 21.7 & 30.7 & 39.6 & 42.2 \\
                                           & \multicolumn{1}{c|}{\textbf{ru}} & 26.5 & 33.8 & 43.5 & 44.1 \\
                                           & \multicolumn{1}{c|}{\textbf{sv}} & 21.1 & 30.5 & 40.9 & 42.2 \\
                                           & \multicolumn{1}{c|}{\textbf{zh}} & 25.8 & 35.0 & 44.8 & 41.7 \\ \midrule
\multicolumn{2}{c|}{\textbf{Average}}                                         & 25.2 & 33.1 & 43.9 & 44.5 \\ \bottomrule
\end{tabular}
}
\caption{Detailed results of Multi-SFT on the MMWP benchmark across 48 languages.
\label{tab:MMWP-details-multisft}
}
\end{table*}

\begin{table*}[!htbp]
\centering
\resizebox{.925\linewidth}{!}{
\begin{tabular}{@{}cccccc@{}}
\toprule
\textbf{Resource Level} &
  \textbf{Language} &
  \textbf{Translate-En} &
  \textbf{Langbridge} &
  \textbf{MindMerger} &
  \textbf{AlignIFT-MetaMath} \\ \midrule
\multirow{21}{*}{\textbf{Low-Resource}} &
  \multicolumn{1}{c|}{\textbf{af}} &
  33.0 &
  40.7 &
  40.2 &
  46.4 \\
  & \multicolumn{1}{c|}{\textbf{ar}}  & 34.9 & 33.4 & 35.8 & 42.5 \\
  & \multicolumn{1}{c|}{\textbf{be}}  & 31.2 & 35.1 & 37.1 & 39.6 \\
  & \multicolumn{1}{c|}{\textbf{bn}}  & 27.7 & 31.9 & 37.6 & 39.5 \\
  & \multicolumn{1}{c|}{\textbf{eu}}  & 18.8 & 30.2 & 35.4 & 39.6 \\
  & \multicolumn{1}{c|}{\textbf{gu}}  & 21.6 & 31.8 & 35.8 & 41.0 \\
  & \multicolumn{1}{c|}{\textbf{ha}}  & 19.9 & 28.0 & 29.1 & 34.4 \\
  & \multicolumn{1}{c|}{\textbf{hi}}  & 27.4 & 35.3 & 39.4 & 41.7 \\
  & \multicolumn{1}{c|}{\textbf{hy}}  & 31.5 & 33.3 & 34.7 & 39.7 \\
  & \multicolumn{1}{c|}{\textbf{is}}  & 31.5 & 34.7 & 35.6 & 41.9 \\
  & \multicolumn{1}{c|}{\textbf{kn}}  & 26.7 & 29.7 & 33.8 & 37.4 \\
  & \multicolumn{1}{c|}{\textbf{lb}}  & 29.2 & 36.0 & 37.6 & 43.4 \\
  & \multicolumn{1}{c|}{\textbf{mk}}  & 35.9 & 40.8 & 40.7 & 43.9 \\
  & \multicolumn{1}{c|}{\textbf{ml}}  & 19.3 & 28.4 & 37.4 & 40.7 \\
  & \multicolumn{1}{c|}{\textbf{mr}}  & 25.6 & 30.5 & 34.3 & 37.6 \\
  & \multicolumn{1}{c|}{\textbf{ne}}  & 25.9 & 31.8 & 37.9 & 40.6 \\
  & \multicolumn{1}{c|}{\textbf{sk}}  & 29.2 & 36.4 & 38.6 & 42.5 \\
  & \multicolumn{1}{c|}{\textbf{sw}}  & 27.3 & 41.2 & 36.4 & 47.6 \\
  & \multicolumn{1}{c|}{\textbf{ta}}  & 24.4 & 27.1 & 35.0 & 40.5 \\
  & \multicolumn{1}{c|}{\textbf{te}}  & 28.5 & 29.7 & 37.1 & 41.9 \\
  & \multicolumn{1}{c|}{\textbf{th}}  & 30.3 & 34.9 & 38.7 & 41.7 \\ \midrule
\multicolumn{2}{c|}{\textbf{Average}} & 27.6 & 33.4 & 36.6 & 41.2 \\ \midrule
\multirow{17}{*}{\textbf{Medium-Resource}} &
  \multicolumn{1}{c|}{\textbf{bg}} &
  38.4 &
  40.3 &
  40.6 &
  46.5 \\
  & \multicolumn{1}{c|}{\textbf{ca}}  & 35.3 & 37.0 & 41.7 & 45.4 \\
  & \multicolumn{1}{c|}{\textbf{cs}}  & 37.7 & 37.4 & 44.1 & 47.4 \\
  & \multicolumn{1}{c|}{\textbf{da}}  & 39.4 & 39.5 & 42.1 & 46.0 \\
  & \multicolumn{1}{c|}{\textbf{fi}}  & 34.1 & 33.4 & 37.7 & 41.3 \\
  & \multicolumn{1}{c|}{\textbf{hr}}  & 31.0 & 33.7 & 37.4 & 40.2 \\
  & \multicolumn{1}{c|}{\textbf{hu}}  & 36.1 & 33.9 & 37.1 & 41.4 \\
  & \multicolumn{1}{c|}{\textbf{id}}  & 38.0 & 36.9 & 39.8 & 43.0 \\
  & \multicolumn{1}{c|}{\textbf{ko}}  & 30.7 & 33.1 & 39.5 & 42.6 \\
  & \multicolumn{1}{c|}{\textbf{nb}}  & 38.2 & 40.0 & 40.9 & 48.1 \\
  & \multicolumn{1}{c|}{\textbf{pl}}  & 38.8 & 38.5 & 40.6 & 43.8 \\
  & \multicolumn{1}{c|}{\textbf{pt}}  & 42.4 & 41.7 & 42.9 & 49.1 \\
  & \multicolumn{1}{c|}{\textbf{ro}}  & 38.0 & 37.5 & 42.3 & 43.7 \\
  & \multicolumn{1}{c|}{\textbf{sl}}  & 34.4 & 34.8 & 40.4 & 44.1 \\
  & \multicolumn{1}{c|}{\textbf{sr}}  & 38.1 & 36.7 & 42.4 & 44.8 \\
  & \multicolumn{1}{c|}{\textbf{uk}}  & 36.3 & 37.9 & 41.9 & 43.7 \\
  & \multicolumn{1}{c|}{\textbf{vi}}  & 31.6 & 33.9 & 39.8 & 42.0 \\ \midrule
\multicolumn{2}{c|}{\textbf{Average}} & 36.4 & 36.8 & 40.7 & 44.3 \\ \midrule
\multirow{10}{*}{\textbf{High-resource}} &
  \multicolumn{1}{c|}{\textbf{de}} &
  41.0 &
  40.7 &
  43.9 &
  44.6 \\
  & \multicolumn{1}{c|}{\textbf{en}}  & 42.5 & 47.1 & 47.8 & 49.7 \\
  & \multicolumn{1}{c|}{\textbf{es}}  & 40.6 & 39.6 & 43.9 & 49.0 \\
  & \multicolumn{1}{c|}{\textbf{fr}}  & 40.2 & 39.3 & 42.5 & 45.6 \\
  & \multicolumn{1}{c|}{\textbf{it}}  & 42.6 & 41.9 & 42.7 & 47.7 \\
  & \multicolumn{1}{c|}{\textbf{ja}}  & 39.6 & 37.0 & 40.2 & 40.4 \\
  & \multicolumn{1}{c|}{\textbf{nl}}  & 39.8 & 41.9 & 43.5 & 47.5 \\
  & \multicolumn{1}{c|}{\textbf{ru}}  & 38.9 & 40.4 & 42.4 & 46.5 \\
  & \multicolumn{1}{c|}{\textbf{sv}}  & 38.8 & 31.0 & 42.2 & 41.4 \\
  & \multicolumn{1}{c|}{\textbf{zh}}  & 41.5 & 35.1 & 43.3 & 42.1 \\ \midrule
\multicolumn{2}{c|}{\textbf{Average}} & 40.6 & 39.4 & 43.2 & 45.5 \\ \bottomrule
\end{tabular}
}
\caption{Detailed results(Accuracy) of Leveraging External Tools or Models baseline and LinguaLIFT on the MMWP benchmark across 48 languages.
}
\label{tab:MMWP-details-external}
\end{table*}

\subsection{Evaluation Results on MGSM and MMWP}
The complete experimental results on MGSM and MSVAMP are shown in Table~\ref{main-result-mgsm}, and \ref{main-result-msvamp}.

\begin{table*}[ht]
\centering
\resizebox{\linewidth}{!}{
\begin{tabular}{ccccccccccccccc}
\hline
\multicolumn{1}{l}{\textbf{LLaMA-2-7B as base model}} &
  \textbf{Bn} &
  \textbf{De} &
  \textbf{En} &
  \textbf{Es} &
  \textbf{Fr} &
  \textbf{Ja} &
  \textbf{Ru} &
  \textbf{Sw} &
  \textbf{Te} &
  \textbf{Th} &
  \textbf{Zh} &
  \textbf{LR.} &
  \textbf{HR.} &
  \textbf{Avg.} \\ \hline
\multicolumn{15}{c}{Baseline} \\
\rowcolor[HTML]{E6F9E6} 
\multicolumn{15}{l}{\cellcolor[HTML]{E6F9E6}\textit{Mono-SFT}} \\
\rowcolor[HTML]{E6F9E6} 
\cellcolor[HTML]{E6F9E6}MAmmoTH\textsuperscript{\textdagger} &
  3.20 &
  33.2 &
  47.2 &
  34.4 &
  32.4 &
  23.2 &
  30.4 &
  2.00 &
  1.60 &
  6.80 &
  26.0 &
  3.40 &
  32.4 &
  21.9 \\
\rowcolor[HTML]{E6F9E6} 
\cellcolor[HTML]{E6F9E6}WizardMath\textsuperscript{\textdagger} &
  4.40 &
  37.2 &
  51.6 &
  43.2 &
  39.2 &
  27.2 &
  36.8 &
  4.40 &
  1.60 &
  5.60 &
  28.8 &
  4.00 &
  37.7 &
  25.5 \\
\rowcolor[HTML]{E6F9E6} 
\cellcolor[HTML]{E6F9E6}MetaMath\textsuperscript{\textdagger} &
  5.60 &
  56.4 &
  68.0 &
  54.4 &
  56.0 &
  34.8 &
  52.8 &
  5.20 &
  2.00 &
  5.60 &
  39.6 &
  4.60 &
  51.7 &
  34.6 \\
\rowcolor[HTML]{E6F9E6} 
\cellcolor[HTML]{E6F9E6}OpenMath2\textsuperscript{\textdaggerdbl} &
  6.80 &
  65.6 &
  79.2 &
  66.0 &
  62.4 &
  42.4 &
  57.6 &
  6.00 &
  2.80 &
  6.80 &
  48.4 &
  5.60 &
  60.2 &
  40.4 \\
\rowcolor[HTML]{FFFCC1} 
\multicolumn{15}{l}{\cellcolor[HTML]{FFFCC1}\textit{Multi-SFT}} \\
\rowcolor[HTML]{FFFCC1} 
\cellcolor[HTML]{FFFCC1}MathOctopus-Parallel\textsuperscript{\textdagger} &
  31.2 &
  46.0 &
  51.6 &
  42.8 &
  43.6 &
  34.0 &
  40.0 &
  36.4 &
  10.8 &
  33.6 &
  38.8 &
  28.0 &
  42.4 &
  37.2 \\
\rowcolor[HTML]{FFFCC1} 
\cellcolor[HTML]{FFFCC1}MathOctopus-MAPO-DPO\textsuperscript{\textdagger} &
  33.2 &
  47.2 &
  46.8 &
  43.2 &
  39.6 &
  41.6 &
  40.8 &
  37.2 &
  13.6 &
  38.4 &
  44.8 &
  30.6 &
  43.4 &
  38.8 \\
\rowcolor[HTML]{FFFCC1} 
\cellcolor[HTML]{FFFCC1}MetaMathOctopus-MAPO-DPO\textsuperscript{\textdagger} &
  35.6 &
  52.4 &
  70.0 &
  58.0 &
  51.2 &
  46.4 &
  56.0 &
  42.4 &
  2.00 &
  44.0 &
  55.2 &
  31.0 &
  55.6 &
  46.7 \\
\rowcolor[HTML]{FFFCC1} 
\cellcolor[HTML]{FFFCC1}QAlign-MetaMathQA\textsuperscript{\textdagger} &
  29.6 &
  54.0 &
  68.4 &
  57.6 &
  59.2 &
  45.2 &
  58.4 &
  35.6 &
  2.40 &
  37.6 &
  48.8 &
  26.3 &
  55.9 &
  45.2 \\
\rowcolor[HTML]{DCF2FC} 
\multicolumn{15}{l}{\cellcolor[HTML]{DCF2FC}\textit{Leveraging External Tools or Models}} \\
\rowcolor[HTML]{DCF2FC} 
\cellcolor[HTML]{DCF2FC}Translate-En-MetaMath\textsuperscript{\textdaggerdbl} &
  49.0 &
  59.6 &
  65.6 &
  59.8 &
  56.2 &
  49.0 &
  48.4 &
  37.4 &
  34.6 &
  37.2 &
  47.0 &
  39.6 &
  55.1 &
  49.4 \\
\rowcolor[HTML]{DCF2FC} 
\cellcolor[HTML]{DCF2FC}LangBridge-MetaMath\textsuperscript{\textdagger} &
  41.2 &
  53.2 &
  62.4 &
  58.0 &
  51.6 &
  39.6 &
  55.2 &
  39.6 &
  28.0 &
  44.8 &
  43.2 &
  38.4 &
  51.9 &
  47.0 \\
\rowcolor[HTML]{DCF2FC} 
MindMerger-Soft-MetaMath\textsuperscript{\textdagger} &
  50.4 &
  59.6 &
  67.2 &
  58.4 &
  55.6 &
  50.0 &
  61.6 &
  55.2 &
  52.8 &
  54.2 &
  53.2 &
  53.1 &
  57.9 &
  56.2 \\
\rowcolor[HTML]{DCF2FC} 
Translate-En-OpenMath2\textsuperscript{\textdaggerdbl} &
  52.1 &
  69.8 &
  76.9 &
  71.6 &
  63.4 &
  57.4 &
  54.2 &
  38.4 &
  35.6 &
  39.2 &
  53.2 &
  41.3 &
  63.8 &
  55.6 \\
\rowcolor[HTML]{DCF2FC} 
LangBridge-OpenMath2\textsuperscript{\textdaggerdbl} &
  43.2 &
  63.0 &
  73.8 &
  69.8 &
  58.9 &
  47.6 &
  60.2 &
  41.0 &
  39.8 &
  46.2 &
  52.2 &
  42.6 &
  60.8 &
  54.2 \\
\rowcolor[HTML]{DCF2FC} 
MindMerger-Soft-OpenMath2\textsuperscript{\textdaggerdbl} &
  60.0 &
  69.4 &
  77.8 &
  74.2 &
  63.4 &
  57.8 &
  71.8 &
  61.2 &
  57.2 &
  63.6 &
  58.4 &
  60.5 &
  67.5 &
  65.0 \\ \hline
\multicolumn{15}{c}{Our Methods} \\
\textbf{LinguaLIFT-MetaMathQA} &
  54.4 &
  62.0 &
  64.8 &
  63.6 &
  56.8 &
  50.0 &
  60.4 &
  55.6 &
  54.0 &
  57.6 &
  54.0 &
  55.4 &
  58.8 &
  57.6 \\
\textbf{LinguaLIFT-MetaMathQA\textsuperscript{$\Diamond$}} &
  50.2 &
  57.2 &
  61.4 &
  60.2 &
  52.6 &
  45.8 &
  55.4 &
  51.2 &
  49.8 &
  53.6 &
  50.6 &
  51.2 &
  54.7 &
  53.5 \\
\textbf{LinguaLIFT-OpenMathInstruct-2} &
  63.0 &
  69.6 &
  76.0 &
  75.2 &
  62.8 &
  52.0 &
  71.6 &
  64.2 &
  61.2 &
  66.6 &
  58.0 &
  63.8 &
  66.5 &
  65.5 \\
\textbf{LinguaLIFT-OpenMathInstruct-2\textsuperscript{$\Diamond$}} &
  59.2 &
  65.0 &
  72.8 &
  71.6 &
  59.4 &
  48.6 &
  67.2 &
  60.0 &
  57.4 &
  62.8 &
  55.4 &
  59.9 &
  62.9 &
  61.8 \\ \hline
\end{tabular}
}
\caption{Experimental Results on the MGSM Dataset. "LR." "HR." and "Avg." represent the average performance for low-resource languages, high-resource languages, and all languages, respectively. Following prior work \citep{yoon-etal-2024-langbridge}, we classify Bn, Te, Th, and Sw as low-resource languages, while the remaining languages are categorized as high-resource. The dagger symbol (\textdagger) indicates results obtained using officially released models, the diamond symbol ($\Diamond$) signifies the LoRA implementation, and the double dagger symbol (\textdaggerdbl) denotes results from our local implementation.}
\label{main-result-mgsm}
\end{table*}

\begin{table*}[ht]
\centering
\resizebox{\linewidth}{!}{
\begin{tabular}{cccccccccccccc}
\hline
\multicolumn{1}{l}{\textbf{LLaMA-2-7B as base model}} &
  \textbf{Bn} &
  \textbf{De} &
  \textbf{En} &
  \textbf{Es} &
  \textbf{Fr} &
  \textbf{Ja} &
  \textbf{Ru} &
  \textbf{Sw} &
  \textbf{Th} &
  \textbf{Zh} &
  \textbf{LR.} &
  \textbf{HR.} &
  \textbf{Avg.} \\ \hline
\multicolumn{14}{c}{\textsc{baselines}} \\
\rowcolor[HTML]{E6F9E6} 
\multicolumn{14}{l}{\cellcolor[HTML]{E6F9E6}\textit{\textbf{Mono-SFT}}} \\
\rowcolor[HTML]{E6F9E6} 
\cellcolor[HTML]{E6F9E6}MAmmoTH\textsuperscript{\textdagger} &
  6.2 &
  44.1 &
  39.5 &
  45.4 &
  42.3 &
  34.1 &
  38.1 &
  5.1 &
  8.4 &
  37.5 &
  6.57 &
  40.1 &
  30.1 \\
\rowcolor[HTML]{E6F9E6} 
\cellcolor[HTML]{E6F9E6}WizardMath\textsuperscript{\textdagger} &
  16.4 &
  49.1 &
  56.1 &
  50.5 &
  50.8 &
  45.4 &
  44.8 &
  13.4 &
  17.2 &
  43.1 &
  15.7 &
  48.5 &
  38.7 \\
\rowcolor[HTML]{E6F9E6} 
\cellcolor[HTML]{E6F9E6}MetaMath\textsuperscript{\textdagger} &
  12.5 &
  63.5 &
  67.2 &
  64.7 &
  64.9 &
  54.2 &
  58.2 &
  16.7 &
  16.5 &
  55.5 &
  15.2 &
  61.2 &
  47.4 \\
\rowcolor[HTML]{E6F9E6} 
\cellcolor[HTML]{E6F9E6}OpenMath2\textsuperscript{\textdaggerdbl} &
  19.9 &
  72.7 &
  78.6 &
  72.5 &
  72.9 &
  64.6 &
  68.5 &
  13.1 &
  23.5 &
  63.8 &
  18.8 &
  70.5 &
  55.0 \\
\rowcolor[HTML]{FFFCC1} 
\multicolumn{14}{l}{\cellcolor[HTML]{FFFCC1}\textit{\textbf{Multi-SFT}}} \\
\rowcolor[HTML]{FFFCC1} 
\cellcolor[HTML]{FFFCC1}MathOctopus-Parallel\textsuperscript{\textdagger} &
  27.8 &
  43.8 &
  46.1 &
  45.4 &
  44.9 &
  41.4 &
  43.8 &
  38.1 &
  34.9 &
  41.4 &
  33.6 &
  43.8 &
  40.8 \\
\rowcolor[HTML]{FFFCC1} 
\cellcolor[HTML]{FFFCC1}MathOctopus-MAPO-DPO\textsuperscript{\textdagger} &
  48.3 &
  57.6 &
  59.0 &
  59.4 &
  60.0 &
  58.7 &
  58.0 &
  54.6 &
  54.7 &
  59.1 &
  52.5 &
  58.8 &
  56.9 \\
\rowcolor[HTML]{FFFCC1} 
\cellcolor[HTML]{FFFCC1}MetaMathOctopus-MAPO-DPO\textsuperscript{\textdagger} &
  50.3 &
  67.7 &
  71.5 &
  68.7 &
  67.5 &
  65.4 &
  64.7 &
  61.3 &
  61.7 &
  66.0 &
  57.8 &
  67.4 &
  64.5 \\
\rowcolor[HTML]{FFFCC1} 
\cellcolor[HTML]{FFFCC1}QAlign-MetaMathQA\textsuperscript{\textdagger} &
  41.9 &
  63.5 &
  65.5 &
  63.0 &
  62.8 &
  58.0 &
  61.5 &
  53.6 &
  49.7 &
  56.2 &
  48.4 &
  61.5 &
  57.6 \\
\rowcolor[HTML]{DCF2FC} 
\multicolumn{14}{l}{\cellcolor[HTML]{DCF2FC}\textit{\textbf{Leveraging External Tools or Models}}} \\
\rowcolor[HTML]{DCF2FC} 
\cellcolor[HTML]{DCF2FC}Translate-En-MetaMath\textsuperscript{\textdaggerdbl} &
  47.9 &
  43.9 &
  60.6 &
  51.4 &
  50.9 &
  50.4 &
  53.4 &
  43.1 &
  51.3 &
  55.8 &
  47.4 &
  52.3 &
  50.9 \\
\rowcolor[HTML]{DCF2FC} 
\cellcolor[HTML]{DCF2FC}LangBridge-MetaMath\textsuperscript{\textdagger} &
  39.6 &
  58.8 &
  60.1 &
  56.8 &
  57.9 &
  45.2 &
  53.6 &
  45.8 &
  46.3 &
  49.4 &
  43.9 &
  54.5 &
  51.4 \\
\rowcolor[HTML]{DCF2FC} 
\cellcolor[HTML]{DCF2FC}MindMerger-Soft-MetaMath\textsuperscript{\textdagger} &
  52.0 &
  61.1 &
  64.5 &
  62.9 &
  60.8 &
  59.0 &
  58.6 &
  54.0 &
  52.1 &
  57.3 &
  52.7 &
  60.6 &
  58.2 \\
\rowcolor[HTML]{DCF2FC} 
\cellcolor[HTML]{DCF2FC}Translate-En-OpenMath2\textsuperscript{\textdaggerdbl} &
  56.3 &
  54.1 &
  72.0 &
  59.5 &
  59.3 &
  60.9 &
  64.6 &
  40.2 &
  59.2 &
  64.9 &
  51.9 &
  62.2 &
  59.1 \\
\rowcolor[HTML]{DCF2FC} 
\cellcolor[HTML]{DCF2FC}LangBridge-OpenMath2\textsuperscript{\textdaggerdbl} &
  48.0 &
  68.6 &
  71.9 &
  64.8 &
  66.6 &
  56.5 &
  64.4 &
  42.3 &
  53.5 &
  58.3 &
  47.9 &
  64.4 &
  59.5 \\
\rowcolor[HTML]{DCF2FC} 
\cellcolor[HTML]{DCF2FC}MindMerger-Soft-OpenMath2\textsuperscript{\textdaggerdbl} &
  61.2 &
  77.2 &
  77.1 &
  73.6 &
  73.5 &
  74.1 &
  70.8 &
  64.8 &
  64.1 &
  72.8 &
  63.4 &
  74.2 &
  70.9 \\ \hline
\multicolumn{14}{c}{\textsc{our methods}} \\
\textbf{LinguaLIFT-MetaMathQA} &
  54.3 &
  62.8 &
  65.2 &
  59.9 &
  62.7 &
  57.9 &
  59.7 &
  56.8 &
  57.2 &
  56.0 &
  56.1 &
  60.6 &
  59.3 \\
\textbf{LinguaLIFT-MetaMathQA\textsuperscript{$\Diamond$}} &
  49.8 &
  57.6 &
  61.1 &
  55.4 &
  58.2 &
  53.5 &
  55.3 &
  52.1 &
  52.7 &
  51.4 &
  51.5 &
  56.1 &
  54.7 \\
\textbf{LinguaLIFT-OpenMathInstruct-2} &
  65.2 &
  76.4 &
  77.9 &
  74.2 &
  74.5 &
  73.5 &
  71.4 &
  68.6 &
  67.9 &
  72.2 &
  67.2 &
  74.3 &
  72.2 \\
\textbf{LinguaLIFT-OpenMathInstruct-2\textsuperscript{$\Diamond$}} &
  61.3 &
  72.1 &
  74.2 &
  70.5 &
  70.8 &
  69.7 &
  67.5 &
  64.9 &
  64.2 &
  68.5 &
  63.5 &
  70/5 &
  68.4 \\ \hline
\end{tabular}
}
\caption{Experimental Results on the MSVAMP Dataset. "LR." "HR." and "Avg." represent the average performance for low-resource languages, high-resource languages, and all languages, respectively. Following prior work \citep{yoon-etal-2024-langbridge}, we classify Bn, Th, and Sw as low-resource languages, while the remaining languages are categorized as high-resource. The dagger symbol (\textdagger) indicates results obtained using officially released models, while the double dagger symbol (\textdaggerdbl) denotes results from our local implementation.}
\label{main-result-msvamp}
\end{table*}

\subsection{Evaluation Results on XNLI and X-CSQA}
The complete experimental results on XNLI and X-CSQA are shown in Table~\ref{tab:xnli-results}, and \ref{tab:xcsqa-results}.

\begin{table*}[!ht]
\centering
\resizebox{\linewidth}{!}{
\begin{tabular}{ccccccccccccccccccc}
\hline
\multicolumn{1}{l}{\textbf{LLaMA-2-7B as base model}} &
  \textbf{Ar} &
  \textbf{Bg} &
  \textbf{Sw} &
  \textbf{Th} &
  \textbf{Tr} &
  \textbf{Ur} &
  \textbf{El} &
  \textbf{Hi} &
  \textbf{Zh} &
  \textbf{Ru} &
  \textbf{Vi} &
  \textbf{De} &
  \textbf{Fr} &
  \textbf{Es} &
  \textbf{En} &
  \textbf{LR.} &
  \textbf{HR.} &
  \textbf{Avg.} \\ \hline
\multicolumn{19}{c}{\textsc{baselines}} \\
\rowcolor[HTML]{E6F9E6} 
\multicolumn{19}{l}{\cellcolor[HTML]{E6F9E6}\textit{\textbf{Mono-SFT}}} \\
\rowcolor[HTML]{E6F9E6} 
\cellcolor[HTML]{E6F9E6}Mono-SFT\textsuperscript{\textasteriskcentered} &
  60.9 &
  76.7 &
  45.9 &
  55.4 &
  61.9 &
  49.2 &
  63.7 &
  55.7 &
  74.7 &
  77.6 &
  73.7 &
  80.6 &
  82.2 &
  82.2 &
  90.0 &
  58.7 &
  80.1 &
  68.7 \\
\rowcolor[HTML]{FFFCC1} 
\multicolumn{19}{l}{\cellcolor[HTML]{FFFCC1}\textit{\textbf{Multi-SFT}}} \\
\rowcolor[HTML]{FFFCC1} 
\cellcolor[HTML]{FFFCC1}Multi-SFT\textsuperscript{\textasteriskcentered} &
  61.7 &
  78.7 &
  56.3 &
  60.1 &
  65.6 &
  57.5 &
  67.0 &
  61.7 &
  79.1 &
  79.7 &
  73.7 &
  82.3 &
  82.9 &
  83.9 &
  88.8 &
  63.6 &
  81.5 &
  71.9 \\
\rowcolor[HTML]{FFFCC1} 
\cellcolor[HTML]{FFFCC1}QAlign\textsuperscript{\textasteriskcentered} &
  67.0 &
  79.4 &
  65.2 &
  65.2 &
  67.9 &
  62.2 &
  66.5 &
  63.3 &
  76.6 &
  79.2 &
  73.7 &
  80.9 &
  83.1 &
  83.8 &
  89.1 &
  67.1 &
  80.9 &
  73.5 \\
\rowcolor[HTML]{DCF2FC} 
\multicolumn{19}{l}{\cellcolor[HTML]{DCF2FC}\textit{\textbf{Leveraging External Tools or Models}}} \\
\rowcolor[HTML]{DCF2FC} 
\cellcolor[HTML]{DCF2FC}LangBridge\textsuperscript{\textasteriskcentered} &
  75.2 &
  79.6 &
  71.7 &
  72.4 &
  74.8 &
  66.9 &
  79.1 &
  71.1 &
  77.4 &
  77.4 &
  78.5 &
  78.8 &
  79.9 &
  80.5 &
  83.4 &
  73.9 &
  79.4 &
  76.5 \\
\rowcolor[HTML]{DCF2FC} 
\cellcolor[HTML]{DCF2FC}Translate-En\textsuperscript{\textasteriskcentered} &
  68.9 &
  80.8 &
  65.3 &
  69.5 &
  74.5 &
  61.6 &
  79.3 &
  68.7 &
  74.8 &
  76 &
  76.7 &
  80.6 &
  80.4 &
  81.4 &
  87.4 &
  71.1 &
  79.6 &
  75.1 \\
\rowcolor[HTML]{DCF2FC} 
\cellcolor[HTML]{DCF2FC}MindMerger-Soft\textsuperscript{\textasteriskcentered} &
  76.2 &
  82.4 &
  66.6 &
  71.8 &
  75.7 &
  69.4 &
  78.5 &
  74.7 &
  80.0 &
  80.7 &
  80.3 &
  83.5 &
  83.9 &
  84.4 &
  88.7 &
  74.4 &
  83.1 &
  78.4 \\ \hline
\multicolumn{19}{c}{\textsc{our methods}} \\
\textbf{LinguaLIFT} &
  78.4 &
  83.5 &
  75.1 &
  77.0 &
  77.1 &
  72.4 &
  82.2 &
  75.1 &
  80.2 &
  80.5 &
  80.8 &
  83.6 &
  84.0 &
  84.4 &
  89.5 &
  77.6 &
  83.3 &
  80.3 \\ \hline
\end{tabular}
}
\caption{Experimental Results on the XNLI Dataset. "LR." "HR." and "Avg." represent the average performance for low-resource languages, high-resource languages, and all languages, respectively. The asterisk symbol (\textasteriskcentered) indicates results obtained directly from the published results~\citep{huang2024mindmerger}.}
\label{tab:xnli-results}
\end{table*}

\begin{table*}[!ht]
\centering
\resizebox{\linewidth}{!}{
\begin{tabular}{cccccccccccccccccrrc}
\hline
\multicolumn{1}{l}{\textbf{LLaMA-2-7B as base model}} &
  \textbf{Ar} &
  \textbf{De} &
  \textbf{En} &
  \textbf{Es} &
  \textbf{Fr} &
  \textbf{Hi} &
  \textbf{It} &
  \textbf{Ja} &
  \textbf{Nl} &
  \textbf{Pl} &
  \textbf{Pt} &
  \textbf{Ru} &
  \textbf{Sw} &
  \textbf{Ur} &
  \textbf{Vi} &
  \multicolumn{1}{l}{\textbf{Zh}} &
  \multicolumn{1}{c}{\textbf{LR.}} &
  \multicolumn{1}{c}{\textbf{HR.}} &
  \textbf{Avg.} \\ \hline
\multicolumn{20}{c}{\textsc{baselines}} \\
\rowcolor[HTML]{E6F9E6} 
\multicolumn{19}{l}{\cellcolor[HTML]{E6F9E6}\textit{\textbf{Mono-SFT}}} &
  \multicolumn{1}{l}{\cellcolor[HTML]{E6F9E6}} \\
\rowcolor[HTML]{E6F9E6} 
\cellcolor[HTML]{E6F9E6}Mono-SFT\textsuperscript{\textasteriskcentered} &
  32.3 &
  61.2 &
  76.3 &
  64.0 &
  63.5 &
  32.9 &
  56.0 &
  49.1 &
  57.5 &
  50.6 &
  61.7 &
  56.0 &
  24.2 &
  25.1 &
  50.9 &
  56.5 &
  28.6 &
  58.6 &
  51.3 \\
\rowcolor[HTML]{FFFCC1} 
\multicolumn{19}{l}{\cellcolor[HTML]{FFFCC1}\textit{\textbf{Multi-SFT}}} &
  \multicolumn{1}{l}{\cellcolor[HTML]{FFFCC1}} \\
\rowcolor[HTML]{FFFCC1} 
\cellcolor[HTML]{FFFCC1}Multi-SFT\textsuperscript{\textasteriskcentered} &
  28.7 &
  49.1 &
  67.2 &
  54.3 &
  52.1 &
  32.0 &
  50.2 &
  38.7 &
  45.9 &
  45.5 &
  51.2 &
  46.5 &
  27.6 &
  29.2 &
  38.8 &
  43.8 &
  29.4 &
  48.6 &
  43.8 \\
\rowcolor[HTML]{FFFCC1} 
\cellcolor[HTML]{FFFCC1}QAlign\textsuperscript{\textasteriskcentered} &
  36.3 &
  58.8 &
  75.7 &
  63.1 &
  60.3 &
  37.8 &
  58.3 &
  49.2 &
  56.3 &
  51.3 &
  59.8 &
  56.3 &
  35.1 &
  32.6 &
  50.5 &
  54.8 &
  35.5 &
  57.9 &
  52.3 \\
\rowcolor[HTML]{DCF2FC} 
\multicolumn{19}{l}{\cellcolor[HTML]{DCF2FC}\textit{\textbf{Leveraging External Tools or Models}}} &
  \multicolumn{1}{l}{\cellcolor[HTML]{DCF2FC}} \\
\rowcolor[HTML]{DCF2FC} 
\cellcolor[HTML]{DCF2FC}LangBridge\textsuperscript{\textasteriskcentered} &
  30.6 &
  37.4 &
  44.4 &
  38.4 &
  38.2 &
  30.6 &
  39.1 &
  33.9 &
  38.4 &
  39.8 &
  36.3 &
  35.1 &
  31.8 &
  30.5 &
  33.3 &
  39.8 &
  30.9 &
  37.8 &
  36.1 \\
\rowcolor[HTML]{DCF2FC} 
\cellcolor[HTML]{DCF2FC}Translate-En\textsuperscript{\textasteriskcentered} &
  44.6 &
  57.3 &
  71.3 &
  55.5 &
  57.2 &
  48.4 &
  56.3 &
  47.1 &
  55.0 &
  53.3 &
  54.7 &
  54.4 &
  36.5 &
  41.3 &
  51.8 &
  51.5 &
  42.7 &
  55.5 &
  52.3 \\
\rowcolor[HTML]{DCF2FC} 
\cellcolor[HTML]{DCF2FC}MindMerger-Soft\textsuperscript{\textasteriskcentered} &
  51.4 &
  67.0 &
  78.1 &
  69.1 &
  68.1 &
  48.4 &
  66.8 &
  53.9 &
  63.8 &
  63.3 &
  67.1 &
  63.7 &
  45.5 &
  46.2 &
  60.6 &
  62.9 &
  47.9 &
  65.4 &
  61.0 \\ \hline
\multicolumn{20}{c}{\textsc{our methods}} \\
\textbf{LinguaLIFT} &
  \multicolumn{1}{r}{54.1} &
  \multicolumn{1}{r}{68.7} &
  \multicolumn{1}{r}{78.3} &
  \multicolumn{1}{r}{67.9} &
  \multicolumn{1}{r}{68.7} &
  \multicolumn{1}{r}{50.6} &
  \multicolumn{1}{r}{65.7} &
  \multicolumn{1}{r}{53.8} &
  \multicolumn{1}{r}{63.5} &
  \multicolumn{1}{r}{62.9} &
  \multicolumn{1}{r}{68.2} &
  \multicolumn{1}{r}{63.5} &
  \multicolumn{1}{r}{46.7} &
  \multicolumn{1}{r}{46.8} &
  \multicolumn{1}{r}{60.4} &
  \multicolumn{1}{r}{64.1} &
  \multicolumn{1}{c}{49.6} &
  \multicolumn{1}{c}{65.5} &
  61.5 \\ \hline
\end{tabular}
}
\caption{Experimental Results on the X-CSQA Dataset. "LR." "HR." and "Avg." represent the average performance for details of the low-resource, high-resource, and all languages, respectively. The asterisk symbol (\textasteriskcentered) indicates results obtained directly from the published results~\citep{huang2024mindmerger}.}
\label{tab:xcsqa-results}
\end{table*}

\section{Analysis Experiments Details}
This section outlines the experimental implementation for the various analysis experiments presented in the main text.

\subsection{Ablation of Two-Stage Training}
\label{apdx:two-stage-ablation}

To demonstrate the necessity of the proposed two-stage instruction tuning approach, we perform an ablation study on the distinct components of the two-stage process: the Language-Align stage and the Task-Transfer stage. This study is conducted on the mathematical reasoning datasets MGSM and MSVAMP to assess the impact of each stage’s removal on reasoning performance across all languages. The detailed results are presented in Tables~\ref{apdx-ablation-two-stage-mgsm} and \ref{apdx-ablation-two-stage-msvamp} for MGSM and MSVAMP, respectively, where ``w/o'' indicates the absence of the specific stage.

When the Language-Align stage was removed, we fine-tuned both the language-alignment layer and the LLM using only English reasoning data. This resulted in a noticeable decline in performance in low-resource languages, such as a 10.2\% performance drop in Bengali (Bn), 13.2\% performance drop in Thai (Th), 21.2\% performance drop in Telugu (Te), and 7.6\% performance drop in Swahili (Sw) on the MGSM dataset and a 14.4\% performance drop in Bengali (Bn), 15.7\% performance drop in Swahili (Sw) and 19.8\% performance drop in Thai (Th) on the MSVAMP dataset. This decline highlights the importance of using code-switched multilingual input as a warm-up strategy before directly applying English-only instruction data. By incorporating this stage, LinguaLIFT can better leverage the multilingual model’s representation space, which contains valuable information from low-resource languages.

When the Task-Transfer stage was removed, we fine-tuned the language alignment layer using only code-switched translation data without updating the LLM’s parameters. After ablating the Task-Transfer stage, the models couldn't complete reasoning tasks across all languages, resulting in an average accuracy of just 8.49\% on the MGSM dataset and 9.74\% on the MSVAMP dataset. This demonstrates that reasoning instruction tuning is critical for enabling the LLM to perform reasoning operations effectively.

These results underscore the complementary nature of the two stages: the Language-Align stage enhances the cross-lingual transferability within models, which allows the models to better transfer task-solving capabilities to low-resource languages in the Task-Transfer stage.

\begin{table*}[!htbp]
\centering
\resizebox{\linewidth}{!}{
\begin{tabular}{l|ccccccccccc|ccc}
\hline
\multicolumn{1}{c|}{\textbf{MGSM}} &
  \textbf{Bn} &
  \textbf{De} &
  \textbf{En} &
  \textbf{Es} &
  \textbf{Fr} &
  \textbf{Ja} &
  \textbf{Ru} &
  \textbf{Sw} &
  \textbf{Te} &
  \textbf{Th} &
  \textbf{Zh} &
  \textbf{LR.} &
  \textbf{HR.} &
  \textbf{Avg.} \\ \hline
\textbf{LinguaLIFT}                              & 52.8 & 62.8 & 64.8 & 65.2 & 55.6 & 48.4 & 61.2 & 53.2 & 50.4 & 56.0 & 54.8  & 54.0 & 59.0 & 57.5 \\
\multicolumn{1}{r|}{\textit{w/o language-align}} & 41.6 & 56.8 & 67.2 & 60.4 & 58.8 & 38.4 & 56.4 & 45.6 & 29.2 & 42.8 & 46.4  & 39.8 & 54.9 & 49.4 \\
\multicolumn{1}{r|}{\textit{w/o task-transfer}}  & 3.20 & 12.8 & 14.6 & 12.4 & 12.0 & 9.20 & 11.2 & 2.40 & 1.60 & 3.60 & 10.8  & 2.60 & 11.9 & 8.49 \\
Mono-SFT                                         & 6.00 & 58.0 & 64.8 & 55.6 & 55.6 & 37.6 & 52.4 & 4.40 & 0.00 & 4.80 & 40.80 & 3.80 & 52.1 & 34.6 \\ \hline
\end{tabular}
}
\caption{Two-stage Training Ablation Experimental Results on the MGSM Dataset. "LR." "HR." and "Avg." represent the average performance for low-resource languages, high-resource languages, and all languages, respectively. Following prior work \citep{huang2024mindmerger}, we classify Bn, Te, Th, and Sw as low-resource languages, while the remaining languages are categorized as high-resource.}
\label{apdx-ablation-two-stage-mgsm}
\end{table*}

\begin{table*}[!htbp]
\centering
\resizebox{\linewidth}{!}{
\begin{tabular}{l|cccccccccc|ccc}
\hline
\multicolumn{1}{c|}{\textbf{MSVAMP}} &
  \textbf{Bn} &
  \textbf{De} &
  \textbf{En} &
  \textbf{Es} &
  \textbf{Fr} &
  \textbf{Ja} &
  \textbf{Ru} &
  \textbf{Sw} &
  \textbf{Th} &
  \textbf{Zh} &
  \textbf{LR.} &
  \textbf{HR.} &
  \textbf{Avg.} \\ \hline
\textbf{LinguaLIFT}                              & 51.3 & 62.8 & 67.2 & 59.9 & 62.7 & 54.3 & 59.7 & 56.5 & 55.2 & 56.0 & 54.3 & 60.4 & 58.6 \\
\multicolumn{1}{r|}{\textit{w/o language-align}} & 36.9 & 58.7 & 63.6 & 58.8 & 62.1 & 53.5 & 55.3 & 40.8 & 35.4 & 53.1 & 37.7 & 57.9 & 51.8 \\
\multicolumn{1}{r|}{\textit{w/o task-transfer}}  & 3.8  & 10.4 & 16.5 & 11.7 & 15.3 & 7.7  & 12.0 & 2.2  & 4.1  & 13.7 & 3.37 & 12.5 & 9.74 \\
Mono-SFT &
  \multicolumn{1}{r}{15.2} &
  \multicolumn{1}{r}{59.5} &
  \multicolumn{1}{r}{64.9} &
  \multicolumn{1}{r}{62.9} &
  \multicolumn{1}{r}{61.7} &
  \multicolumn{1}{r}{52.4} &
  \multicolumn{1}{r}{58.3} &
  \multicolumn{1}{r}{15.2} &
  \multicolumn{1}{r}{18.3} &
  \multicolumn{1}{r|}{53.7} &
  \multicolumn{1}{r}{16.2} &
  \multicolumn{1}{r}{59.1} &
  \multicolumn{1}{r}{46.2} \\ \hline
\end{tabular}
}
\caption{Two-stage Training Ablation Experimental Results on the MSVAMP Dataset. "LR." "HR." and "Avg." represent the average performance for low-resource languages, high-resource languages, and all languages, respectively. Following prior work \citep{huang2024mindmerger}, we classify Bn, Te, Th, and Sw as low-resource languages, while the remaining languages are categorized as high-resource.}
\label{apdx-ablation-two-stage-msvamp}
\end{table*}

\subsection{Ablation of Trainable Modules}
To further demonstrate the essential design of the two-stage training approach, we conduct an ablation study on the trainable modules during each stage. This study is performed on the mathematical reasoning datasets MGSM and MSVAMP to assess the impact of training each module at the appropriate stage. The detailed results are presented in Tables~\ref{apdx-ablation-trainable-mgsm} and \ref{apdx-ablation-trainable-msvamp} for MGSM and MSVAMP, respectively, where ``w/o'' indicates the absence of a specific operation.

The results reveal that performance on high-resource languages suffers when the LLM is frozen during the Language-Align stage. We attribute this decline to catastrophic forgetting, where the initial training on low-resource languages interferes with the model’s ability to retain its high-resource language capabilities. Specifically, the average performance on high-resource languages decreases by 3.5\% on the MGSM dataset and 2.6\% on the MSVAMP dataset.

When the LLM training is ablated in the Task-Transfer stage, performance significantly deteriorates across all languages. This indicates that relying solely on the language alignment layer is insufficient for learning high-level reasoning tasks and effectively transferring knowledge to low-resource languages. The average performance across all languages drops to 35.4\% on the MGSM dataset and 37.8\% on the MSVAMP dataset.

Furthermore, freezing the language alignment layer during the Task-Transfer stage results in a decline in performance on low-resource languages. This suggests that the alignment learned in the language alignment layer is disrupted during continuous training, as the average performance on low-resource languages drops by 9.1\% on the MGSM dataset and 5.3\% on the MSVAMP dataset.

These findings emphasize the benefit of training the language alignment layer first, followed by the LLM training, to improve performance on both low-resource and high-resource languages.

\label{apdx:trainable-ablation}
\begin{table*}[!htbp]
\centering
\resizebox{\linewidth}{!}{
\begin{tabular}{r|ccccccccccc|ccc}
\hline
\multicolumn{1}{c|}{\textbf{MGSM}} &
  \textbf{Bn} &
  \textbf{De} &
  \textbf{En} &
  \textbf{Es} &
  \textbf{Fr} &
  \textbf{Ja} &
  \textbf{Ru} &
  \textbf{Sw} &
  \textbf{Te} &
  \textbf{Th} &
  \textbf{Zh} &
  \textbf{LR.} &
  \textbf{HR.} &
  \textbf{Avg.} \\ \hline
\multicolumn{1}{l|}{\textbf{LinguaLIFT}}          & 52.8 & 62.8 & 64.8 & 65.2 & 55.6 & 48.4 & 61.2 & 53.2 & 50.4 & 56.0 & 54.8 & 54.0 & 59.0 & 57.5 \\
\textit{w/o freezing LLM in Language-Align}       & 50.6 & 59.3 & 62.0 & 60.1 & 52.7 & 45.0 & 58.2 & 52.6 & 48.2 & 54.5 & 51.5 & 51.5 & 55.5 & 54.1 \\
\textit{w/o training LLM in Task-Transfer}        & 26.4 & 42.4 & 47.6 & 45.6 & 40.4 & 30.4 & 41.2 & 26.8 & 24.8 & 28.4 & 35.6 & 26.6 & 40.5 & 35.4 \\
\textit{w/o freezing Language Alignment Layer in Task-Transfer} & 46.8 & 60.2 & 66.4 & 59.2 & 57.8 & 46.8 & 58.8 & 45.2 & 39.6 & 48.0 & 49.6 & 44.9 & 57.0 & 52.6 \\ \hline
\end{tabular}
}
\caption{Trainable Modules Ablation Experimental Results on the MGSM Dataset. "LR." "HR." and "Avg." represent the average performance for low-resource languages, high-resource languages, and all languages, respectively. Following prior work \citep{huang2024mindmerger}, we classify Bn, Te, Th, and Sw as low-resource languages, while the remaining languages are categorized as high-resource.}
\label{apdx-ablation-trainable-mgsm}
\end{table*}

\begin{table*}[!htbp]
\centering
\resizebox{\linewidth}{!}{
\begin{tabular}{r|cccccccccc|ccc}
\hline
\multicolumn{1}{c|}{\textbf{MSVAMP}} &
  \textbf{Bn} &
  \textbf{De} &
  \textbf{En} &
  \textbf{Es} &
  \textbf{Fr} &
  \textbf{Ja} &
  \textbf{Ru} &
  \textbf{Sw} &
  \textbf{Th} &
  \textbf{Zh} &
  \textbf{LR.} &
  \textbf{HR.} &
  \textbf{Avg.} \\ \hline
\multicolumn{1}{l|}{\textbf{LinguaLIFT}}          & 51.3 & 62.8 & 67.2 & 59.9 & 62.7 & 54.3 & 59.7 & 56.5 & 55.2 & 56.0 & 54.3 & 60.4 & 58.6 \\
\textit{w/o freezing LLM in Language-Align}        & 50.7 & 59.9 & 65.4 & 57.4 & 55.6 & 50.1 & 57.5 & 55.6 & 54.9 & 53.0 & 53.7 & 57.0 & 56.0 \\
\textit{w/o training LLM in Task-Transfer}        & 29.3 & 49.7 & 51.3 & 45.6 & 47.4 & 28.8 & 35.8 & 27.1 & 24.8 & 38.3 & 27.1 & 42.4 & 37.8 \\
\textit{w/o freezing Language Alignment Layer in Task-Transfer} & 42.4 & 60.4 & 65.6 & 60.2 & 60.8 & 53.3 & 58.5 & 53.1 & 51.6 & 50.8 & 49.0 & 58.8 & 55.9 \\ \hline
\end{tabular}
}
\caption{Trainable Modules Ablation Experimental Results on the MSVAMP Dataset. "LR." "HR." and "Avg." represent the average performance for low-resource languages, high-resource languages, and all languages, respectively. Following prior work \citep{huang2024mindmerger}, we classify Bn, Te, Th, and Sw as low-resource languages, while the remaining languages are categorized as high-resource.}
\label{apdx-ablation-trainable-msvamp}
\end{table*}

\subsection{Impact of Multilingual Encoder Size on Language Alignment}
\label{apdx:encoder}
We experimented with various multilingual encoders regarding types and sizes and evaluated the corresponding reasoning performance on the XNLI dataset. Figure~\ref{fig:mt5} illustrates the XNLI performance across five different sizes of the mT5 encoder: 270M (Small), 470M (Base), 820M (Large), 2.2B (XL), and 6.7B (XXL). Additionally, we evaluate five types of multilingual encoders: mT5, mBERT, XLM, XLM-R, and LaBSE.
Our findings indicate that performance improves significantly as the encoder size increases from 270M to 2.2B for the mT5 encoder, with diminishing returns observed as the model scales beyond 2.2B to 6.7B. These results suggest that while enlarging the encoder size leads to better performance, there is a point beyond which further scaling offers limited improvements. In terms of encoder types, we observe that encoders with stronger language alignment capabilities yield better performance. Specifically, more powerful multilingual models like LaBSE, which is fine-tuned with parallel corpus, are more effective at enhancing the capabilities of low-resource languages, demonstrating that greater language alignment leads to better transfer and generalization across diverse languages.

\subsection{Adapting to different types and scales of LLMs}
\label{apdx:differentLLMs}
LinguaLIFT can be flexibly adapted to different LLMs. To validate this, we conducted experiments on Mistral-7B~\citep{jiang2023mistral} and the larger Llama-2-13B~\citep{touvron2023llama}. As shown in Tables~\ref{apdx-llama2-13b-mgsm} and \ref{apdx-mistral-mgsm}, LinguaLIFT outperforms various baselines, with average accuracy improvements of at least 1.6\% and 0.9\% for low-resource languages based on Llama-2-13B and Mistral-7B, respectively. These results demonstrate LinguaLIFT's potential for broader applicability across LLMs.

\begin{table*}[!htbp]
\centering
\resizebox{\linewidth}{!}{
\begin{tabular}{cccccccccccccc}
\hline
\multicolumn{1}{l}{\textbf{Llama-2-13B as base model}} &
  \textbf{Bn} &
  \textbf{De} &
  \textbf{En} &
  \textbf{Es} &
  \textbf{Fr} &
  \textbf{Ja} &
  \textbf{Ru} &
  \textbf{Sw} &
  \textbf{Th} &
  \textbf{Zh} &
  \textbf{LR.} &
  \textbf{HR.} &
  \textbf{Avg.} \\ \hline
\multicolumn{14}{c}{\textsc{baseline}} \\
\rowcolor[HTML]{E6F9E6} 
\multicolumn{14}{l}{\cellcolor[HTML]{E6F9E6}\textit{\textbf{Mono-SFT}}} \\
\rowcolor[HTML]{E6F9E6} 
\cellcolor[HTML]{E6F9E6}MetaMath\textsuperscript{\textasteriskcentered} &
  11.6 &
  64.8 &
  67.2 &
  65.2 &
  65.2 &
  42.8 &
  63.6 &
  7.60 &
  6.40 &
  49.2 &
  8.53 &
  59.7 &
  44.4 \\
\rowcolor[HTML]{FFFCC1} 
\multicolumn{14}{l}{\cellcolor[HTML]{FFFCC1}\textit{\textbf{Multi-SFT}}} \\
\rowcolor[HTML]{FFFCC1} 
\cellcolor[HTML]{FFFCC1}MathOctopus-Parallel\textsuperscript{\textasteriskcentered} &
  35.2 &
  44.4 &
  53.2 &
  \cellcolor[HTML]{FFFCC1}48.0 &
  48.4 &
  43.2 &
  47.6 &
  42.8 &
  46.8 &
  48.8 &
  41.6 &
  47.7 &
  45.8 \\
\rowcolor[HTML]{FFFCC1} 
\cellcolor[HTML]{FFFCC1}QAlign-MetaMathQA\textsuperscript{\textasteriskcentered} &
  38.4 &
  62.0 &
  69.2 &
  \cellcolor[HTML]{FFFCC1}67.2 &
  62.4 &
  52.4 &
  64.4 &
  46.0 &
  49.6 &
  59.2 &
  44.7 &
  62.4 &
  57.1 \\
\rowcolor[HTML]{DCF2FC} 
\multicolumn{14}{l}{\cellcolor[HTML]{DCF2FC}\textit{\textbf{Leveraging External Tools or Models}}} \\
\rowcolor[HTML]{DCF2FC} 
\cellcolor[HTML]{DCF2FC}Translate-En\textsuperscript{\textasteriskcentered} &
  34.8 &
  53.6 &
  70.8 &
  62.4 &
  54.0 &
  44.4 &
  45.6 &
  44.4 &
  54.0 &
  58.0 &
  44.4 &
  55.5 &
  52.2 \\
\rowcolor[HTML]{DCF2FC} 
LangBridge\textsuperscript{\textasteriskcentered} &
  39.2 &
  55.2 &
  65.2 &
  60.8 &
  54.8 &
  33.6 &
  58.8 &
  42.0 &
  42.8 &
  42.0 &
  41.3 &
  52.9 &
  49.4 \\
\rowcolor[HTML]{DCF2FC} 
\cellcolor[HTML]{DCF2FC}MindMerger-Soft\textsuperscript{\textasteriskcentered} &
  55.2 &
  65.2 &
  68.8 &
  69.6 &
  63.6 &
  60.0 &
  68.0 &
  56.4 &
  59.6 &
  60.4 &
  57.1 &
  65.1 &
  62.7 \\ \hline
\multicolumn{14}{c}{\textsc{our methods}} \\
\textbf{LinguaLIFT} &
  57.6 &
  64.4 &
  67.8 &
  70.4 &
  64.0 &
  59.8 &
  67.2 &
  57.8 &
  60.8 &
  59.4 &
  58.7 &
  64.7 &
  62.9 \\ \hline
\end{tabular}
}
\caption{Experimental Results on the MGSM Dataset. "LR." "HR." and "Avg." represent the average performance for low-resource languages, high-resource languages, and all languages, respectively. Following prior work (citation: LLaMA2), we classify Bn, Te, Th, and Sw as low-resource languages, while the remaining languages are categorized as high-resource. The asterisk symbol (\textasteriskcentered) denotes results taken directly from the published results of~\citet{zhu-etal-2024-question,huang2024mindmerger}.}
\label{apdx-llama2-13b-mgsm}
\end{table*}

\begin{table*}[!htbp]
\centering
\resizebox{\linewidth}{!}{
\begin{tabular}{cccccccccccccc}
\hline
\multicolumn{1}{l}{\textbf{Mistral-7B as base model}} &
  \textbf{Bn} &
  \textbf{De} &
  \textbf{En} &
  \textbf{Es} &
  \textbf{Fr} &
  \textbf{Ja} &
  \textbf{Ru} &
  \textbf{Sw} &
  \textbf{Th} &
  \textbf{Zh} &
  \textbf{LR.} &
  \textbf{HR.} &
  \textbf{Avg.} \\ \hline
\multicolumn{14}{c}{\textsc{baseline}} \\
\rowcolor[HTML]{E6F9E6} 
\multicolumn{14}{l}{\cellcolor[HTML]{E6F9E6}\textit{\textbf{Mono-SFT}}} \\
\rowcolor[HTML]{E6F9E6} 
\cellcolor[HTML]{E6F9E6}MetaMath\textsuperscript{\textasteriskcentered} &
  38.4 &
  70.4 &
  78.0 &
  71.2 &
  70.8 &
  50.8 &
  67.2 &
  16.8 &
  34.8 &
  57.2 &
  30.0 &
  66.5 &
  55.6 \\
\rowcolor[HTML]{FFFCC1} 
\multicolumn{14}{l}{\cellcolor[HTML]{FFFCC1}\textit{\textbf{Multi-SFT}}} \\
\rowcolor[HTML]{FFFCC1} 
\cellcolor[HTML]{FFFCC1}MathOctopus-Parallel\textsuperscript{\textasteriskcentered} &
  44.0 &
  50.0 &
  58.4 &
  53.2 &
  47.2 &
  48.0 &
  49.6 &
  51.6 &
  48.8 &
  51.6 &
  48.1 &
  51.1 &
  50.2 \\
\rowcolor[HTML]{FFFCC1} 
\cellcolor[HTML]{FFFCC1}QAlign-MetaMathQA\textsuperscript{\textasteriskcentered} &
  45.6 &
  59.2 &
  65.8 &
  63.6 &
  59.8 &
  49.4 &
  60.2 &
  55.2 &
  51.2 &
  57.2 &
  50.7 &
  59.3 &
  56.7 \\
\rowcolor[HTML]{DCF2FC} 
\multicolumn{14}{l}{\cellcolor[HTML]{DCF2FC}\textit{\textbf{Leveraging External Tools or Models}}} \\
\rowcolor[HTML]{DCF2FC} 
\cellcolor[HTML]{DCF2FC}Translate-En\textsuperscript{\textasteriskcentered} &
  54.6 &
  50.4 &
  69.7 &
  58.6 &
  56.7 &
  57.2 &
  64.9 &
  47.7 &
  58.7 &
  63.1 &
  53.7 &
  60.1 &
  58.2 \\
\rowcolor[HTML]{DCF2FC} 
\cellcolor[HTML]{DCF2FC}LangBridge\textsuperscript{\textasteriskcentered} &
  50.0 &
  68.4 &
  65.6 &
  65.6 &
  68.8 &
  58.4 &
  68.4 &
  47.2 &
  60.0 &
  65.6 &
  52.4 &
  65.8 &
  61.8 \\
\rowcolor[HTML]{DCF2FC} 
\cellcolor[HTML]{DCF2FC}MindMerger-Soft\textsuperscript{\textasteriskcentered} &
  57.6 &
  69.2 &
  79.2 &
  71.6 &
  69.6 &
  57.2 &
  68.4 &
  53.2 &
  59.6 &
  68.8 &
  56.8 &
  69.1 &
  65.4 \\ \hline
\multicolumn{14}{c}{\textsc{our methods}} \\
\textbf{LinguaLIFT} &
  58.4 &
  67.2 &
  77.6 &
  72.1 &
  70.6 &
  56.4 &
  66.8 &
  58.4 &
  63.2 &
  68.2 &
  60.0 &
  68.4 &
  65.9 \\ \hline
\end{tabular}
}
\caption{Experimental Results on the MGSM Dataset. "LR." "HR." and "Avg." represent the average performance for low-resource languages, high-resource languages, and all languages, respectively. Following prior work \citep{huang2024mindmerger}, we classify Bn, Te, Th, and Sw as low-resource languages, while the remaining languages are categorized as high-resource.  The asterisk symbol (\textasteriskcentered) denotes results taken directly from the published results of~\citet{zhu-etal-2024-question,huang2024mindmerger}.}
\label{apdx-mistral-mgsm}
\end{table*}

\subsection{Language Transferability in Language Families and Writing Systems}
\label{apdx:language-family}

\begin{table*}[!htbp]
\centering
\resizebox{\linewidth}{!}{
\begin{tabular}{@{}cccccc@{}}
\toprule
\textbf{Language Family} &
  \textbf{Language} &
  \textbf{MetaMath-7B} &
  \textbf{MathOctopus-Parallel-7B} &
  \textbf{MindMerger} &
  \textbf{AlignIFT-MetaMath} \\ \midrule
\multicolumn{1}{c|}{\multirow{4}{*}{\textbf{Indo-European-Germanic}}} & af               & 32.4 & 19.1 & 40.2 & 46.4 \\
\multicolumn{1}{c|}{}                                                 & is               & 14.3 & 10.5 & 35.6 & 41.9 \\
\multicolumn{1}{c|}{}                                                 & lb               & 14.6 & 8.75 & 37.6 & 43.4 \\
\multicolumn{1}{c|}{}                                                 & \textbf{Average} & 20.4 & 12.8 & 37.8 & 43.9 \\ \midrule
\multicolumn{1}{c|}{\multirow{4}{*}{\textbf{Indo-European-Slavic}}}   & be               & 17.9 & 14.6 & 37.1 & 39.6 \\
\multicolumn{1}{c|}{}                                                 & mk               & 30.1 & 15.9 & 40.7 & 42.9 \\
\multicolumn{1}{c|}{}                                                 & sk               & 32.3 & 16.3 & 38.6 & 42.5 \\
\multicolumn{1}{c|}{}                                                 & \textbf{Average} & 26.8 & 15.6 & 38.8 & 41.7 \\ \midrule
\multicolumn{1}{c|}{\multirow{6}{*}{\textbf{Indo-European-Indo-Iranian}}} &
  bn &
  9.86 &
  20.1 &
  37.6 &
  39.5 \\
\multicolumn{1}{c|}{}                                                 & gu               & 4.81 & 6.17 & 35.8 & 39.0 \\
\multicolumn{1}{c|}{}                                                 & hi               & 24.5 & 11.1 & 39.4 & 41.7 \\
\multicolumn{1}{c|}{}                                                 & mr               & 9.62 & 6.78 & 34.3 & 32.6 \\
\multicolumn{1}{c|}{}                                                 & ne               & 8.38 & 6.54 & 37.9 & 38.6 \\
\multicolumn{1}{c|}{}                                                 & \textbf{Average} & 11.4 & 10.1 & 37.0 & 38.3 \\ \midrule
\multicolumn{1}{c|}{\multirow{5}{*}{\textbf{Dravidian}}}              & ta               & 6.29 & 3.95 & 35.0 & 36.5 \\
\multicolumn{1}{c|}{}                                                 & te               & 6.17 & 10.2 & 37.1 & 36.9 \\
\multicolumn{1}{c|}{}                                                 & kn               & 6.41 & 5.43 & 33.8 & 33.4 \\
\multicolumn{1}{c|}{}                                                 & ml               & 5.18 & 5.92 & 37.4 & 36.7 \\
\multicolumn{1}{c|}{}                                                 & \textbf{Average} & 6.01 & 6.38 & 35.8 & 35.9 \\ \midrule
\multicolumn{1}{c|}{\multirow{7}{*}{\textbf{Other}}}                  & ar               & 23.6 & 8.63 & 40.2 & 42.5 \\
\multicolumn{1}{c|}{}                                                 & eu               & 6.29 & 5.55 & 35.4 & 35.6 \\
\multicolumn{1}{c|}{}                                                 & ha               & 5.92 & 5.55 & 29.1 & 33.4 \\
\multicolumn{1}{c|}{}                                                 & hy               & 5.80 & 6.04 & 34.7 & 39.7 \\
\multicolumn{1}{c|}{}                                                 & sw               & 7.52 & 21.7 & 36.4 & 47.6 \\
\multicolumn{1}{c|}{}                                                 & th               & 13.4 & 21.3 & 38.7 & 41.7 \\
\multicolumn{1}{c|}{}                                                 & \textbf{Average} & 10.4 & 11.5 & 35.8 & 40.1 \\ \bottomrule
\end{tabular}
}
\caption{The detailed experimental results of LinguaLIFT on MMWP benchmark grouped by language family.}
\label{language-family-details}
\end{table*}

\begin{table*}[!htbp]
\centering
\resizebox{\linewidth}{!}{
\begin{tabular}{@{}cccccc@{}}
\toprule
\textbf{Writing Scripts} & \textbf{Language} & \textbf{MetaMath-7B} & \textbf{MathOctopus-Parallel-7B} & \textbf{MingMerger} & \textbf{AlignIFT-MetaMath} \\ \midrule
\multicolumn{1}{c|}{\multirow{7}{*}{\textbf{Latin}}}      & af               & 32.4 & 19.1 & 40.2 & 46.4 \\
\multicolumn{1}{c|}{}                                     & eu               & 6.29 & 5.55 & 35.4 & 35.6 \\
\multicolumn{1}{c|}{}                                     & is               & 14.3 & 10.5 & 35.6 & 41.9 \\
\multicolumn{1}{c|}{}                                     & lb               & 14.6 & 8.8  & 37.6 & 43.4 \\
\multicolumn{1}{c|}{}                                     & sw               & 7.52 & 21.7 & 36.4 & 47.6 \\
\multicolumn{1}{c|}{}                                     & sk               & 32.3 & 16.3 & 38.6 & 42.5 \\
\multicolumn{1}{c|}{}                                     & \textbf{Average} & 17.9 & 13.7 & 37.3 & 42.9 \\ \midrule
\multicolumn{1}{c|}{\multirow{3}{*}{\textbf{Arabic}}}     & ar               & 23.6 & 8.63 & 40.2 & 42.5 \\
\multicolumn{1}{c|}{}                                     & ha               & 5.92 & 5.55 & 29.1 & 33.4 \\
\multicolumn{1}{c|}{}                                     & \textbf{Average} & 14.8 & 7.09 & 34.7 & 38.0 \\ \midrule
\multicolumn{1}{c|}{\multirow{3}{*}{\textbf{Cyrillic}}}   & be               & 17.8 & 14.6 & 37.1 & 39.6 \\
\multicolumn{1}{c|}{}                                     & mk               & 30.1 & 15.9 & 40.7 & 42.9 \\
\multicolumn{1}{c|}{}                                     & \textbf{Average} & 24.0 & 15.3 & 38.9 & 41.3 \\ \midrule
\multicolumn{1}{c|}{\multirow{4}{*}{\textbf{Devanagari}}} & hi               & 24.5 & 11.1 & 39.4 & 41.7 \\
\multicolumn{1}{c|}{}                                     & mr               & 9.6  & 6.8  & 34.3 & 32.6 \\
\multicolumn{1}{c|}{}                                     & ne               & 8.4  & 6.5  & 37.9 & 38.6 \\
\multicolumn{1}{c|}{}                                     & \textbf{Average} & 14.2 & 8.1  & 37.2 & 37.6 \\ \midrule
\multicolumn{1}{c|}{\multirow{9}{*}{\textbf{Other}}}      & ta               & 6.29 & 3.95 & 35.0 & 36.5 \\
\multicolumn{1}{c|}{}                                     & te               & 6.17 & 10.2 & 37.1 & 36.9 \\
\multicolumn{1}{c|}{}                                     & ml               & 5.18 & 5.92 & 37.4 & 36.7 \\
\multicolumn{1}{c|}{}                                     & bn               & 9.86 & 20.1 & 37.6 & 39.5 \\
\multicolumn{1}{c|}{}                                     & hy               & 5.8  & 6.04 & 34.7 & 39.7 \\
\multicolumn{1}{c|}{}                                     & kn               & 6.41 & 5.43 & 33.8 & 33.4 \\
\multicolumn{1}{c|}{}                                     & gu               & 4.81 & 6.17 & 35.8 & 39.0 \\
\multicolumn{1}{c|}{}                                     & th               & 13.4 & 21.3 & 38.7 & 41.7 \\
\multicolumn{1}{c|}{}                                     & \textbf{Average} & 7.6  & 10.8 & 36.3 & 38.3 \\ \bottomrule
\end{tabular}
}
\caption{The detailed experimental results of LinguaLIFT on MMWP benchmark grouped by writing systems.}
\label{writing-scripts-details}
\end{table*}

In our analysis, we considered the influence of language families and writing systems on LinguaLIFT's performance. This approach allowed us to delve into the nuances of language transferability, particularly in low-resource languages that share linguistic characteristics with English.

We organized the MMWP benchmark results according to language family and writing system, following the categorization proposed by~\citet{zhang-etal-2023-multilingual}. As depicted in Figure~\ref{fig:language-family-writing-systems}, and detailed results of specific low-resource languages reasoning tasks presented in Table~\ref{language-family-details}, LinguaLIFT demonstrates superior performance over all baseline models across different language families. Notably, it excels in languages from the Indo-European-Indo-Iranian, Indo-European-Germanic, and Dravidian families. The Indo-European-Germanic languages, which share a common lineage with English, register stronger performance, underscoring the potential benefits of shared ancestry in cross-lingual transfer. In contrast, languages that diverge significantly from English in terms of syntax, such as those from the Dravidian family, present a more challenging scenario, yet LinguaLIFT still yields impressive results.

In Figure~\ref{fig:language-family-writing-systems} and in Table~\ref{writing-scripts-details}, we also observed similar trends in the context of writing system transfer. LinguaLIFT exhibits exceptional performance in languages that employ shared scripts, particularly those using Latin-based orthographies. This observation underscores the critical role of script similarity in facilitating the transfer of knowledge across languages. The shared orthographic features can provide an additional layer of shared structure that aids in the cross-lingual transfer process, enabling more effective learning and translation across related languages.

In summary, our findings underscore the impressive transferability of LinguaLIFT across different low-resource languages from various language families and writing systems. The method significantly enhances cross-lingual transfer, particularly when linguistic factors such as language family and writing system align with English ones. These results highlight the potential of LinguaLIFT as a robust approach for multilingual processing, capable of leveraging shared linguistic characteristics for improved performance, even in low-resource settings.

\subsection{Analysis of Code-Switch Tuning}
\label{apdx:code-switch}

In our experiment, we aimed to scrutinize the influence of the part-of-speech (POS) of the substitution words on the multilingual reasoning performance in the context of code-switched tuning. This analysis is essential as it provides insights into the role of various word categories and their combinations in the effectiveness of language model tuning and reasoning capabilities.

\begin{table}[!htbp]
\centering
\begin{tabular}{cc}
\hline
\textbf{Part-of-Speech} & \textbf{\#Nums} \\ \hline
Noun                    & 84726           \\
Verb                    & 42451           \\
Adposition              & 37727           \\
Pronoun                 & 25012           \\
Adjective               & 21790           \\
Conjunction             & 20764           \\
Auxiliary               & 16409           \\ 
Adverb                  & 9150            \\ \hline
\end{tabular}
\caption{Part-of-speech statistics results from the MGSM English test set.}
\label{pos-stats}
\end{table}

\begin{figure*}[!htbp]
    \centering
    \includegraphics[width=1\linewidth]{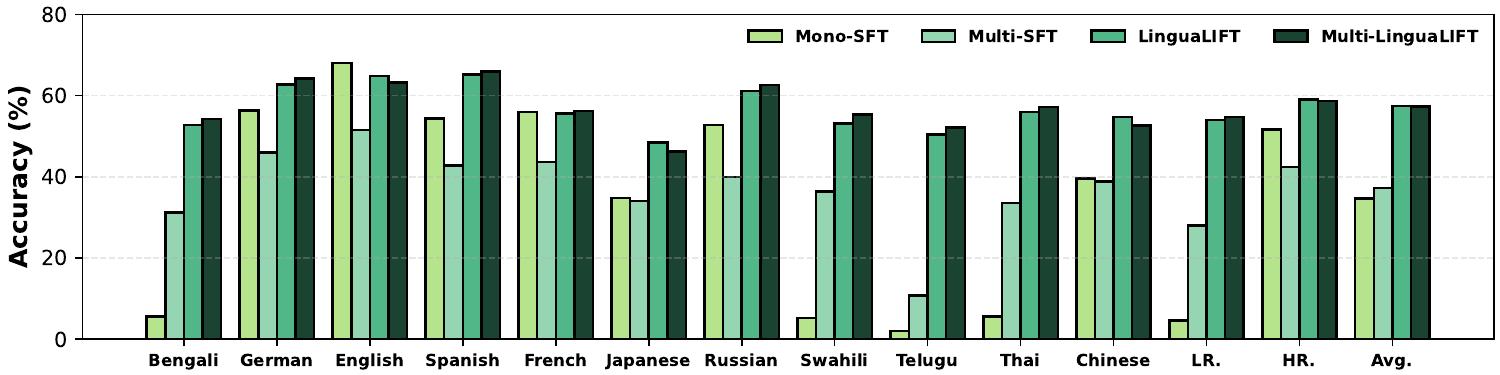}
    \caption{Effects of tuning LLM with mixed supervised data. Generally, incorporating multilingual supervised data into LinguaLIFT can achieve a higher ceiling for low-resource language reasoning performance.}
    \label{fig:multilingual-LinguaLIFT}
\end{figure*}

Our initial step involved an examination of the distribution of part-of-speech (POS) tags in the reasoning queries, the results of which are tabulated in Table~\ref{pos-stats}. This analysis facilitated a preliminary understanding of the prevalence of different word classes within the reasoning queries, setting the stage for further investigation into their impact on multilingual reasoning.

Subsequently, we categorized the POS combinations into three distinct groups for a more granular analysis. The first group comprised individual POS categories, specifically Verbs, Adpositions, and Pronouns. These categories were selected due to their fundamental role in sentence construction and their potential to affect the meaning and structure of statements significantly.

The second group consisted of syntactic function combinations, including Verb+Adverb, Adjective+Adverb, and Pronoun+Auxiliary+Conjunction combinations. These combinations were chosen based on their syntactic roles and capacity to influence sentence structure and meaning. They are integral to creating complex sentence structures and are often pivotal in conveying nuanced meanings.

The third group focused on key syntactic structures: Subject-Verb (Noun+Verb) and Prepositional Phrases (Adposition+Noun). These structures were selected due to their central role in sentence construction and their potential to encapsulate core semantic information. They form the backbone of many sentence structures and play a crucial role in interpreting a sentence's meaning.

By analyzing the impact of these POS categories and combinations on the performance of code-switched tuning, we aim to provide a comprehensive understanding of the interplay between syntax and semantics in the context of multilingual reasoning. This analysis will shed light on the importance of different word classes and their combinations in improving the efficacy of multilingual language models.

Further experiments were conducted to investigate the effects of replacing different types of words (e.g., nouns, verbs, prepositions, etc.) on the multilingual reasoning capabilities of LinguaLIFT models. As indicated in Figure~\ref{fig:ablation_pos}, we observed that replacing nouns significantly impacted the model's reasoning performance within the individual POS categories. This finding leads to the hypothesis that reasoning in query sentences may be closely tied to noun-based understanding, possibly due to the central role nouns play in representing key entities and concepts in the task.

Additionally, we found that syntactic structures involving Subject-Verb combinations and Prepositional Phrases performed the best, surpassing all other POS combinations. This suggests that core arguments (subjects and verbs) and their relational elements (prepositions) are crucial for capturing the key relationships in reasoning tasks.

In contrast, the Adjective+Adverb combination showed the weakest performance, indicating that modifiers are less critical for reasoning tasks. This supports the rationale of previous work~\citep{gaur-saunshi-2023-reasoning,mirzadeh2024gsmsymbolicunderstandinglimitationsmathematical} that proposed transforming mathematical word problems into symbolic reasoning tasks.

Overall, these findings highlight that, for reasoning tasks, word alignment—particularly involving nouns and verbs—is sufficient to enable the model to generalize reasoning abilities across multiple languages.

\subsection{Incorporating multilingual reasoning data in LinguaLIFT}
\label{apdx:incorporate-multilingual}

In some settings, we observed that incorporating multilingual reasoning data during the first stage can further enhance task transfer performance in the second stage. This is evidenced by the improved performance compared to the pure English-only reasoning task training. In the mixed supervision setting, we first fine-tune the language alignment layer during the Language-Align stage using multilingual reasoning instruction data \textsc{GSM8KINSTRUCT}~\citep{chen2023breaking} and code-switching translation instruction data, followed by fine-tuning LLM with English-only instruction data, \textsc{MetaMathQA}. The experimental results on \textsc{MGSM} are presented in Figure~\ref{fig:multilingual-LinguaLIFT}. We find that incorporating additional multilingual supervision yields an average performance gain of 1.0\% on low-resource reasoning tasks and allows the model to achieve comparable performance on high-resource reasoning tasks relative to the vanilla setting.

\section{Supplementary Experiments}

\subsection{Quantitative Analysis of the Correlation between Multilingual Alignment and Reasoning Performance}
\label{apdx:correlation}

To evaluate the cross-lingual alignment capabilities of LLMs, we employ the multilingual sentence retrieval benchmark \textbf{Tatoeba} \citep{artetxe-schwenk-2019-massively}, a widely used dataset for assessing ML-LMs. The dataset includes up to 1,000 sentences per language, along with their English translations. In our study, we focus on a subset of the original benchmark that aligns with the language categories used in the reasoning test sets MGSM, ensuring consistency with our experimental design.

We adopt the evaluation procedure outlined in XTREME~\citep{pmlr-v119-hu20b} to evaluate cross-lingual alignment. For each language pair, we compute the cosine similarity between sentences in the source language and their closest counterparts in the target language. Following the methods of \citet{yang-etal-2021-simple} and \citet{xie-etal-2022-discovering}, we use top-1 retrieval accuracy as a metric to quantify alignment between the two languages. A higher retrieval accuracy score indicates a stronger alignment.

\begin{figure*}[!htbp]
	\centering
 	\subfloat[MGSM]{\includegraphics[width=.5\linewidth]{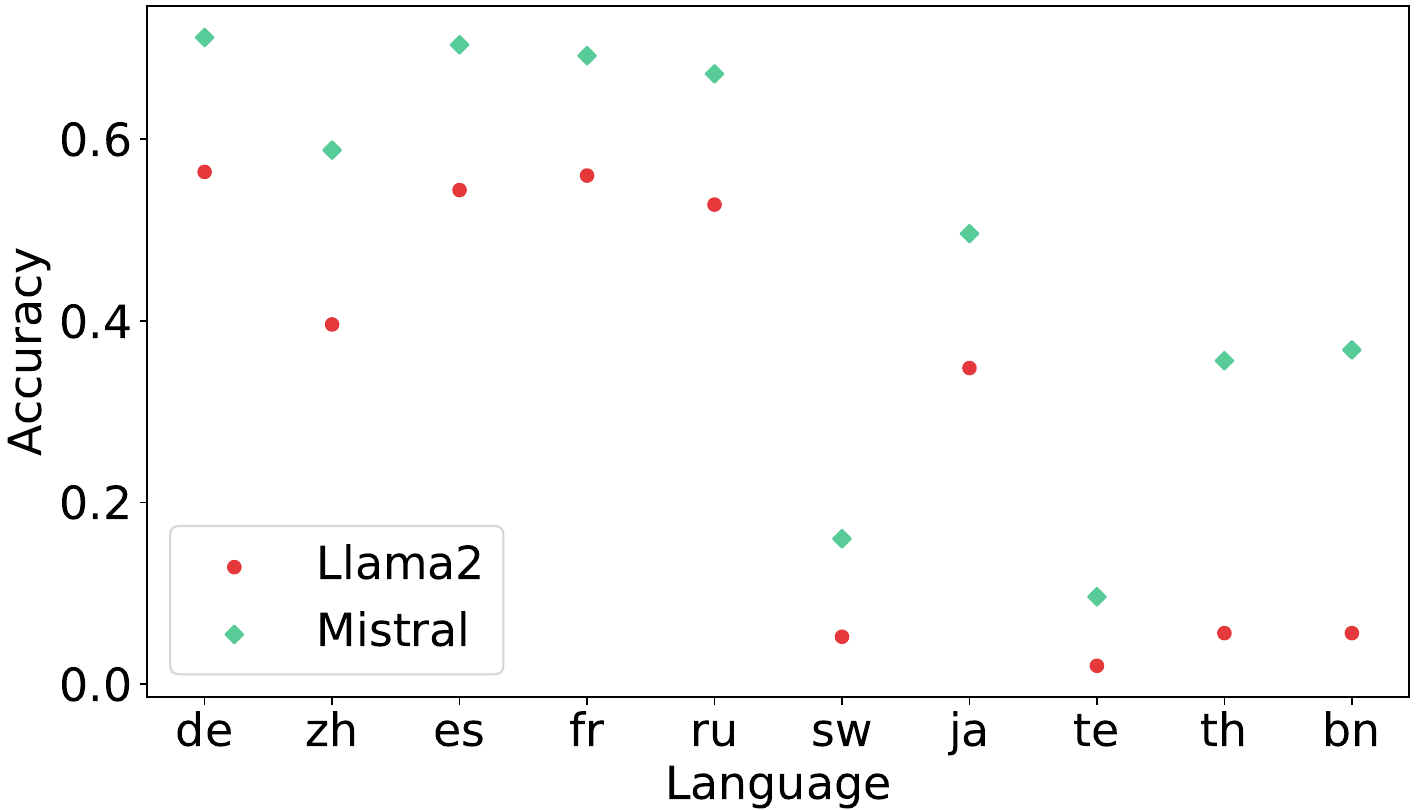}\label{mgsm_trend}}
	\subfloat[MSVAMP]{\includegraphics[width=.5\linewidth]{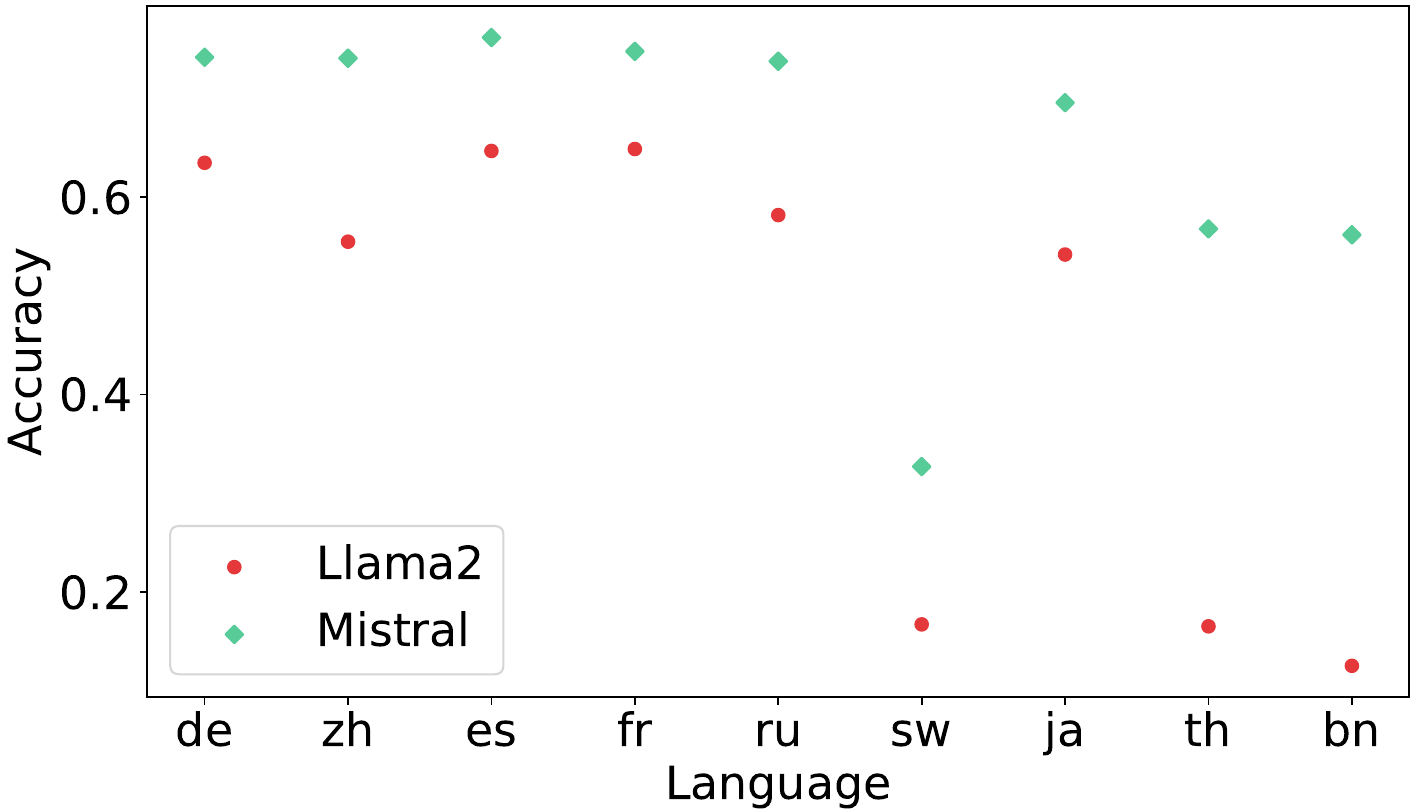}\label{msvamp_trend}}
	\caption{Performances on MGSM and MSVAMP test set, where languages are sorted by decreasing retrieval accuracy. The trend indicates the mathemtical reasoning performance worsens along with the decreasing cross-lingual alignment degree.}
    \label{tab:reasoning_trend}
\end{figure*}

\begin{table*}[!htbp]
\centering
\resizebox{\linewidth}{!}{
\begin{tabular}{@{}c|c|ccccccccccc|ccc@{}}
\toprule
\textbf{Architecture} &
  \textbf{\# Params} &
  \textbf{Bn} &
  \textbf{De} &
  \textbf{En} &
  \textbf{Es} &
  \textbf{Fr} &
  \textbf{Ja} &
  \textbf{Ru} &
  \textbf{Sw} &
  \textbf{Te} &
  \textbf{Th} &
  \textbf{Zh} &
  \textbf{LR.} &
  \textbf{HR.} &
  \textbf{Avg.} \\ \midrule
Linear       & 4 M  & 53.1 & 61.6 & 63.9 & 63.8 & 56.6 & 46.0 & 56.6 & 49.0 & 44.4 & 50.6 & 52.2 & 49.3 & 57.2 & 54.4 \\
2 layers MLP & 10 M & 54.4 & 62.0 & 64.8 & 63.6 & 56.8 & 50.0 & 60.4 & 55.6 & 54.0 & 57.6 & 54.0 & 55.4 & 58.8 & 57.6 \\
3 layers MLP & 14 M & 53.8 & 62.8 & 66.6 & 56.6 & 58.8 & 50.0 & 59.4 & 53.8 & 53.6 & 58.2 & 53.4 & 54.9 & 58.2 & 57.0 \\ \bottomrule
\end{tabular}
}
\caption{The ablation experiments of the selection of language alignment layer results on the MGSM Dataset. "LR." "HR." and "Avg." represent the average performance for low-resource languages, high-resource languages, and all languages, respectively. Following prior work \citep{huang2024mindmerger}, we classify Bn, Te, Th, and Sw as low-resource languages, while the remaining languages are categorized as high-resource.}
\label{apdx:architecture-of-alignment}
\end{table*}

\begin{table}[!htbp]
\centering
\resizebox{\columnwidth}{!}{
\begin{tabular}{ccccc}
\hline
Test set &
  \multicolumn{2}{c}{MGSM} &
  \multicolumn{2}{c}{MSVAMP} \\
Resource-Level &
  Low-Resource &
  High-Resource &
  Low-Resource &
  High-Resource \\ \hline
$\rho$ &
  0.733\textsuperscript{\textasteriskcentered} &
  0.778\textsuperscript{\textasteriskcentered} &
  0.763\textsuperscript{\textasteriskcentered} &
  0.783\textsuperscript{\textasteriskcentered} \\ \hline
\end{tabular}
}
\caption{Spearman's rank correlation coefficient ($\rho$) between retrieval accuracy (\%) on Tatoeba and reasoning accuracy on MGSM. An asterisk (\textasteriskcentered) signifies a statistically significant correlation (p-value < 0.05).}
\label{tab:rho}
\end{table}

Figure \ref{tab:reasoning_trend} illustrates that the performance of low-resource languages lags significantly behind that of other languages in both mathematical reasoning tasks and the cross-lingual sentence retrieval task. It also shows that a higher alignment degree tends to correlate with improved mathematical reasoning performance. Additionally, we compute the Spearman's rank correlation between the retrieval accuracy scores and reasoning performance, as presented in Table~\ref{tab:rho}, which reveals a strong correlation between the two. Both the observed trend and the correlation coefficient confirm that the consistency of multilingual mathematical reasoning is closely tied to the cross-lingual alignment degree.

\subsection{Impact of Different Language Alignment Layers}
\label{apdx:la-layer}
In the main experiment, we utilize two layers of MLP as the language alignment layers to transfer multilingual alignment from the pre-trained multilingual encoder into the LLMs. We also conducted experiments to evaluate the performance of LinguaLIFT when ablating different mapping layers, as shown in Table~\ref{apdx:architecture-of-alignment}. In contrast to the findings of \citet{yoon-etal-2024-langbridge}, the two layers of MLP used in our main experiment achieved the best performance. In comparison, using a linear layer resulted in lower performance, likely due to its limited capacity to effectively adapt the pre-trained multilingual encoder to LLM, attributed to the smaller number of parameters.

\subsection{LinguaLIFT CoT Examples}
\label{apdx:case_study}
\begin{figure*}[!htbp]
    \centering
    \includegraphics[width=\linewidth]{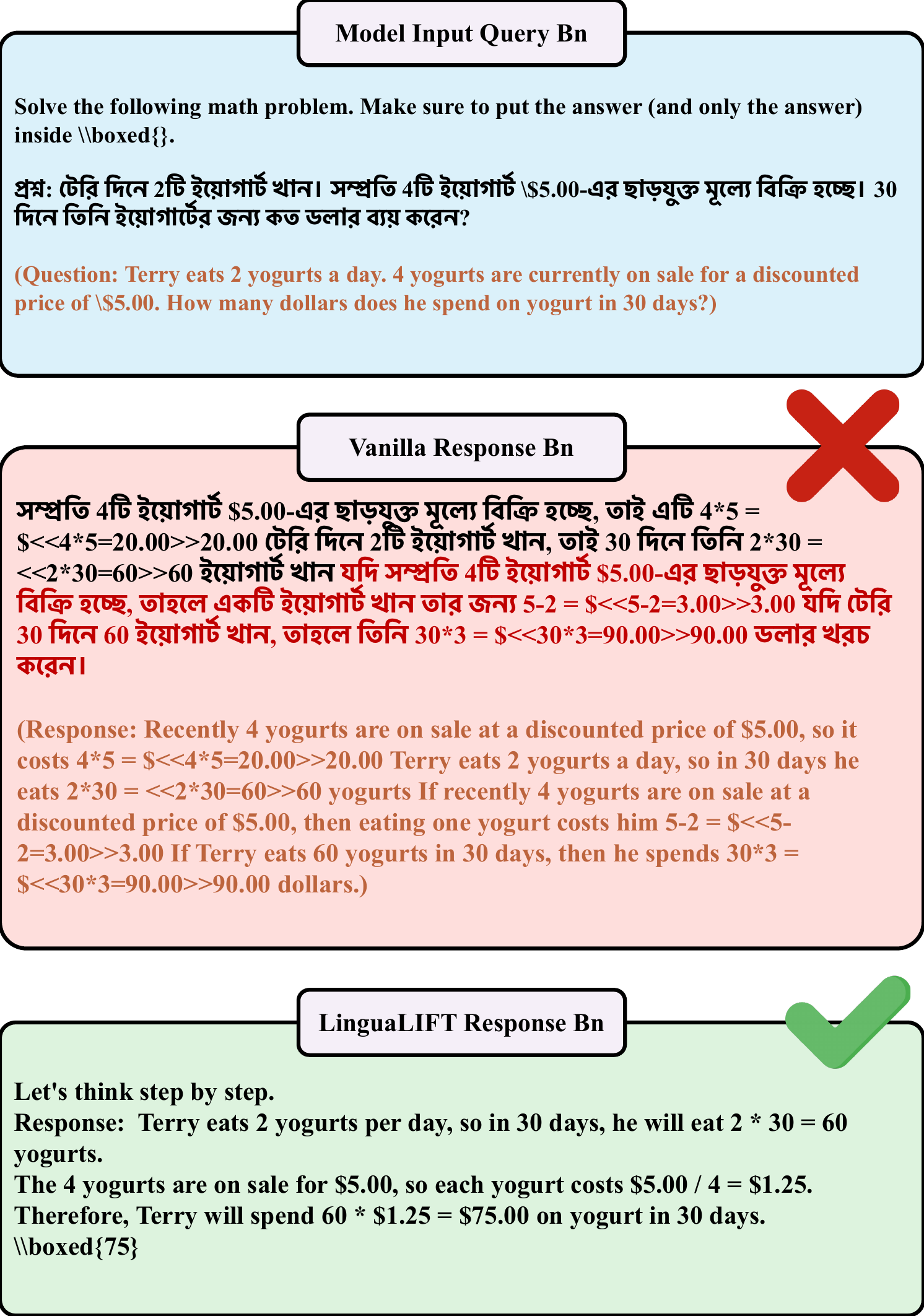}
    \caption{Example of zero-shot CoT generated by MetaMath-7B when prompted with an example from MGSM Bengali. Translation of the input and response is provided in \color[HTML]{BD643D}{brown text}.}
    \label{fig:case-study-bn}
\end{figure*}

\begin{figure*}[!htbp]
    \centering
    \includegraphics[width=1\linewidth]{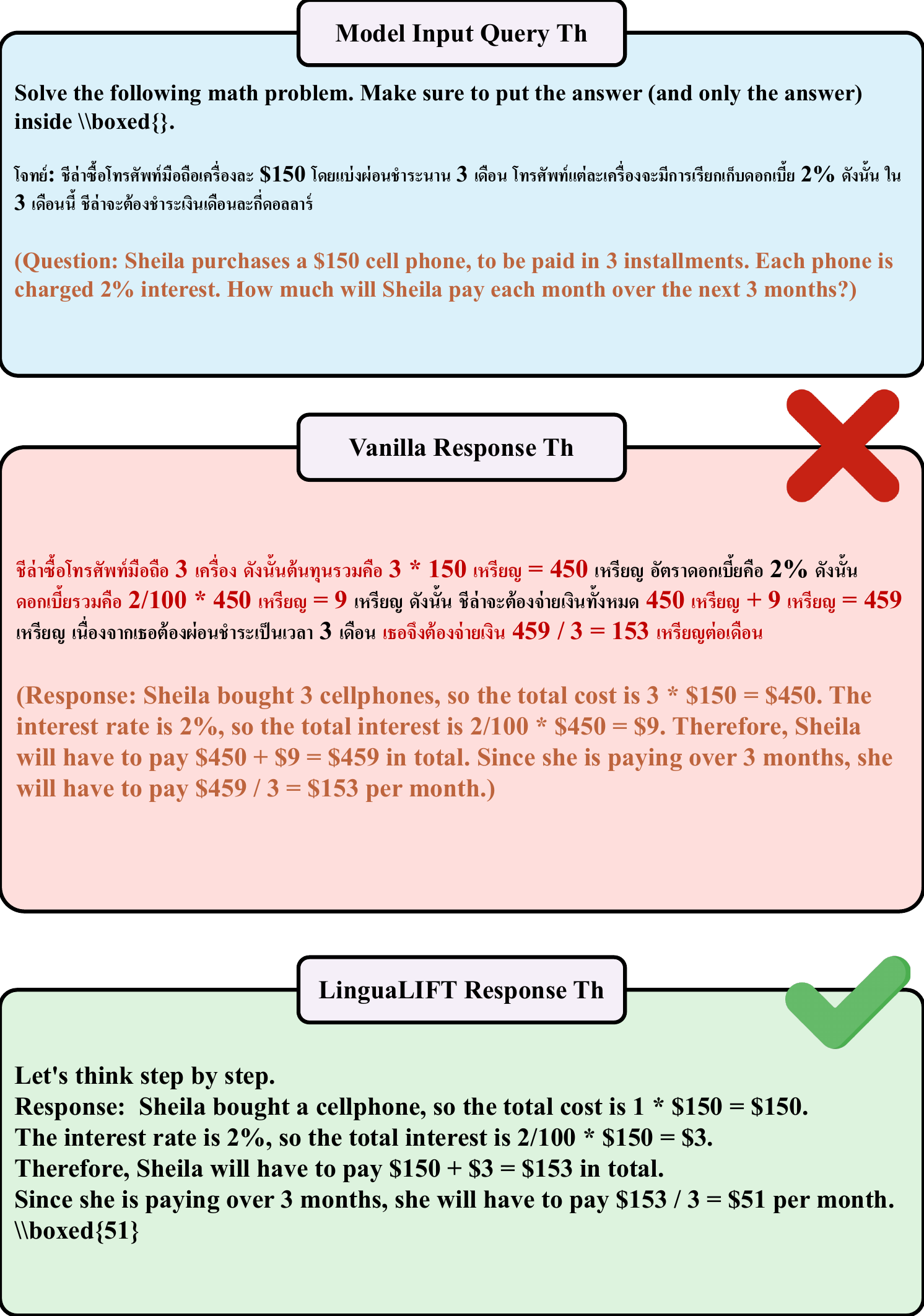}
    \caption{Example of zero-shot CoT generated by MetaMath-7B when prompted with an example from MGSM Thai. Translation of the input and response is provided in \color[HTML]{BD643D}{brown text}.}
    \label{fig:case-study-th}
\end{figure*}

\begin{figure*}[!htbp]
    \centering
    \includegraphics[width=1\linewidth]{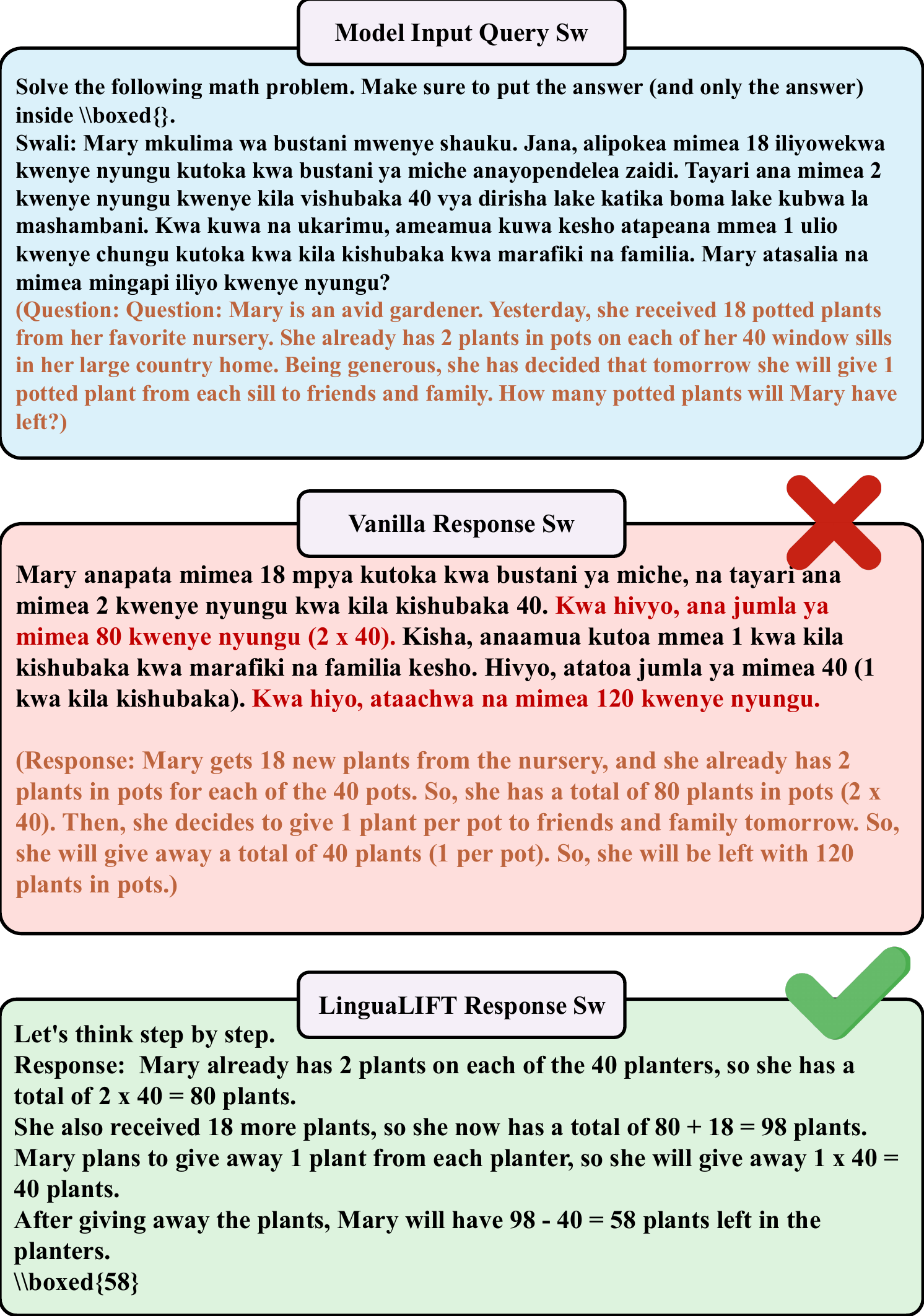}
    \caption{Example of zero-shot CoT generated by MetaMath-7B when prompted with an example from MGSM Swahili. Translation of the input and response is provided in \color[HTML]{BD643D}{brown text}.}
    \label{fig:case-study-sw}
\end{figure*}

We present three examples of CoT reasoning. These examples demonstrate LinguaLIFT's ability to effectively understand low-resource languages and generate corresponding mathematical reasoning processes using the chain-of-thought reasoning strategy.
Figure~\ref{fig:case-study-bn} illustrates a zero-shot CoT example from the MGSM Bengali dataset. The Vanilla Mono-SFT model produces incorrect reasoning, while the LinguaLIFT model generates a correct reasoning process, ultimately leading to the correct answer. Similar trends are observed in the Thai and Swahili, shown in Figures~\ref{fig:case-study-th} and \ref{fig:case-study-sw}, where the LinguaLIFT model corrects the erroneous reasoning produced by the Mono-SFT model when presented with the same problems.

\end{document}